\documentclass[]{fairmeta}
\newcommand{\PWRArxiv}{}

\usepackage{natbib}
\usepackage{multirow}
\usepackage{caption}
\usepackage[utf8]{inputenc} % allow utf-8 input
\usepackage[T1]{fontenc}    % use 8-bit T1 fonts
\usepackage{hyperref}       % hyperlinks
\usepackage{url}            % simple URL typesetting
\usepackage{booktabs}       % professional-quality tables
\usepackage{amsfonts}       % blackboard math symbols
\usepackage{nicefrac}       % compact symbols for 1/2, etc.
\usepackage{microtype}      % microtypography
\usepackage[table]{xcolor}         % colors
\usepackage{graphicx} 
\usepackage{algorithm}
\usepackage{algpseudocode}
\usepackage{amsmath}
\usepackage{amssymb}
\usepackage{enumitem}
\usepackage{colortbl}
\usepackage{pifont}
\usepackage{subcaption}
\usepackage{tcolorbox}
\usepackage{adjustbox}
\usepackage{needspace}

\usepackage{array}
\usepackage{placeins} 

\definecolor{ourgreen}{RGB}{234,243,222}
\definecolor{ablationgray}{RGB}{247,247,247}
\definecolor{darkgreen}{RGB}{0, 127, 0}
\definecolor{colTask}{HTML}{F5F7FF}
\definecolor{colPlan}{HTML}{EEF2FF}
\definecolor{colCue}{HTML}{F0F4FF}
\definecolor{colDialog}{HTML}{F3E8FF}
\definecolor{colGolden}{HTML}{F0FDF4}
\definecolor{colOurs}{HTML}{FFFBEB}
\definecolor{colThreeP}{HTML}{FEF2F2}

\newcommand{\cmark}{\ding{51}}    % ✓
\newcommand{\xmark}{\ding{55}}  
\definecolor{colCat1}{HTML}{2563EB}
\definecolor{colCat2}{HTML}{16A34A}
\definecolor{colCat3}{HTML}{D97706}
\definecolor{colCat4}{HTML}{DC2626}

\definecolor{colCat1Light}{HTML}{DBEAFE}
\definecolor{colCat2Light}{HTML}{DCFCE7}
\definecolor{colCat3Light}{HTML}{FEF9C3}
\definecolor{colCat4Light}{HTML}{FFE4E6}

\definecolor{metablue}{HTML}{0064E0}     % from fairmeta.cls
\colorlet{avgblue}{metablue!12}          % light tint for the Avg column

\definecolor{avggray}{gray}{0.93}

\newsavebox{\catbox}
\newsavebox{\turnvbox}
\newcommand{\catstripe}[2]{%
  \sbox{\catbox}{\arrayrulecolor{black!55}\setlength{\arrayrulewidth}{0.4pt}#2}%
  \leavevmode
  \begin{adjustbox}{max width=\linewidth}
  \textcolor{#1}{\vrule width 5pt height \ht\catbox depth \dp\catbox}%
  \usebox{\catbox}%
  \textcolor{#1}{\vrule width 5pt height \ht\catbox depth \dp\catbox}\end{adjustbox}}

\newcommand{\catbanner}[3]{%
  \vspace{4pt}\noindent%
  \colorbox{#1Light}{\makebox[\dimexpr\linewidth-2\fboxsep][l]{%
    \hspace{6pt}\small\textcolor{#1}{\bfseries #2}\quad\small #3\hspace{6pt}}}%
  \vspace{2pt}\\}

\definecolor{colTurnTxt}{HTML}{334155}

\usepackage{microtype}
\usepackage{silence}
\definecolor{colTurnRib}{HTML}{6D28D9}
\newcommand{\turnrib}[3]{% #1 = turn number, #2 = time range (unused), #3 = content
  \begingroup\arrayrulecolor{black!55}\setlength{\arrayrulewidth}{0.4pt}%
  \sbox{\turnvbox}{#3}%
  \noindent\begin{adjustbox}{max width=\linewidth}\begin{tabular}{@{}>{\columncolor{white}}c@{}c@{}}%
    \parbox[c][\dimexpr\ht\turnvbox+\dp\turnvbox\relax][s]{1.7em}{\centering\small\bfseries\textcolor{colTurnTxt}{%
      \strut % \vspace*
      I\\
      [\stretch{0.1}]N\\
      [\stretch{0.1}]T\\
      [\stretch{0.1}]E\\
      [\stretch{0.1}]R\\
      [\stretch{0.1}]V\\
      [\stretch{0.1}]A\\
      [\stretch{0.1}]L\\
      [\stretch{0.5}]#1\\
      \strut}}
    & \usebox{\turnvbox} \\%
  \end{tabular}\end{adjustbox}\endgroup}
\newcommand{\turnpad}[1]{%
  \noindent\begingroup\arrayrulecolor{black!55}\setlength{\arrayrulewidth}{0.4pt}%
  \begin{adjustbox}{max width=\linewidth}\begin{tabular}{@{}m{1.7em}@{}c@{}}{} & #1 \\\end{tabular}\end{adjustbox}\endgroup}

\crefname{figure}{Figure}{Figures}
\Crefname{figure}{Figure}{Figures}

\title{Plan, Watch, Recover: A Benchmark and Architectures for Proactive Procedural Assistance}

\author[1,2,*]{Kaustav Kundu}
\author[1,*]{Ritvik Shrivastava}
\author[1]{Maxim Arap}
\author[1]{Nanshu Wang}
\author[1]{Xianhui Zhu}
\author[1,2]{Quintin Fettes}
\author[1]{Gautam Tiwari}
\author[1]{Parth Suresh}
\author[1]{Prateek Singhi}
\author[2]{Théo Moutakanni}
\author[2]{Alejandro Castillejo Munoz}
\author[2]{Allen Bolourchi}
\author[2]{Pascale Fung}
\author[1]{Pinar Donmez}
\author[1]{Babak Damavandi}
\author[1]{Anuj Kumar}
\author[1, \dagger]{Seungwhan Moon}

\affiliation[1]{Meta Reality Labs}
\affiliation[2]{Meta Superintelligence Labs}

\contribution[*]{Joint First Authors}
\contribution[\dagger]{Project Lead}

\abstract{\begin{abstract}
    
Proactive multimodal assistants must decide \textit{when} guidance is needed and \textit{what} to say, while tracking progress through extended procedures and helping users recover when they depart from the expected plan.
Progress is limited by benchmarks that score isolated clips, cover narrow domains, or omit realistic deviations.
We introduce \textbf{EgoProactive}, a smart-glasses dataset of procedures across four domains in which users commit step-level deviations and recover from them, paired with reference guidance.
We combine it with five re-annotated egocentric and instructional datasets to form \textbf{Pro\textsuperscript{2}Bench}, a cross-domain benchmark that scores intervention timing, procedure completion, and response quality.
Zero-shot, no frontier or open-weight model completes more than $26\%$ of a procedure's interventions before its first error.
We propose \textbf{PWR} (Plan, Watch, Recover), which keeps an explicit plan state in the model's context and revises it after every intervention, including recovery steps after a deviation.
Across three open-weight backbones, training on Pro\textsuperscript{2}Bench without a plan barely changes performance, while adding the dynamic plan raises average procedure completion from $20\%$ to $27\%$ and improves turn-level decisions and response quality for every backbone.
The gains hold on held-out datasets and on deviations never seen in training.
\end{abstract}}

\date{October 12, 2026}
\correspondence{Kaustav Kundu at \email{kkundu@meta.com} | Ritvik Shrivastava at \email{ritvik@meta.com}}
\metadata[Data \& Code]{\url{https://huggingface.co/datasets/facebook/wearable-ai}}

\definecolor{upgreen}{RGB}{0,128,0}

\providecommand{\probench}{Pro\textsuperscript{2}Bench}

\begin{document}

\maketitle

% \input{tex/intro} 
% ============================================================
% Introduction -- PWR (NeurIPS 2026)
% ============================================================

\section{Introduction}
\label{sec:intro}
\vspace{-5pt}

Procedural tasks such as servicing a bike or cooking a souffl\'{e} are ordered sequences of steps, and errors in ordering, technique, or timing have real consequences.
A proactive assistant on smart glasses must (i)~track procedural state from streaming egocentric video, (ii)~decide \emph{when} to intervene, and (iii)~decide \emph{what} to say, including how to recover when the user goes off plan.
It must do this without the user repeatedly asking \textit{``What is the next step?''}
These capabilities are coupled, since a wrong estimate of state produces mistimed interventions.

Recent work addresses these pieces separately, including online state and mistake detection~\citep{Seminara2024,Flaborea2024PREGO}, cognitively grounded timing~\citep{Li2024Satori,Zhang2024PrISM}, and streaming video LLMs at interactive latency~\citep{Qian2025Dispider,Wang2025StreamBridge}.
ProAssist~\citep{Zhang2025ProAssist} is the first end-to-end system for proactive assistance from streaming egocentric video.
Its central finding is that dialogue quality is limited by procedural reasoning rather than perception.
We ask how procedural reasoning should be evaluated and what a model needs in order to do it.

% (30K synthetic dialogues over six benchmarks on VideoLLM-Online)

% \begin{figure*}[htbp]
%     \centering
%     \includegraphics[width=\linewidth]{figures/procedural_teaser.png}
%     \caption{\textbf{PWR in action on an Out-of-Plan deviation.} The duplex interaction model continuously processes streaming egocentric video; when the user deviates from the procedural plan (e.g., wrong tool selected), it emits an \texttt{interrupt} at the deviation moment and the planner injects a recovery plan into the next forward pass. Baseline VLMs typically remain \texttt{silent} under plan conditioning (plan-induced silent bias) or continue the canonical procedure, while PWR produces a calibrated recovery instruction. See Appendix~\ref{app:wp-oop} for additional examples.}
%     \label{fig:teaser}
% \end{figure*}

% Despite these advancements, three main barriers block deployment of proactive assistant systems to real-world scenarios. \textbf{(B1)~No benchmark covers user deviations.} Existing egocentric corpora (\textit{e.g.}, HoloAssist, ProAssist)  assume canonical execution, while real users skip steps, reorder, substitute, make mistakes and recover (\emph{Out-of-Plan}, OOP); no public dataset pairs OOP with recovery guidance or carries per-decision-point intervention labels under wearable capture. 

\textbf{Evaluation lacks deviations and procedure-level scoring.}
Real users skip, reorder, and substitute steps, which we call out-of-plan (OOP) deviations.
Existing corpora either assume execution without deviations or provide recovery guidance only for cooking and for errors within a single step~\citep{Bhattacharyya2026EgoMC}.
Existing protocols also score each decision in isolation, which cannot tell whether a model carries a user through a procedure.

\textbf{Knowing the procedure is not the same as knowing where the user is.}
A model given the task or even the full list of steps must still infer from video which step is in progress and whether the user has left the expected path.
Neither prompting with a plan nor training on in-domain data closes this gap alone.
The model needs an explicit account of its position that stays current as the user acts and deviates.

\begin{figure*}[t!]
    \centering
    \includegraphics[width=\linewidth]{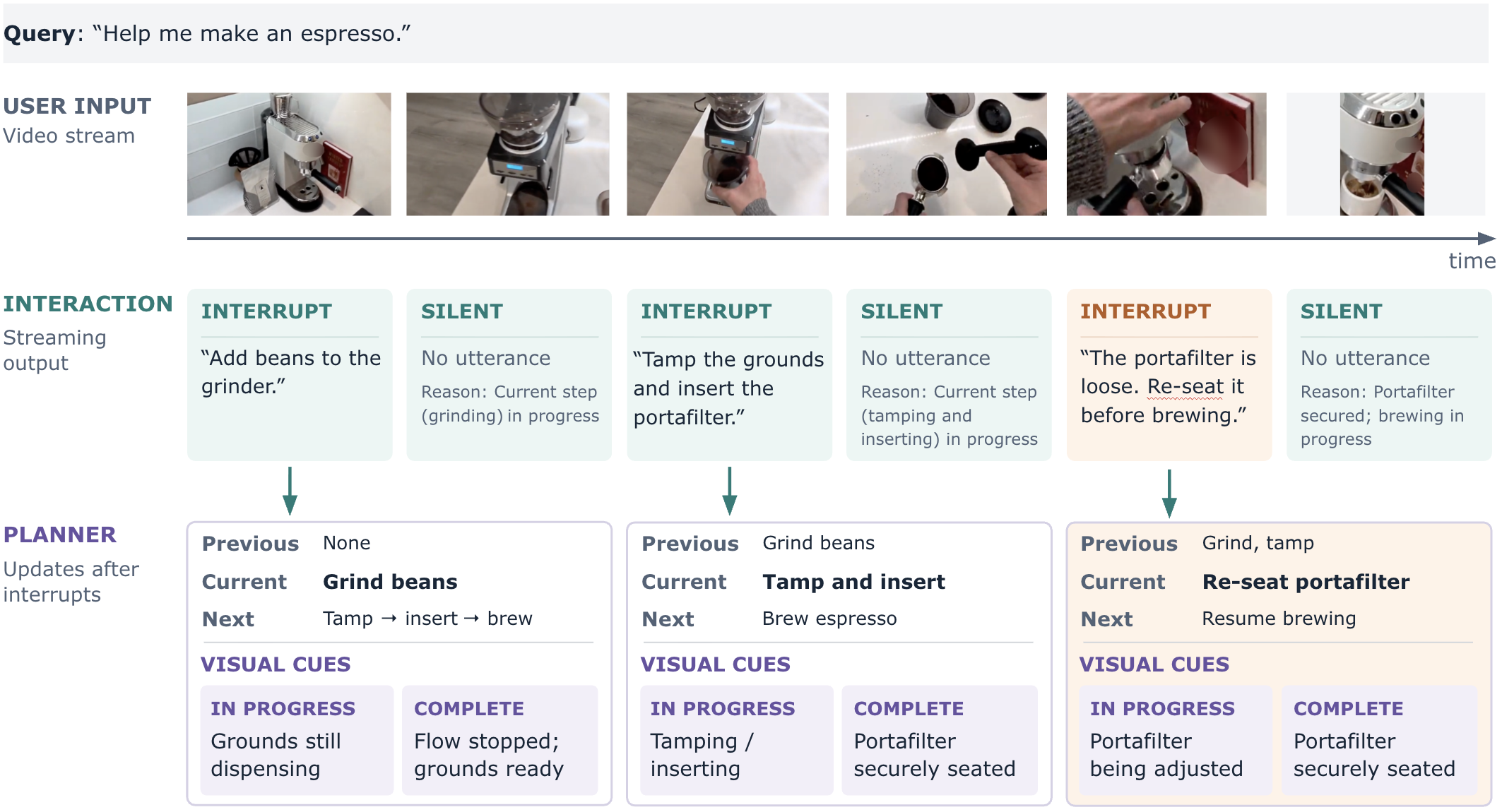}
    \caption{\textbf{Plan, Watch, Recover (PWR) on an out-of-plan deviation.}
    Given a request such as \textit{``Help me make an espresso,''} (a)~the \textbf{planner} writes the steps and the visual cues that mark each step's completion.
    (b)~The \textbf{interaction model} watches the stream against the current step and emits \textsc{Silent} or \textsc{Interrupt}.
    (c)~When the user deviates, the planner replaces the remaining steps with \textbf{recovery steps}.
    More examples are in App.~\ref{app:wp-oop}.}
    \label{fig:teaser}
\end{figure*}

\textbf{Our contributions.} 

\begin{enumerate}[leftmargin=1.4em,itemsep=1pt,topsep=2pt,parsep=0pt]
\item \textbf{EgoProactive, a multi-domain dataset of step-level deviations and their recoveries.}
We record $700$ sessions ($31.4$\,h) on Ray-Ban Meta glasses from about $30$ participants at home, across cooking, crafts, DIY, and tutorials.
In $247$ sessions the participant omits, reorders, or substitutes a step and then recovers within one continuous take, so one video grounds both detection and recovery.
To our knowledge, it is the only public dataset that pairs step-level deviations with recovery guidance across multiple domains (\S\ref{sec:egoproactive}).\footnote{\url{https://huggingface.co/datasets/facebook/wearable-ai}}
 
\item \textbf{\probench{}, a unified proactive procedural benchmark.}
We re-annotate Ego4D, Ego-Exo4D, EPIC-KITCHENS, HoloAssist, and HowTo100M with plan states, intervention decisions, and reference guidance under one schema, and combine them with EgoProactive.
\probench{} provides $26{,}874$ evaluation and $249{,}584$ training decision points across $14$ domains and $3$ viewpoints.
It scores systems per procedure and per turn (\S\ref{sec:dataset}).
 
\item \textbf{PWR, dynamic procedural context for proactive assistance.}
PWR places a plan state in the context of the interaction model and revises it after every intervention, so the model checks each observation against the current step and its visual cues.
A planner writes the revisions, including recovery steps after a deviation, and runs asynchronously so the interaction model keeps deciding in real time (\S\ref{sec:method}).
 
\item \textbf{Findings.}
Zero-shot, no model completes more than $26\%$ of a procedure's interventions before its first error.
Across three open-weight backbones, training on \probench{} without a plan barely changes performance, while adding the plan improves all nine backbone and metric comparisons, by $+.070$ Procedure Macro, $+.068$ Turn Macro-F1, and $+.176$ PQS on average.
The gains extend to held-out sources and on OOP deviations never seen in training.
\end{enumerate}

\section{Pro\textsuperscript{2}Bench: Proactive Procedural Benchmark}
\label{sec:dataset}

A proactive assistant must decide when to intervene and what to say throughout a multistep procedure.
When users omit, reorder, or substitute steps~\citep{Peddi2023,Sener2022}, it must also detect the deviation and guide them to a correct end state.
Evaluating an assistant for deployment therefore requires intervention decisions with reference guidance, and deviation detection with recovery guidance.
No existing dataset provides both across diverse domains and capture settings.
Large egocentric and instructional corpora such as Ego4D~\citep{Grauman2022}, EPIC-KITCHENS~\citep{Damen2022EPIC100}, Ego-Exo4D~\citep{Grauman2024EgoExo4D}, and HowTo100M~\citep{Miech2019HowTo100M} contain neither.
ProAssist~\citep{Zhang2025ProAssist} covers only procedures executed without deviations, CaptainCook4D~\citep{Peddi2023} lacks guidance, and HoloAssist~\citep{Wang2023} does not mark boundaries between steps.

QICD~\citep{Bhattacharyya2025QICD} and Ego-MC-Bench~\citep{Bhattacharyya2026EgoMC} provide recovery guidance, but only for cooking and for deviations within a single step, such as a wrong quantity.
We close this gap with two contributions: we re-annotate five public corpora with intervention decisions and reference guidance under one schema, and we collect EgoProactive, a new smart-glasses dataset with step-level deviations and their recoveries.
Together they form \probench{}, with $26{,}874$ evaluation and $249{,}584$ training decision points from $29{,}367$ videos across $14$ domains and three viewpoints (Table~\ref{tab:benchmark}).

% ------------------------------------------------------------
\subsection{Task and Annotation Schema}
\label{sec:task}
 
\paragraph{Task.}
A procedure is one user carrying out one task, beginning with a query $q$ in which the user states the task.
At time $t$ the assistant receives the observation $o_t = (v_{1:t}, \mathcal{H}_{<t}, q)$, where $v_{1:t}$ is the video stream sampled at $2$\,fps and $\mathcal{H}_{<t}$ is the dialogue history.
It emits an intervention decision $d_t \in \{\textsc{Silent}, \textsc{Interrupt}\}$ and, when $d_t = \textsc{Interrupt}$, an utterance $u_t$.
An interrupt is warranted when the user completes a step, or departs from the procedure at the level of steps, which we call an out-of-plan (OOP) deviation.

\paragraph{Decision points.}
Each procedure is annotated as a time-ordered sequence of decision points.
A decision point is a time $t$ with a reference decision $d_t^\star$, the $8$\,s clip ending at $t$, and the reference dialogue history $\mathcal{H}^\star_{<t}$.
Interrupt points lie at step boundaries and OOP onsets and carry a reference utterance $u_t^\star$.
Silent points are placed between them.
A silent point has no interrupt point within its clip or the following $\tau = 2$\,s, which excludes frames just before a boundary where the correct decision is ambiguous (\S\ref{section:data_annotation_pipeline}).
% TODO(code): confirm the sampler implements this rule and tau = 2 s; regenerate eval silents if not.
 
\paragraph{Plan states.}
Each decision point also carries a reference plan state $P_t^\star = \bigl((s_1,\sigma_1),\ldots,(s_N,\sigma_N);\,\mathbf{c}\bigr)$.
Here $s_i$ describes the $i$-th of $N$ steps, $\sigma_i \in \{\texttt{completed},\texttt{current},\texttt{next}\}$ is its status at $t$, and the cue set $\mathbf{c}$ describes visual evidence that the \texttt{current} step is in progress or has completed.
After an OOP deviation the step list includes the annotated recovery steps.
Reference plan states supervise training (\S\ref{sec:train-infer}) and are never given to evaluated systems.

% ------------------------------------------------------------
\subsection{EgoProactive}
\label{sec:egoproactive}
\label{sec:sources}
 
EgoProactive records step-level deviations and their recoveries in continuous egocentric video from consumer smart glasses.
It contains $700$ sessions totaling $31.4$ hours from about $30$ participants, who wore Ray-Ban Meta glasses in their own homes for $2$ to $5$ minutes per session.
Each session follows a script with a goal and $5$ to $14$ ordered steps ($9.3$ on average), drawn from $700$ distinct scripts in cooking, crafts, DIY, and tutorials.
In the $247$ OOP sessions, the participant also commits a scripted deviation and fixes it within one continuous take that ends in the correct state, so one video grounds both detection and recovery.
The deviations consist of omissions, reorderings, and substitutions.
The remaining $453$ sessions follow the script without deviation and test step-completion guidance on a device and in settings absent from training.
Against Ego-MC-Bench ($40$ sessions, $7$ participants), EgoProactive has about $17$ times as many sessions and targets deviations across steps rather than within one, so the two are complementary.

EgoProactive carries the OOP evaluation, hence annotators reviewed every video in full:  corrected steps and boundaries, marked each deviation onset by hand, linked to corresponding recovery steps, wrote the reference utterances for OOP decision points, and edited the pipeline's candidates (\S\ref{section:data_annotation_pipeline}) for step completions (protocol, demographics, and ethics in App.~\ref{app:wp}).

\subsection{Unified Annotation}
\label{section:data_annotation_pipeline}
 
Ego4D~\citep{Grauman2022}, EPIC-KITCHENS~\citep{Damen2022EPIC100}, Ego-Exo4D~\citep{Grauman2024EgoExo4D}, HoloAssist~\citep{Wang2023}, and HowTo100M~\citep{Miech2019HowTo100M} supply viewpoint, device, and domain diversity (Table~\ref{tab:benchmark}).
Their annotations, from action segments to transcripts, describe what the user does, not when to intervene or what to say.
One pipeline maps these five corpora and EgoProactive onto the schema of \S\ref{sec:task}.
Source-specific converters recover the goal and steps where available, and a VLM proposes the remaining fields as candidates.
This is followed by an iterative refinement stage, where a VLM judge checks each candidate's steps, boundaries, decisions, and guidance against the frames.
For each failure, the judge returns localized frame evidence and identifies the affected fields.
A separate correction pass receives the original candidate, the field-level critique, and the same frames; it revises only the flagged step text, temporal boundary, decision label, or guidance utterance while preserving accepted fields.
The revised candidate is judged again, and any item that remains inconsistent is quarantined rather than released.

\begin{table*}[t]
\centering
\setlength{\tabcolsep}{2pt}
\renewcommand{\arraystretch}{1.10}
\scriptsize
\begin{tabular*}{\textwidth}{@{\extracolsep{\fill}}l l l r r r r r@{}}
\toprule
\textbf{Source} & \textbf{View} & \textbf{Source Annot.}
  & \textbf{Doms} & \textbf{Vids}
  & \textbf{Train DPs} & \textbf{Eval DPs} & \textbf{OOP DPs} \\
\midrule
Ego4D$^\ddagger$
  & Ego (head-cam) & Narrations & 14 & 24{,}474 & 214{,}511 & -- & 0 \\
EgoExo4D (EE4D)
  & Ego + Exo & Keysteps & 8 & 462 & 13{,}726 & 2{,}734 & 0 \\
EPIC-KITCHENS (EK)
  & Ego (head-cam) & Actions & 1 & 468 & 8{,}954 & 6{,}399 & 0 \\
HowTo100M (HowTo)
  & 3rd-person & Transcripts & 12 & 3{,}026 & 12{,}393 & 4{,}654 & 0 \\
HoloAssist (HA)$^\dagger$
  & Ego (AR) & Dialogue & 4 & 237 & -- & 3{,}188 & 0 \\
\midrule
\rowcolor{blue!8}
\textbf{EgoProactive (ours)}$^\dagger$
  & Ego (glasses) & Scripts + OOP & 4 & 700 & -- & 9{,}899 & 3{,}810 \\
\midrule
\textbf{\probench{} (ours)}
  & All 3 views & Unified schema
  & \textbf{14} & \textbf{29{,}367} & \textbf{249{,}584}
  & \textbf{26{,}874} & \textbf{3{,}810} \\
\bottomrule
\end{tabular*}
\vspace{-2mm}
\caption{\small \textbf{\probench{} composition.}
\textbf{View} is the viewpoint of the procedure.
\textbf{Source Annot.} is the annotation each source provided before conversion to the \probench{} schema.
\textbf{DPs} are decision points, counted after splitting, filtering, deduplication, and leakage checks.
\textbf{OOP DPs} come from the $247$ EgoProactive sessions with a deviation and its recovery.
The last row counts unique domains and videos.
$^\dagger$Evaluation only. $^\ddagger$Training only.}
\label{tab:benchmark}
\end{table*}

\subsection{Benchmark Metrics}
\label{sec:tasks-metrics}
\paragraph{Procedure-level metrics.}
A system runs along all decision points of a procedure in time order until its
first failure, a decision point where the predicted decision $\hat d_t$ differs
from $d_t^\star$.
Let $\mathcal{P}$ be the set of evaluated procedures.
For procedure $i$, let $L_i$ be the number of reference interrupt points and
$k_i$ the number of those interrupts correctly reached before the first
failure. Correct silent decisions extend the valid sequence but do not
increment $k_i$. We report $\mathrm{Procedure\ Macro} = 1/|\mathcal{P}|\sum_{i\in\mathcal{P}} k_i/L_i$.
% \[
%   \mathrm{Procedure\ Macro}
%   = \frac{1}{|\mathcal{P}|}\sum_{i\in\mathcal{P}}\frac{k_i}{L_i}.
% \]
Each decision point uses the reference dialogue history
$\mathcal{H}^\star_{<t}$. Errors compound because evaluation stops at the
first incorrect decision: no later interrupt contributes to $k_i$, even if it is predicted correctly.
% Stopping at the first failure reflects a user who waits to be told a step is complete, unlike protocols that stream past a miss~\citep{Bhattacharyya2026EgoMC}.

\paragraph{Turn-level metrics.}
Turn-level metrics use all 26{,}874 frozen decision points, without class
rebalancing or metric-specific subsampling (App.~\ref{app:eval-population}).
We report interrupt F1 (IF1), silent F1 (SF1), and
$\mathrm{Turn\ Macro\text{-}F1}=(\mathrm{IF1}+\mathrm{SF1})/2$.
An invalid or unparseable output is scored as an incorrect decision.

\paragraph{Proactive Quality Score (PQS).}
\label{score:pqs}
For each correct interrupt, GPT-5.6 scores the utterance $\hat u_t$ against $u_t^\star$ from $1$ to $5$ on relevance, specificity, actionability, and conciseness.
With $\bar r_t$ their mean, the normalized quality is $g_t=(\bar r_t-1)/4\in[0,1]$, and exact matches receive $g_t = 1$.
Let $\mathcal{T}$ be the set of evaluation decision points and $s_t$ equal $1$ for a correct silence, $g_t$ for a correct interrupt, and $0$ otherwise.
Then $\mathrm{PQS}=\frac{1}{|\mathcal{T}|}\sum_{t\in\mathcal{T}}s_t$, which rewards speaking at the right time and saying something useful.
Utterance quality alone is reported in App.~\ref{app:results-subj-patterns}.

\section{Method}
\label{sec:method}

A proactive assistant must know where the user is in the procedure.
Knowing the steps is not enough.
The assistant must also infer from video which step is in progress, whether it has just finished, and whether the user has left the expected path.
We make this state explicit and keep it current.
We introduce \textbf{PWR} (Plan, Watch, Recover), which places a plan state in the context of the interaction model and revises it as the procedure unfolds.
A planner writes the initial plan and revises it after each intervention (\emph{Plan}).
The interaction model checks each observation against the current step and decides whether to speak (\emph{Watch}).
After an OOP deviation, the planner replaces the remaining steps with recovery steps (\emph{Recover}).
% With reference plans, which remove planner errors from the comparison, dynamic context improves OOP procedure completion over static context for every backbone (\S\ref{sec:results}).
Both models are post-trained on plan-conditioned trajectories from \probench{} (\S\ref{sec:train-infer}). We adopt the task of \S\ref{sec:task}.
In practice $o_t$ contains the $8$\,s clip ending at $t$ rather than the full stream $v_{1:t}$.
Let $\mathcal{V}$ denote model vocabulary, so $u_t \in \mathcal{V}^{*}$ when $d_t = \textsc{Interrupt}$ and $u_t = \emptyset$ otherwise.
We write $P_t$ for a predicted plan state with the structure of \S\ref{sec:task}, serialized as a structured text block in the input of the interaction model (Figure~\ref{fig:teaser}).

% ------------------------------------------------------------
\subsection{From Implicit to Dynamic Procedural Context}
\label{sec:context}

We consider three ways of giving procedural knowledge to an interaction model $p_{\theta}$, differing only in input context.
The backbone, video clips (App.~\ref{sec:clip-selection}), and training data are held fixed.
 
\paragraph{Implicit and Static context.}
The two baselines condition on
\par\vspace{\abovedisplayskip}
\noindent
\makebox[0.5\linewidth][c]{%
  $\displaystyle p_{\theta}\bigl(d_t,\, u_t \mid o_t\bigr)$%
  \quad\refstepcounter{equation}\label{eq:implicit}(\theequation)}%
\makebox[0.5\linewidth][c]{%
  $\displaystyle p_{\theta}\bigl(d_t,\, u_t \mid o_t,\, P_0\bigr)$%
  \quad\refstepcounter{equation}\label{eq:static}(\theequation)}%
\par\vspace{\belowdisplayskip}
\noindent
Implicit context (Eq.~\ref{eq:implicit}) is the formulation of \citet{Zhang2025ProAssist}.
The procedure exists only in $q$ and $\mathcal{H}_{<t}$, and one model must recover the steps, track the current one, and decide whether to speak.
Static context (Eq.~\ref{eq:static}) adds a plan $P_0$ generated once from $q$.
It states what the task involves.
The model must still infer its position from the video and from steps it already confirmed in $\mathcal{H}_{<t}$.
$P_0$ cannot change when the user departs from it.

\paragraph{Dynamic context (ours).}
PWR conditions on the most recent plan state (Eq.~\ref{eq:dynamic}) and revises it as the procedure unfolds (Eq.~\ref{eq:update-rule}),
\begin{subequations}
\label{eq:dynamic-context}
\par\vspace{\abovedisplayskip}
\noindent
\makebox[0.5\linewidth][c]{%
  $\displaystyle p_{\theta}\bigl(d_t,\, u_t \mid o_t,\, P_{t-1}\bigr)$%
  \quad\refstepcounter{equation}\label{eq:dynamic}(\theequation)}%
\makebox[0.5\linewidth][c]{%
  $\displaystyle P_t = \mathcal{U}\bigl(P_{t-1},\, d_t,\, u_t,\, o_t\bigr)$%
  \quad\refstepcounter{equation}\label{eq:update-rule}(\theequation)}%
\par\vspace{\belowdisplayskip}
\end{subequations}
\noindent
where $\mathcal{U}$ is an update rule realized by a learned planner (\S\ref{sec:update}).
Static context is the special case $\mathcal{U}(P_{t-1}, \cdot) = P_{t-1}$, so dynamic context subsumes it.
With a current $P_{t-1}$ the interaction model no longer has to localize the user in the task.
It checks the observation against the \texttt{current} step and its cues.
A step completion is a match with the completion cues, and an OOP deviation is a mismatch with the current step.
 
Static and dynamic contexts differ most when the plan itself must change.
On procedures that follow the script, the dialogue history already records which steps the assistant has confirmed, and static and dynamic context perform similarly (App.~\ref{app:eval}).
After an OOP deviation, only a dynamic plan can replace the remaining steps with recovery steps.
With reference plans, which remove planner errors from the comparison, dynamic context improves OOP procedure completion over static context for every backbone (\S\ref{sec:results}).
After an OOP deviation, only a dynamic plan can replace the remaining steps with recovery steps.
Dynamic context thus handles step completion and recovery with one mechanism.

\subsection{Updating the Plan}
\label{sec:update}

\begin{figure}[t]
    \centering
    \includegraphics[width=\linewidth]{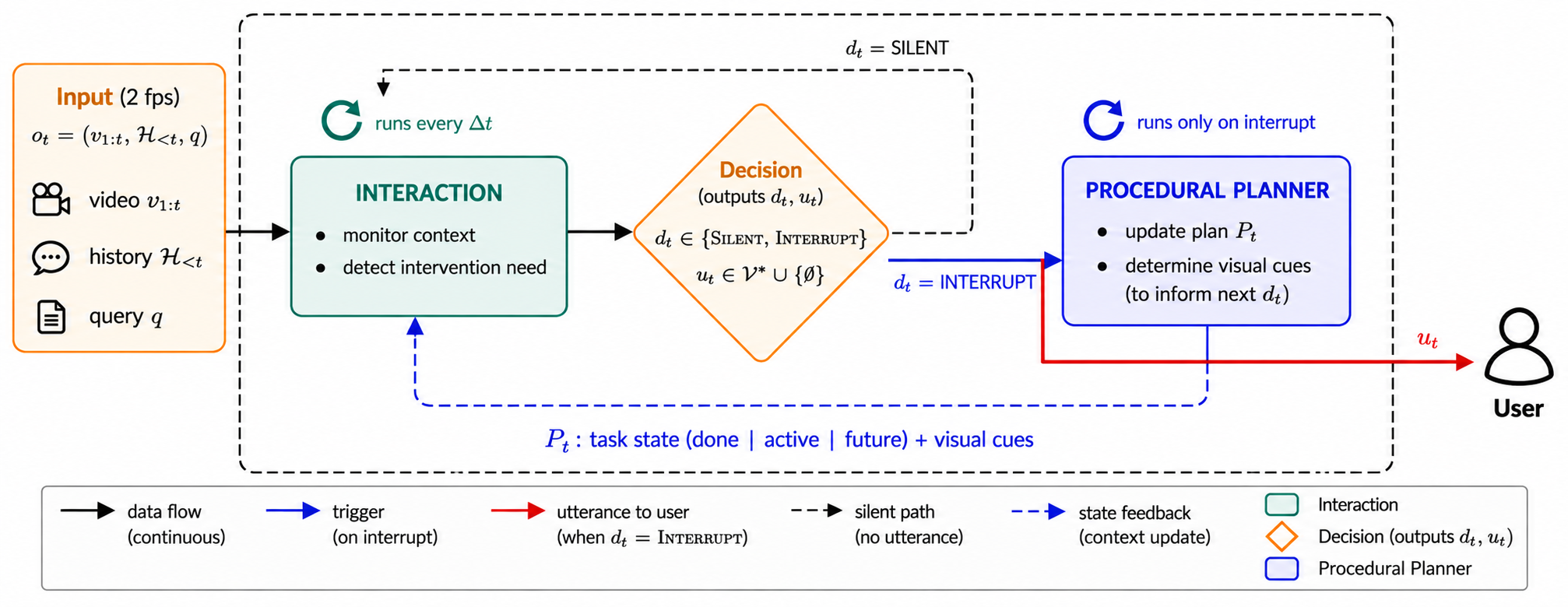}
    % \vspace{-6mm}
    \caption{\textbf{PWR.}
    At observation $o_t$, the interaction model conditions on the latest plan state $P_{t-1}$ to emit decision $d_t$ and, when interrupting, utterance $u_t$.
    An interrupt triggers an asynchronous planner call.
    The revised state $P_t$ replaces $P_{t-1}$ for later decisions once it passes schema validation.}
    \label{fig:architecture}
    \vspace{-3mm}
\end{figure}

The plan changes only when a step completes or the user deviates.
These are the moments at which the interaction model interrupts, so we revise the plan only at interrupts.
We model $\mathcal{U}$ with a planner $p_{\phi}$.
Letting $p$ denote the joint distribution over the outputs of the system at time $t$,
\begin{equation}
\boxed{
    p\bigl(d_t, u_t, P_t \mid o_t, P_{t-1}\bigr)
    =
    \underbrace{p_{\theta}\bigl(d_t, u_t \mid o_t, P_{t-1}\bigr)}_{\text{Interaction model}}
    \cdot
    \underbrace{p_{\phi}\bigl(P_t \mid d_t, u_t, o_t, P_{t-1}\bigr)}_{\text{Planner}}
}
\label{eq:factorization}
\end{equation}
where the first factor is Eq.~\ref{eq:dynamic} and the second is a learned form of $\mathcal{U}$ with
\begin{equation}
    p_{\phi}\bigl(P_t \mid d_t, u_t, o_t, P_{t-1}\bigr) =
    \begin{cases}
        \delta\bigl(P_t - P_{t-1}\bigr) & d_t = \textsc{Silent} \\
        \text{learned (\S\ref{sec:train-infer})} & d_t = \textsc{Interrupt}
    \end{cases}
    \label{eq:planner-cases}
\end{equation}
and $\delta$ the Dirac delta.
Each procedure opens with the response of the assistant to $q$, treated as an interrupt at $t = 0$ with $P_{-1} = \emptyset$.
On a step completion, the planner advances the \texttt{current} status and writes new cues.
On an OOP deviation it writes a recovery plan.
Conditioning on $u_t$ keeps the revision consistent with what the user was just told.
Because $P_{t-1}$ already encodes the procedural history, the planner sees only the most recent clip.
 
We run the planner asynchronously so that the interaction model continues processing observations while emitting a \texttt{silent} token or a short instruction every $500$\,ms.
A plan averages $114.9$ tokens against $12.3$ for a short instruction.
On our hardware the plan output takes $0.9$ to $1.6$\,s to decode, or two to three observation intervals. % (App.~\ref{sec:latency}).
A single model that also revised its plan would stop observing for this time after every interrupt, precisely while the user acts on the guidance it just gave.
Faster hardware, serving optimizations, and shorter plans reduce this cost, but they reduce it equally for PWR and are orthogonal to the design.
Whatever a revision costs, a single model pays it inside the decision loop, and PWR pays it outside.
The interaction model never waits on the planner, and the expected cost per timestep is $C_{\theta} + \Pr(\textsc{Interrupt})\, C_{\phi}$, where $C_{\theta}$ and $C_{\phi}$ are the per-call costs of the interaction model and the planner.

% ------------------------------------------------------------
\subsection{Training and Inference}
\label{sec:train-infer}

A plan in context is necessary but not sufficient.
Zero-shot models given a plan gain little (App.~\ref{app:eval}), so we post-train.
The planner and the interaction model are trained with the same supervised recipe on the same \probench{} decision points.
Each decision point pairs the reference decision $d_t^\star$ and utterance $u_t^\star$ with the latest reference plan state $P^\star_{t-1}$.
The interaction model learns $p_{\theta}(d_t^\star, u_t^\star \mid o_t, P^\star_{t-1})$ at every decision point.
The planner learns $p_{\phi}(P^\star_t \mid d_t^\star, u_t^\star, o_t, P^\star_{t-1})$ at interrupt points.
A silent point teaches the interaction model to stay silent given a specific current step and its cues.
An interrupt point teaches it to speak because the observation matched or broke that expectation.
At inference, each interrupt triggers an asynchronous planner call, and user prompts and OOP recovery need no dedicated machinery.
Objectives, sampling, and inference details are in App.~\ref{app:method}.

% ArXiv layout of results3_iclr.tex: same prose and results, tables placed beside their discussion.

\section{Experiments and Results}
\label{sec:experiments}
\label{sec:results}

\paragraph{Setup.}

We evaluate on the \probench{} evaluation split (Table~\ref{tab:benchmark}) with the metrics of \S\ref{sec:tasks-metrics}.
Each decision point
contains up to 16 chronological frames from an 8\,s window sampled at 2\,fps
and the dialogue history.
PWR carries its own predicted plan state forward and calls the planner only after a predicted \textsc{Interrupt}, as in Eq.~\ref{eq:factorization}.
We report Turn Macro-F1,

Procedure Macro derived from the same turn predictions, and PQS for the joint
quality of when to speak and what to say (\S\ref{sec:tasks-metrics}). We
compare GPT-5.6, Gemini 3.6 Flash, and Claude Opus 5 with the open-weight Kimi
K3, Inkling, Gemma~4 31B, Muse Glimmer 30B, and Qwen3.8-VL-27B. Annotation tags, targets, and future mistake or recovery information are
removed from every model input. Training inputs, controls and prompts are in App. ~\ref{app:eval}.

% \paragraph{Setup.}
% We evaluate on the 40{,}008-instance Pro\textsuperscript{2}Bench (Table~\ref{tab:benchmark}) and score each decision point on G-Mean F1 for the timing decision and PQS for the joint quality of when to speak and what to say (\S\ref{sec:tasks-metrics}).
% We compare three frontier proprietary models (Gemini 3.1 Pro, Claude Opus 4.6, GPT-5.2) and three open-weight models (Qwen3-VL-235B, Qwen3.6-VL-27B, Llama~4-Maverick), all zero-shot, with GPT-5.2 also serving as the PQS judge and excluded as a candidate.
% Our PWR systems fine-tune two open-weight backbones, Llama~4 17B×128 MoE (Maverick) and Qwen3.6-VL-27B, on the four-dataset subset \{EE4D, EK, HowTo, Ego4D\}, leaving HoloAssist and EgoProactive held out to measure generalization.
% We also report PWR-Oracle, the trained interaction paired with a ground-truth plan, as a non-deployable upper bound. Model configurations, the three planning conditions, prompts, and training details are in Appendix~\ref{app:eval-models}.
\paragraph{Training Recipe.}
For Gemma, Muse, and Qwen we train the planner and the interaction model with LoRA on EgoExo4D, EPIC-KITCHENS, HowTo100M, and Ego4D, using the recipe of \S\ref{sec:train-infer}.
HoloAssist and EgoProactive are held out, so no OOP session is seen in training.
We sample sources in proportion to the square root of their size and select checkpoints on a validation split whose procedures do not overlap training.
The \emph{FT, no plan} control trains the interaction model on the same data with the plan removed from its context, so it differs from PWR only in the input context.
Details are in App.~\ref{app:training-recipe}.

\subsection{Main Findings}
\label{sec:main-findings}

\begin{table}[H]
\centering\scriptsize
\setlength{\tabcolsep}{1.2pt}
\renewcommand{\arraystretch}{1.08}
\begin{tabular*}{\textwidth}{@{\extracolsep{\fill}}l l ccccc|ccc@{}}
\toprule
& & \multicolumn{5}{c|}{\emph{Zero-shot only}} & \multicolumn{3}{c}{\emph{Trained}} \\
\cmidrule(lr){3-7}\cmidrule(lr){8-10}
\textbf{Metric} & \textbf{Config} & \textbf{GPT} & \textbf{Gem.} & \textbf{Opus} & \textbf{Kimi} & \textbf{Ink.} & \textbf{Gemma} & \textbf{Muse} & \textbf{Qwen} \\
\midrule
Procedure Macro & Zero-shot & .068 & .058 & .186 & .096 & .249 & .216 & .078 & .262 \\
 & FT, no plan & -- & -- & -- & -- & -- & .280 & .080 & .248 \\
 & PWR & -- & -- & -- & -- & -- & \cellcolor{blue!8}\textbf{.325} & \cellcolor{blue!8}\textbf{.213} & \cellcolor{blue!8}\textbf{.279} \\
 & \cellcolor{gray!15}$\Delta_{\text{plan}}$ & \cellcolor{gray!15} & \cellcolor{gray!15} & \cellcolor{gray!15} & \cellcolor{gray!15} & \cellcolor{gray!15} & \cellcolor{gray!15}\textcolor{upgreen}{$\uparrow$.045} & \cellcolor{gray!15}\textcolor{upgreen}{$\uparrow$.133} & \cellcolor{gray!15}\textcolor{upgreen}{$\uparrow$.031} \\
\midrule
Turn Macro-F1 & Zero-shot & .490 & .461 & .567 & .465 & .532 & .539 & .432 & .463 \\
 & FT, no plan & -- & -- & -- & -- & -- & .555 & .425 & .452 \\
 & PWR & -- & -- & -- & -- & -- & \cellcolor{blue!8}\textbf{.592} & \cellcolor{blue!8}\textbf{.501} & \cellcolor{blue!8}\textbf{.542} \\
 & \cellcolor{gray!15}$\Delta_{\text{plan}}$ & \cellcolor{gray!15} & \cellcolor{gray!15} & \cellcolor{gray!15} & \cellcolor{gray!15} & \cellcolor{gray!15} & \cellcolor{gray!15}\textcolor{upgreen}{$\uparrow$.037} & \cellcolor{gray!15}\textcolor{upgreen}{$\uparrow$.076} & \cellcolor{gray!15}\textcolor{upgreen}{$\uparrow$.090} \\
\midrule
PQS & Zero-shot & .416 & .398 & .461 & .408 & .412 & .430 & .280 & .392 \\
 & FT, no plan & -- & -- & -- & -- & -- & .400 & .358 & .412 \\
 & PWR & -- & -- & -- & -- & -- & \cellcolor{blue!8}\textbf{.625} & \cellcolor{blue!8}\textbf{.468} & \cellcolor{blue!8}\textbf{.605} \\
 & \cellcolor{gray!15}$\Delta_{\text{plan}}$ & \cellcolor{gray!15} & \cellcolor{gray!15} & \cellcolor{gray!15} & \cellcolor{gray!15} & \cellcolor{gray!15} & \cellcolor{gray!15}\textcolor{upgreen}{$\uparrow$.225} & \cellcolor{gray!15}\textcolor{upgreen}{$\uparrow$.110} & \cellcolor{gray!15}\textcolor{upgreen}{$\uparrow$.193} \\
\bottomrule
\end{tabular*}
\vspace{-3mm}
\caption{\textbf{Results on \probench{}.}
Each metric is the equal-weight mean over the five evaluation datasets, with EgoProactive OOP and non-OOP sessions pooled as one source.
\emph{FT, no plan} trains the interaction model on the same data as PWR without a plan in context.
PWR adds the plan and the learned planner.
$\Delta_{\text{plan}}$ is PWR minus \emph{FT, no plan}, the gain from the plan alone.
\textbf{Bold} is the best configuration for each trained backbone.
\colorbox{blue!8}{Ours}.}
\label{tab:main}
\end{table}

\textbf{\probench{} is far from solved, and procedure-level scoring shows why.}
Zero-shot, no model completes more than $26\%$ of reference interventions before its first error, and GPT-5.6 and Gemini complete under $7\%$ (Table~\ref{tab:main}).
Turn Macro-F1 ranges from $.461$ to $.567$.
Turn level and procedure level metrics also rank the models differently. 
Qwen has the highest zero-shot Procedure Macro ($.262$) but amongst the lowest Turn Macro-F1 ($.463$), while Claude Opus 5 leads on turns ($.567$) and trails on procedures ($.186$).
Deciding well on isolated turns does not mean a model can carry a user through a procedure.

\textbf{The gains come from the plan, not from the training data.}
PWR improves over its zero-shot backbone on all three metrics for Gemma, Muse, and Qwen (Table~\ref{tab:main}).
Training the same interaction model on the same data without a plan recovers little of this gain.
Averaged over the three backbones, it changes Procedure Macro by $+.017$, Turn Macro-F1 by $-.001$, and PQS by $+.023$, and it lowers at least one metric for every backbone.
Adding the plan then improves all nine comparisons, by $+.070$ Procedure Macro, $+.068$ Turn Macro-F1, and $+.176$ PQS on average.
Gemma is the one backbone where data alone accounts for most of the procedure-level gain ($.216$ to $.280$ of $.325$).
Because zero-shot models given a plan also gain little (App.~\ref{app:eval}), neither the plan nor the training data suffices alone.
The plan helps once the model is trained to read it.
% Qwen gains least on procedures ($+.017$), which App.~\ref{app:eval} traces to \tbd.
% \textbf{PWR delivers consistent gains across interaction metrics.}
% For Gemma, Muse, and Qwen, PWR improves over the corresponding no-plan model in
% all nine aggregate comparisons across Turn Macro-F1, Procedure Macro, and PQS
% (Table~\ref{tab:main}). Averaged equally across the three families, PWR raises
% Turn Macro-F1 from .478 to .545, Procedure Macro from .185 to .272, and PQS from
% .367 to .566.
Figure~\ref{fig:combined-radar} shows the corresponding
model-wise profiles, including subjective response quality. The repeated gains across distinct backbones support the joint
benefit of learned procedural state and an interaction model trained to consume it.

% \textbf{The strongest overall results come from PWR.}
% Gemma PWR achieves the highest aggregate Turn Macro-F1 (.592), Procedure Macro
% (.325), and PQS (.625). Qwen PWR obtains the second-highest Procedure Macro
% (.279) and PQS (.605). Muse also improves substantially, from .432 to .501 in
% Turn Macro-F1, .078 to .213 in Procedure Macro, and .280 to .468 in PQS.
% \textbf{The gains transfer across backbones and held-out sources.}
% On both held-out sources, HoloAssist and EgoProactive, PWR improves Turn
% Macro-F1 and PQS over the corresponding no-plan configuration for all three
% trained model families (Tables~\ref{tab:detail-holoassist} and
% \ref{tab:detail-egoproactive}). On EgoProactive, PWR reaches .598, .509, and
% .563 Turn Macro-F1 for Gemma, Muse, and Qwen, respectively, while PQS reaches
% .582, .533, and .594. The consistent improvements across three backbones and
% two unseen sources demonstrate that the learned system generalizes beyond its
% training mixture.

\textbf{The gains hold on held-out sources and on out-of-plan deviations.}
On HoloAssist and EgoProactive, neither seen in training, PWR improves Turn Macro-F1 and PQS for all three backbones (Tables~\ref{tab:detail-holoassist} and~\ref{tab:detail-egoproactive}).
The OOP sessions of EgoProactive are the hardest test, since every deviation is scripted but none appears in training (Table~\ref{tab:oop-results}).
PWR raises OOP Procedure Macro for every backbone, most for Qwen ($.038$ to $.291$), and raises OOP PQS by $.094$ to $.250$.
This is also where revising the plan matters.
Replacing the dynamic plan with a static one lowers OOP Procedure Macro for every backbone, by $2$ to $8$ points relatively.

\begin{table}[H]
\centering
\begin{minipage}{0.72\textwidth}
\centering\scriptsize
\setlength{\tabcolsep}{1.5pt}
\renewcommand{\arraystretch}{1.08}
\begin{tabular*}{\linewidth}{@{\extracolsep{\fill}}l l ccc@{}}
\toprule
& & \multicolumn{3}{c}{\emph{Trained}} \\
\cmidrule(lr){3-5}
\textbf{Metric} & \textbf{Config} & \textbf{Gemma} & \textbf{Muse} & \textbf{Qwen} \\
\midrule
Procedure Macro & Zero-shot & .167 & .024 & .038 \\
 & FT, no plan & .185 & .038 & .194 \\
 & PWR & \cellcolor{blue!8}\textbf{.245} & \cellcolor{blue!8}\textbf{.134} & \cellcolor{blue!8}\textbf{.291} \\
 & \cellcolor{gray!15}$\Delta_{\text{plan}}$ & \cellcolor{gray!15}\textcolor{upgreen}{$\uparrow$.060} & \cellcolor{gray!15}\textcolor{upgreen}{$\uparrow$.096} & \cellcolor{gray!15}\textcolor{upgreen}{$\uparrow$.097} \\
\midrule
Turn Macro-F1 & Zero-shot & .579 & .451 & .434 \\
 & FT, no plan & .524 & .406 & .498 \\
 & PWR & \cellcolor{blue!8}\textbf{.582} & \cellcolor{blue!8}\textbf{.484} & \cellcolor{blue!8}\textbf{.553} \\
 & \cellcolor{gray!15}$\Delta_{\text{plan}}$ & \cellcolor{gray!15}\textcolor{upgreen}{$\uparrow$.058} & \cellcolor{gray!15}\textcolor{upgreen}{$\uparrow$.078} & \cellcolor{gray!15}\textcolor{upgreen}{$\uparrow$.055} \\
\midrule
PQS & Zero-shot & .457 & .254 & .338 \\
 & FT, no plan & .433 & .433 & .251 \\
 & PWR & \cellcolor{blue!8}\textbf{.551} & \cellcolor{blue!8}\textbf{.504} & \cellcolor{blue!8}\textbf{.566} \\
 & \cellcolor{gray!15}$\Delta_{\text{plan}}$ & \cellcolor{gray!15}\textcolor{upgreen}{$\uparrow$.118} & \cellcolor{gray!15}\textcolor{upgreen}{$\uparrow$.071} & \cellcolor{gray!15}\textcolor{upgreen}{$\uparrow$.315} \\
\bottomrule
\end{tabular*}
\vspace{-3mm}
\captionsetup{font=footnotesize}
\captionof{table}{\textbf{Out-of-plan recovery on EgoProactive.} Each metric is computed  on the EgoProactive-OOP procedures. \emph{FT, no plan} trains the interaction model on the same data as PWR without a plan in context. PWR adds the plan and learned planner. $\Delta_{\text{plan}}$ is PWR minus \emph{FT, no plan}. \textbf{Bold}: best config for each trained model. \colorbox{blue!8}{Ours}.}
\label{tab:oop-results}
\end{minipage}
% \par\vspace{4pt}

\end{table}
\newcommand{\PWRRadarFigure}{%
\begin{figure}[H]
\centering
\begin{minipage}{0.96\textwidth}
\centering
\includegraphics[width=\linewidth]{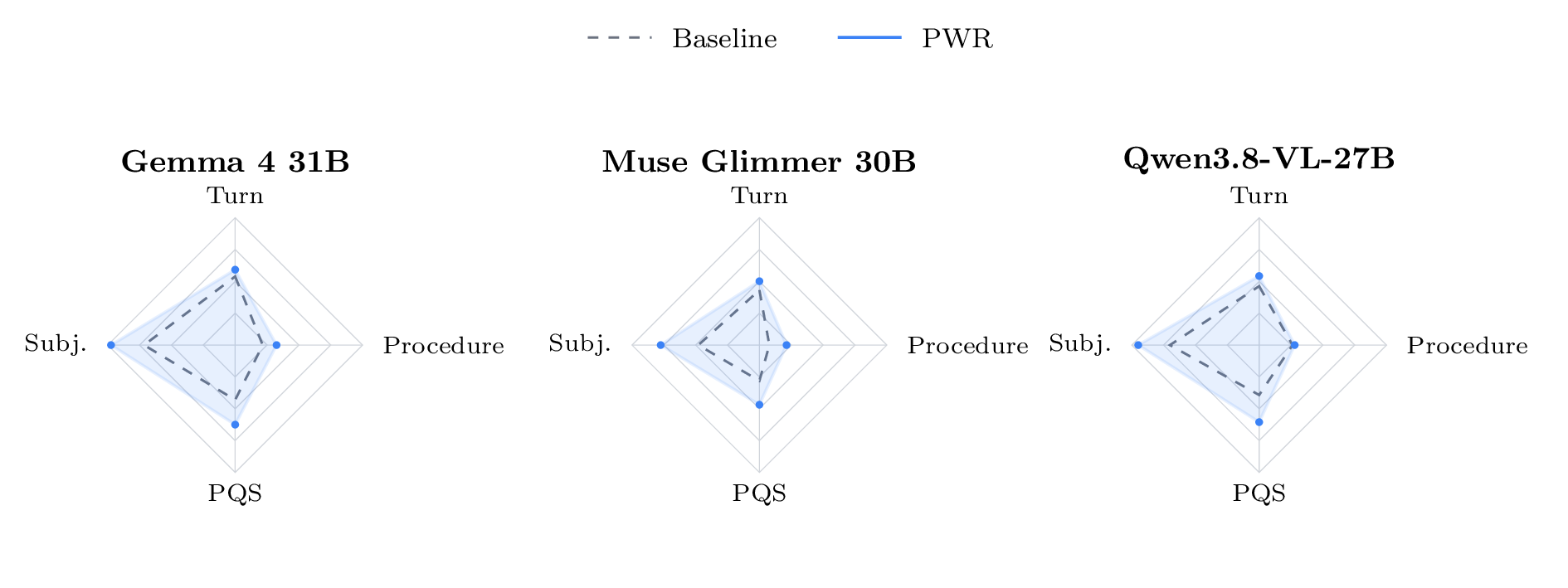}
\vspace{-1.2cm}
\captionsetup{font=footnotesize}
\captionof{figure}{\textbf{Pooled performance profiles.} Zero-Shot and PWR are compared on Turn Macro-F1, Procedure Macro, PQS, and subjective response quality (with scores rescaled from 1--5 to 0--1).}
\label{fig:combined-radar}
\end{minipage}
\end{figure}
}

\Needspace{8\baselineskip}

\section{Qualitative Results}\label{sec:qualitative}

\Cref{fig:qual_summary} shows two distinct failure modes: baselines can speak during an ongoing step or remain silent when guidance is warranted. In the tool-mistake case, Claude also interrupts but moves on to the handle without naming the needed tool change. PWR gives the specific correction. This exposes a guidance failure that an interrupt-only judgment would miss.

\ifdefined\PWRArxiv
\begin{figure}[H]
\else
\begin{figure}[t]
\fi
\centering
\includegraphics[width=\linewidth]{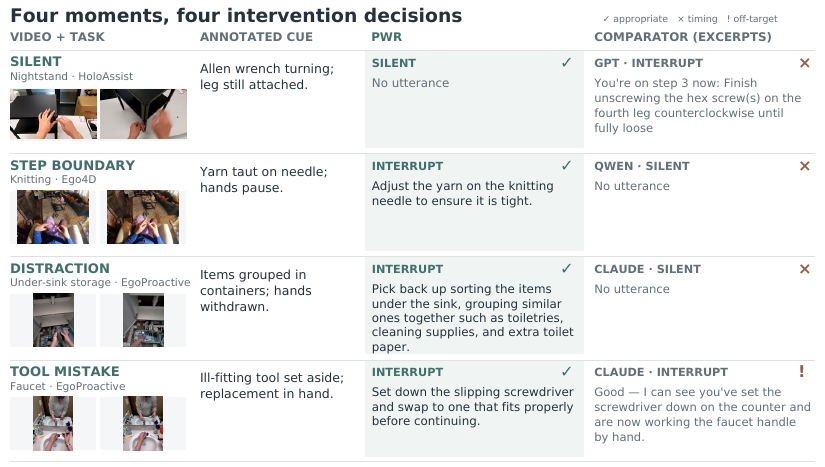}
\vspace{-6mm}
\caption{Qualitative analysis of four decision regimes. Frame pairs are sampled from each source clip. Cue descriptions are derived from the source annotations; PWR utterances are complete and longer comparator utterances are excerpted for display. More examples in App.\ref{app:more_examples}.}
\label{fig:qual_summary}
\end{figure}
\PWRRadarFigure

\section{Related Work}
\label{sec:related-work}
\paragraph{Egocentric procedural understanding.}
Ego4D~\citep{Grauman2022}, EPIC-KITCHENS~\citep{Damen2022EPIC100},
Ego-Exo4D~\citep{Grauman2024EgoExo4D}, and HowTo100M~\citep{Miech2019HowTo100M}
provide scale and domain coverage, but target recognition rather than
assistance. Assembly101~\citep{Sener2022}, CaptainCook4D~\citep{Peddi2023},
and IndustReal~\citep{Schoonbeek2024IndustReal} annotate mistakes without
paired recovery guidance, while HoloAssist~\citep{Wang2023} provides 166 hours
of expert-novice HoloLens~2 interaction with real-time guidance. Related
methods include \citet{Seminara2024} on task graphs,
\citet{Shen2024ProTAS} on progress-aware streaming segmentation, and
\citet{Li2025EgoProceduralSurvey} on the broader procedural-understanding
space. None combines wearable capture,
per-decision-point intervention labels, paired deviation and recovery, and a
shared multi-domain proactive-guidance schema.
\paragraph{From detection to recovery.}
Anticipation methods use LLMs as mental simulators~\citep{Zhao2024AntGPT} or
translate actions into language for long-term prediction~\citep{Kim2024PALM}.
Complementary work detects online egocentric mistakes~\citep{Flaborea2024PREGO},
unifies procedural and execution-state detection~\citep{Patsch2025MistSense},
or reasons about object-state effects~\citep{Guo2025ActionEffect}. These
methods identify that execution has diverged, but do not jointly evaluate
when to intervene and how to guide recovery across multi-domain wearable
procedures. EgoProactive makes this transition an explicit target.
\paragraph{Streaming perception and proactive intervention.}
A proactive assistant must process streaming video and intervene without
disrupting the user. Dispider~\citep{Qian2025Dispider} separates perception,
decision, and reaction, while StreamBridge~\citep{Wang2025StreamBridge} adapts
offline video language models to streaming inference. For timing,
\citet{Li2024Satori} trigger guidance from belief and task-graph divergence,
\citet{Zhang2024PrISM} use smartwatch sensing, and
\citet{Chen2025LlamaPIE} separate when and what to communicate; DARPA
PTG~\citep{DARPAptg2022}
field-tested augmented-reality guidance. These systems advance streaming
perception and timing, but generally assume canonical execution, a single
domain, or omit joint evaluation of timing and free-form recovery.

\paragraph{Proactive procedural assistance.}
Most closely related, \citet{Zhang2025ProAssist} map streaming egocentric video to proactive dialogue using 30K synthetic dialogues and identify procedural reasoning as the main bottleneck.
We evaluate that reasoning on real step-level deviations across domains and at the procedure level, and we address it with an explicit plan state that stays current as the user acts.
% \paragraph{Proactive procedural assistance.}
% Most closely related, \citet{Zhang2025ProAssist} map streaming
% egocentric video to proactive dialogue using 30K synthetic dialogues over six
% benchmarks and identifies procedural reasoning as the principal bottleneck.
% We differ in four respects. \emph{First}, EgoProactive evaluates canonical and
% out-of-plan execution with 9,899 wearable decision points, including 3,810
% points from 247 deviation and recovery procedures. \emph{Second},
% Pro\textsuperscript{2}Bench unifies six sources under procedural plans,
% per-decision intervention labels, and reference guidance, yielding 26,874
% evaluation points. \emph{Third}, PWR separates frequent interaction decisions
% from longer-horizon planning and recovery updates. \emph{Fourth}, one LoRA
% recipe transfers across Gemma~4 31B, Muse Glimmer 30B, and Qwen3.8-VL-27B.
% Together, these advances contribute EgoProactive,
% Pro\textsuperscript{2}Bench (\S\ref{sec:dataset}), the decoupled architecture
% (\S\ref{sec:method}), and a cross-family post-training recipe.
% Let Related Work fill around floats; keep remaining floats before Conclusion.
\FloatBarrier
\Needspace{13\baselineskip}
\section{Conclusion}
\label{sec:conclusion}

We introduced EgoProactive, a multi-domain smart-glasses dataset of step-level deviations and their recoveries, and \probench{}, a benchmark that scores proactive assistants per turn and per procedure.
Current models complete at most $26\%$ of a procedure's interventions before their first error.
PWR keeps an explicit plan state in context and revises it after every intervention.
Across three backbones the gains come from the plan rather than the training data, and they hold on held-out sources and on deviations never seen in training.
Revising the plan matters most where the plan must change, which makes planning under deviation the natural next target.

{
\small
\bibliographystyle{plainnat}
\bibliography{refs}
}

% \newpage
\appendix
% \input{tex/appendix_arxiv}
% \newpage
\appendix

% ============================================================
% Appendix: Method details (moved from main-paper Method)
% Include inside \appendix via \input{appendix_method}
% ============================================================

\section{Method Details}
\label{app:method}

This appendix gives the full details summarized in \S\ref{sec:train-infer}.

% ------------------------------------------------------------
\subsection{Plan-Anchored Clip Selection}
\label{sec:clip-selection}

The interaction model cannot consume $v_{1:t}$ in full at every timestep.
Rather than a recency-only window, which discards long-range context in sessions spanning minutes to hours, we let the plan decide which past video to keep.
At each timestep the interaction model receives up to $K = 15$ non-contiguous 8\,s clips from $v_{1:t}$.
One clip covers the most recent 8\,s ending at $t$.
The remaining clips are anchored at prior plan-update timestamps, keeping the $K - 1$ most recent when more updates have occurred.
Since the session opens with a plan update at $t = 0$, at least one anchored clip is always available.
The dynamic context therefore shapes both \emph{what} the model looks for, through $\mathbf{c}$, and \emph{where} in the timeline it looks.

\paragraph{Application to baselines.}
All implicit and static context baselines, including our reimplementation of \citet{Zhang2025ProAssist}, receive the same clip selection.
For the implicit and static baselines, anchors are the timestamps of their own past interrupts, which is the same rule as for dynamic context, since plan updates occur exactly at interrupts.
Differences between the three regimes therefore reflect the procedural context and not the visual input.

% ------------------------------------------------------------
\subsection{Post-Training on Plan-Conditioned Trajectories}
\label{sec:training}

This section expands the post-training summary in \S\ref{sec:train-infer}.
We find that the gains of dynamic context are largest after post-training on data that pairs each decision with the plan state that should inform it (\S\ref{sec:experiments}).

\paragraph{Plan-state densification.}
Pro$^2$Bench annotates interrupts sparsely.
For a session with annotated interrupt timesteps $\tau_0 = 0 < \tau_1 < \cdots < \tau_M$, each $\tau_m$ carries a ground-truth utterance $u^{*}_{\tau_m}$ and a ground-truth plan state $P^{*}_{\tau_m}$.
We convert these into dense plan-conditioned supervision.
For every timestep $t$, the context plan is $P^{*}_{t-1} = P^{*}_{\tau_m}$ with $\tau_m$ the latest interrupt strictly before $t$.
Every silent frame thus becomes an example of \emph{staying silent given a specific current step and its cues}, and every interrupt becomes an example of \emph{speaking because the observation matched or broke that expectation}.
This turns a sparse event annotation into a trajectory over plan states, which is the form of data the dynamic context formulation requires.

\paragraph{Objectives.}
Both models are trained with supervised fine-tuning.
The combined objective is $\mathcal{L} = \mathcal{L}_{\theta} + \mathcal{L}_{\phi}$ with
\[
    \mathcal{L}_{\theta} = -\mathbb{E}_{t}\bigl[
        \log p_{\theta}(y_t \mid o_t, P_{t-1}^{*})
    \bigr],
    \qquad
    \mathcal{L}_{\phi} = -\mathbb{E}_{m}\bigl[
        \log p_{\phi}(P_{\tau_m}^{*} \mid u_{\tau_m}^*, o_{\tau_m}, P_{\tau_{m-1}}^{*})
    \bigr],
\]
where $y_t = (d_t, u_t)$ is the target sequence, equal to $(\textsc{Silent}, \emptyset)$ at silent timesteps and $(\textsc{Interrupt}, u_t^{*})$ at interrupt timesteps, and $P^{*}_{\tau_{-1}} = \emptyset$ at the session-opening interrupt, where $o_{\tau_0} = (q, \emptyset)$.
Silent timesteps contribute a single-token target and interrupt timesteps the full utterance, both under the same autoregressive cross-entropy.
The parameters $\theta$ and $\phi$ appear in disjoint terms, so the two models can be trained independently.

\paragraph{Sampling.}
Silent frames dominate procedural video, which creates a severe class imbalance~\citep{Zhang2025ProAssist}.
We sample silent and interrupt timesteps for $\mathcal{L}_{\theta}$ at roughly $1{:}1$.
Clip anchors come from ground-truth plan-update timestamps, and the same clip selection is used at training and inference.
The planner is trained only on annotated interrupts, with the most recent clip as its visual input.

\paragraph{Teacher forcing on the plan.}
Training conditions on ground-truth plan states $P^{*}$, while inference conditions on predicted states $\hat{P}$.
The interaction model never sees an imperfect plan during training.
We quantify this gap by reporting both oracle-plan and predicted-plan variants (\S\ref{sec:experiments}).
Closing it, for example by training on the planner's own rollouts, is left to future work.

% ------------------------------------------------------------
\subsection{Inference}
\label{sec:inference}

At each timestep $t$ the interaction model receives $(o_t, P_{t-1})$ and produces $(d_t, u_t)$.
If $d_t = \textsc{Silent}$, then $P_t = P_{t-1}$ by Eq.~\ref{eq:planner-cases} and the planner is not called.
If $d_t = \textsc{Interrupt}$, the planner is called on $(d_t, u_t, o_t, P_{t-1})$ to produce $P_t$.
Two further properties follow from keeping the procedure in context.

\textbf{Reactive turns for free.}
The interaction model runs at every timestep regardless of who is speaking.
A user question mid-task (e.g., \textit{``How long should I tamp?''}) enters $\mathcal{H}_{<t}$, and the next forward pass can answer it with the current plan state in view.

\textbf{Recovery without OOP-specific machinery.}
Step completions and OOP deviations trigger the same update.
Only the planner's revision differs.
No deviation-specific module or loss is used.
Recovery follows from comparing the observation against an explicit plan state and rewriting that state when they disagree.
We observe sensible recovery plans on held-out EgoProactive deviations with no training on that source (\S\ref{sec:experiments}).

% \input{tex/appendix_planner_judge}

% ============================================================
% Appendix: Results Details
% Include via \input{appendix_results}
% ============================================================
\section{Results: Detailed Analysis}
\label{app:results}
This section expands the main findings in \S\ref{sec:main-findings}. It
summarizes the matched evaluation conditions, reports cross-model and
held-out-source trends, and provides additional analysis of the
plan-conditioned controls. Complete per-dataset Turn and subjective results
are reported in Tables~\ref{tab:detail-egoexo4d}--\ref{tab:detail-egoproactive}.
\subsection{Setup and Comparisons}
\label{app:results-setup}
\paragraph{Models.}
We evaluate three proprietary models, GPT-5.6, Gemini 3.6 Flash, and Claude
Opus 5, and five open-weight models, Kimi K3, Inkling, Gemma~4 31B, Muse
Glimmer 30B, and Qwen3.8-VL-27B. The three trained systems use Gemma, Muse,
and Qwen model families. Each pairs a learned planner with a corresponding LoRA
interaction model trained on EgoExo4D, EPIC-KITCHENS, HowTo100M, and Ego4D.
HoloAssist and EgoProactive are excluded from training and therefore measure
transfer to held-out sources.
\paragraph{Conditions.}
The main paper compares \textbf{Baseline}, \textbf{Fine-tuned, no plan}, and
\textbf{PWR}. The fine-tuned no-plan row uses the same LoRA interaction checkpoint
as PWR but omits planner state, separating training from planner conditioning.
PWR pairs the learned planner with that interaction model.
Appendix tables additionally report a
\textbf{Plan} control. This is the stronger matched result from either a fixed
reference plan or a row-local dynamic plan generated by the same model. The
best-of selection favors the control; the same selected condition is then used
for that model and dataset across Turn, Procedure, and subjective summaries.
Full input and prompting definitions are provided in
Appendix~\ref{app:eval-conditions}.
\paragraph{Metrics.}
Turn Macro-F1 is the arithmetic mean of interrupt F1 and silent F1. Procedure
Macro measures the normalized correct prefix before the first incorrect
decision, using the same Turn predictions under reference dialogue history.
PQS assigns full credit to correct silences, quality-weighted credit to correct
interruptions, and zero credit to incorrect decisions. Subjective response
quality is the mean of relevance, specificity, actionability, and conciseness
on a 1--5 scale. Metric definitions and the frozen judging rubric are given in
Appendix~\ref{app:eval-turn-procedure} and Appendix~\ref{app:eval-subjective}.
\subsection{Aggregate Results}
\label{app:results-findings}
\paragraph{Training and planning contributions are separated.}
PWR improves over the unmodified Baseline in all nine aggregate comparisons
across Turn Macro-F1, Procedure Macro, and PQS (Table~\ref{tab:main}). The
fine-tuned no-plan control separates training from planner conditioning. PWR
leads seven of eight completed comparisons with this control; Qwen Procedure
Macro is .298 without a plan and .279 with PWR. These controls show that the
aggregate gains combine training and plan conditioning.
\paragraph{The gains replicate across model families.}
Gemma improves from .539 to .592 in Turn Macro-F1, .216 to .325 in Procedure
Macro, and .430 to .625 in PQS. Muse improves from .432 to .501, .078 to .213,
and .280 to .468, respectively. Qwen improves from .463 to .542, .262 to .279,
and .392 to .605. Gemma provides the strongest aggregate PWR result on all
three reported metrics. Qwen provides the largest PWR gain over its unmodified
baseline in Turn and PQS, while Muse provides the largest Procedure gain.
\subsection{Generalization to Held-Out Sources}
\label{app:results-held-out}
\paragraph{HoloAssist.}
PWR improves Turn Macro-F1 and PQS for all three trained model families on
HoloAssist (Table~\ref{tab:detail-holoassist}). Gemma improves from
.571/.451 to .600/.630, Muse from .464/.366 to .581/.542, and Qwen from
.452/.372 to .504/.574, where each pair reports Turn Macro-F1/PQS.
\paragraph{EgoProactive.}
The same pattern holds on EgoProactive
(Table~\ref{tab:detail-egoproactive}). Gemma improves from .584/.478 to
.598/.582, Muse from .449/.290 to .509/.533, and Qwen from .450/.354 to
.563/.594. These gains are obtained without using either held-out source for
interaction-model training.
\subsection{Out-of-Plan Recovery}
\label{app:results-oop}
The canonical EgoProactive-OOP subset tests whether the system can recognize
and recover from deviations rather than only follow a nominal procedure.
PWR improves all nine comparisons with the unmodified no-plan baselines in
Table~\ref{tab:oop-results}. For Gemma, Turn Macro-F1, Procedure Macro, and PQS
change from .579/.167/.457 to .582/.245/.551. Muse improves from
.451/.024/.254 to .484/.134/.504. Qwen improves from .434/.038/.338 to
.553/.291/.566. The largest Procedure improvement is obtained by Qwen, and the
largest PQS improvement is obtained by Muse. This result is especially
important because EgoProactive and its OOP deviations are excluded from the
training mixture. Table~\ref{tab:oop-static-dynamic} separately isolates the
effect of updating the plan within these OOP procedures.
\subsection{Plan-Conditioned Controls}
\label{app:results-plan-controls}
The Plan controls measure the effect of supplying procedural state to the
interaction model. At the aggregate
level, Gemma Plan obtains .560 Turn Macro-F1, .361 Procedure Macro, and .581 PQS;
Muse Plan obtains .391, .030, and .320; and Qwen Plan obtains .484, .191, and
.495. PWR exceeds these controls in eight of the nine matched aggregate
comparisons. The only exception is Gemma Procedure Macro, where Plan reaches
.361 and PWR reaches .325. PWR nevertheless provides Gemma's strongest Turn
Macro-F1 and PQS, and it improves all three metrics over Plan for both Muse and
Qwen. The control is therefore not uniformly weaker: procedural context can
already be competitive, and the benefit of the learned planner depends on the
model and metric.
The per-dataset controls are intentionally retained in the appendix rather
than the headline table. They show where a reference plan or same-model
dynamic plan is already sufficient and where the matched PWR pairing adds a
further gain. Tables~\ref{tab:detail-egoexo4d}--\ref{tab:detail-egoproactive}
report the selected Plan control alongside Baseline and PWR under identical
frozen examples and denominators.
\subsection{Subjective Response Quality}
\label{app:results-subj-patterns}
PWR improves pooled subjective response quality over the unmodified baseline
for every trained model family.
On the original 1--5 scale, the pooled four-dimension mean increases from
3.58 to 4.87 for Gemma, 2.40 to 3.87 for Muse, and 3.50 to 4.75 for Qwen.
Gemma PWR remains between 4.79 and 4.98 across the five per-dataset subjective
means; Qwen remains between 4.56 and 4.90. Muse improves on every source and
reaches 4.00 on the held-out EgoProactive set. The accompanying PQS gains show
that these improvements are not obtained by evaluating response fluency in
isolation: a response receives quality credit only after a correct decision to
speak.
\subsection{Evaluation Scope}
\label{app:results-scope}
All comparisons in this section use the same frozen decision points, frames,
preceding canonical dialogue, and denominators. Each decision point is scored
independently. Procedure results are deterministic aggregations of those Turn
predictions and do not invoke a second model evaluation. PWR therefore tests
the planner-to-interaction sequence defined in Appendix~\ref{app:eval-conditions}.
The benchmark evaluates annotated decision points and correct-prefix
consistency; continuous-stream serving measurements are outside its metric
definition.

\section{Evaluation Protocol}
\label{app:eval}
This section specifies the models, frozen inputs, planning conditions,
prompts, and metrics used in the reported experiments.
\subsection{Models and Decoding}
\label{app:eval-models}
We evaluate eight VLMs. All reported interaction predictions use deterministic
decoding (temperature $0$) and must emit either \texttt{\$silent\$} or
\texttt{\$interrupt\$} followed by one concise instruction. Each decision
point uses the same frozen frames, dialogue history, and decoding contract
across matched conditions; the reference response is retained only for
scoring. The three trained model
families use a learned planner and a matched LoRA interaction model.
Table~\ref{tab:model-configs} summarizes their evaluation roles.
\paragraph{Input sanitization.}
\label{app:eval-input-sanitization}
Model inputs contain only information available through the current decision
point. Before prompt construction, we remove annotation-only tags such as
\texttt{(mistake)}, \texttt{(interruption)}, and \texttt{(fix)}, together
with event categories, target decisions, and reference responses. Plan context
is sanitized in the same pass: it contains no future deviation, scripted
mistake, recovery action, or later observation. These fields remain only in
the scoring records and are never exposed to either the planner or interaction
model.
\begin{table}[t]
\centering
\caption{Model configurations used in the reported evaluation.}
\label{tab:model-configs}
\setlength{\tabcolsep}{3.5pt}
\renewcommand{\arraystretch}{1.10}
\small
\begin{tabular}{@{}l l@{}}
\toprule
\textbf{Model} & \textbf{Evaluation role} \\
\midrule
GPT-5.6             & Proprietary baseline; subjective judge, not self-scored \\
Gemini 3.6 Flash    & Proprietary baseline \\
Claude Opus 5       & Proprietary baseline \\
Kimi K3             & Open-weight baseline \\
Inkling             & Open-weight baseline \\
\midrule
Gemma~4 31B         & Baseline + LoRA interaction model + learned planner \\
Muse Glimmer 30B    & Baseline + LoRA interaction model + learned planner \\
Qwen3.8-VL-27B      & Baseline + LoRA interaction model + learned planner \\
\bottomrule
\end{tabular}
\end{table}
GPT-5.6 is used as a text-only judge under a frozen rubric. To avoid
self-evaluation, we do not report PQS or response-quality scores for the
GPT-5.6 candidate. Exact response matches receive deterministic full credit;
all remaining true-positive interrupts from other models are judged with the
same prompt and schema.
\subsection{Frozen Evaluation Conditions}
\label{app:eval-conditions}
The main paper reports the baseline, a matched fine-tuned no-plan control, and
PWR. Additional plan controls are retained in the appendix:
\begin{description}
  \item[Baseline.] No planner state is supplied to the unmodified interaction
  model.
  \item[Fine-tuned, no plan.] No planner state is supplied to the LoRA interaction
  model used by PWR. This isolates interaction-model training from planner
  conditioning.
  \item[Reference plan.] The interaction model receives a fixed reference plan
  for the current procedure.
  \item[Dynamic plan.] The same model updates the reference-seeded plan from
  the current observation and preceding canonical history before producing the
  interaction response. Appendix tables label the stronger result from the
  reference-plan and dynamic-plan controls as \textbf{Plan}. This best-of
  control is deliberately favorable to plan conditioning; one selection is
  then shared across its Turn, Procedure, and subjective summaries.
  \item[PWR.] The planner first produces $P_0$ after the user's procedural
  request. At decision point $t$, the LoRA interaction model receives the current
  frames $o_t$, preceding dialogue $\mathcal{H}_{<t}$, and latest plan
  $P_{t-1}$, then predicts $(\hat d_t,\hat u_t)$. A predicted interrupt invokes
  the planner, which advances the state to $P_t$ for subsequent decision
  points; otherwise $P_t=P_{t-1}$.
\end{description}
The matched conditions use the same decision points, frames, history,
reference response, and decoding configuration, so their denominators match
exactly. PWR is evaluated separately using a dedicated learned planner paired
with the LoRA interaction model.
\paragraph{Reported PWR information flow.}
The evaluation follows the same decision-then-update order as the architecture:
\[
P_{t-1}\rightarrow(o_t,\mathcal{H}_{<t},P_{t-1})
\rightarrow(\hat d_t,\hat u_t)
\rightarrow
\begin{cases}
P_t=\mathcal{U}(o_t,\mathcal{H}_{<t},P_{t-1}),
  & \hat d_t=\texttt{interrupt},\\
P_t=P_{t-1}, & \hat d_t=\texttt{silent}.
\end{cases}
\]
The existing future-step context in $P_{t-1}$ lets the interaction model announce the
next action when it detects completion; that interrupt then advances the
planner state used by later decision points. Visual observations and decision
point identities remain frozen across matched conditions.
\subsection{Frozen Interaction Prompt}
\label{app:eval-prompts}
The interaction model uses this system prompt. In no-plan
mode, the planner-state line is omitted; plan-conditioned evaluations insert
the appropriate reference or learned JSON plan.
\begin{tcolorbox}[colback=gray!5,colframe=gray!40,boxrule=0.4pt,
arc=2pt,top=4pt,bottom=4pt,left=6pt,right=6pt,
fontupper=\footnotesize\ttfamily]
You are a helpful, proactive multimodal assistant guiding a user through a
step-by-step procedural task from egocentric video.\\[2pt]
Track the current plan step using the supplied planner state and prior
conversation. Interrupt when the visible current step has completed and
provide only the next grounded action. Interrupt when a visible mistake,
skipped step, wrong order, substitution, or out-of-procedure deviation
requires recovery. If the step is underway, has not started, or completion is
uncertain, remain silent.\\[2pt]
Output exactly one of:\\
- \$silent\$\\
- \$interrupt\$ followed immediately by one concise actionable instruction\\[2pt]
Use only the progressive conversation, supplied plan state when present, and
chronological frames from the current authoritative interval. Never expose
reasoning, narrate the completed action, repeat an already issued instruction,
or use future information.
\end{tcolorbox}
\subsection{Learned Planner Prompt and State}
\label{app:planner-prompt}
For PWR, the planner is called once after the initial user request to create
$P_0$, then after each predicted interrupt to advance the plan. Update calls
receive the current 16-frame window, the latest generated plan, and the
preceding dialogue. Its frozen system prompt is:
\begin{tcolorbox}[colback=gray!5,colframe=gray!40,boxrule=0.4pt,
arc=2pt,top=4pt,bottom=4pt,left=6pt,right=6pt,
fontupper=\footnotesize\ttfamily]
You are a helpful planning agent. You will be provided with a conversation
between a user and an assistant, the current visual context, and your previous
plan.\\
Identify whether the user is beginning or continuing a procedural task.
Produce the procedural plan that should guide the assistant's next decision.
Preserve the task sequence supported by the supplied context. Include current,
next\_few, last\_few, eta, visual\_cues, and every dataset-specific field
required by the plan.\\
Respond with only one valid JSON object. Do not add an intent token, control
prefix, prose, markdown, or a code fence.
\end{tcolorbox}
The canonical plan object contains
\texttt{vertical}, \texttt{task}, \texttt{eta},
\texttt{visual\_cues.\{interrupt,silent\}}, \texttt{last\_few},
\texttt{current}, and \texttt{next\_few}. Step identifiers must remain unique
and ordered, and \texttt{eta} is constrained to $[0,8]$. The planner output
and interaction-model input must correspond to the same procedure and decision point.
Any malformed JSON, duplicate identity, frame mismatch, cross-procedure state,
future history, or incomplete denominator fails closed.
\subsection{Evaluation Population and Metric Denominators}
\label{app:eval-population}
The evaluation population is the frozen manifest obtained after
source-specific validity checks, deduplication, and train--evaluation leakage
checks. We retain every eligible decision point. We do not balance the class
labels, equalize source sizes, or draw separate samples for different metrics.
Every completed model condition therefore predicts the same 26{,}874 decision
points.
\begin{table}[t]
\centering
\caption{Frozen evaluation denominators. Interrupt and silent counts are the
decision-point denominators for Turn Macro-F1 and PQS. Procedure units are the
denominators for Procedure Macro.}
\label{tab:eval-denominators}
\setlength{\tabcolsep}{4pt}
\renewcommand{\arraystretch}{1.10}
\small
\begin{tabular}{@{}l r r r r@{}}
\toprule
\textbf{Source} & \textbf{Points} & \textbf{Interrupt} & \textbf{Silent} &
\textbf{Procedures} \\
\midrule
EgoExo4D      & 2{,}734 & 1{,}551 & 1{,}183 & 286 \\
EPIC-KITCHENS & 6{,}399 & 4{,}172 & 2{,}227 & 667 \\
HowTo100M     & 4{,}654 & 4{,}168 & 486     & 819 \\
HoloAssist    & 3{,}188 & 1{,}695 & 1{,}493 & 237 \\
EgoProactive  & 9{,}899 & 5{,}338 & 4{,}561 & 700 \\
\midrule
\textbf{Total} & \textbf{26{,}874} & \textbf{16{,}924} &
\textbf{9{,}950} & \textbf{2{,}709} \\
\bottomrule
\end{tabular}
\end{table}
A canonical procedure unit is identified by its source procedure, task, and
reference step sequence; a source video may therefore contribute more than
one procedure unit. Turn Macro-F1 computes interrupt and silent F1 over all
decision points in one source and averages the two class scores. Procedure
Macro reuses those predictions, orders them within each procedure unit, and
averages the normalized correct-prefix score across the procedure units in
that source. PQS again uses every decision point in the source: correct
silences receive 1, incorrect decisions receive 0, exact true-positive
responses receive deterministic full credit, and the remaining true-positive
responses receive their normalized GPT-5.6 quality score. The headline value
for each metric is the equal-weight mean of its five source-level values, not
a micro-average over the 26{,}874 points.
The EgoProactive-OOP analysis is a deterministic slice of the same frozen
population. It contains 3{,}810 decision points grouped into 247 procedures.
These points remain part of the pooled EgoProactive result and are not sampled
again for the OOP table.
\paragraph{Trivial decision baselines.}
Table~\ref{tab:trivial-turn-baselines} reports the Turn Macro-F1 obtained by
predicting one class at every decision point. These baselines make the effect
of source imbalance explicit and remain far below the learned systems.
\begin{table}[t]
\centering\small
\setlength{\tabcolsep}{5pt}
\begin{tabular}{@{}lcc@{}}
\toprule
\textbf{Source} & \textbf{Always interrupt} & \textbf{Always silent} \\
\midrule
EgoExo4D & .362 & .302 \\
EPIC-KITCHENS & .395 & .258 \\
HowTo100M & .472 & .095 \\
HoloAssist & .347 & .319 \\
EgoProactive & .350 & .315 \\
\midrule
\textbf{Five-source mean} & \textbf{.385} & \textbf{.258} \\
\bottomrule
\end{tabular}
\caption{Turn Macro-F1 for constant-decision baselines on the frozen
evaluation population.}
\label{tab:trivial-turn-baselines}
\end{table}
\subsection{Turn and Procedure Metrics}
\label{app:eval-turn-procedure}
Turn Macro-F1 is the arithmetic mean of the interrupt-class and silent-class
F1 scores. Invalid outputs count as errors. G-mean recall is retained as a
secondary diagnostic but is not the aggregate timing metric in
Table~\ref{tab:main}.
Procedure Macro measures sustained consistency across the ordered decision
points of a procedure. Correct silent turns extend the valid prefix and correct
interrupt turns increment the credited-interrupt count. The first wrong
decision or invalid output ends that prefix, and later predictions receive no
credit. For procedure $p$, let $c_p$
be the number of gold interrupt points reached within the correct prefix and
$L_p$ the total number of gold interrupt points. Its correct-prefix score is
$c_p/L_p$, and Procedure Macro is
$|\mathcal{P}|^{-1}\sum_{p\in\mathcal{P}} c_p/L_p$. It measures
correct-prefix consistency across a procedure under reference history.
EgoProactive-OOP is obtained by a deterministic procedure-level metadata split
and uses the same scoring rule.
\paragraph{Output validity.}
Invalid interaction outputs are counted as incorrect decisions. In the final
PWR artifacts, all three model families produce a parseable decision for all
26{,}874 points, and all planner records pass the required JSON schema after
validation.
\subsection{Subjective Quality and PQS}
\label{app:eval-subjective}
Only true-positive interrupts are scored for response quality. GPT-5.6 assigns
integer scores from 1 to 5 for relevance, specificity, actionability, and
conciseness under this frozen rubric. Exact matches to the reference
response receive deterministic full credit.
\begin{tcolorbox}[colback=gray!5,colframe=gray!40,boxrule=0.4pt,
arc=2pt,top=4pt,bottom=4pt,left=6pt,right=6pt,
fontupper=\footnotesize\ttfamily]
You are a STRICT evaluator of procedural AI assistant responses. Your job is
to find flaws, not praise.\\[2pt]
The assistant guides a user through step-by-step tasks via egocentric video.
When a step is complete, it should proactively interrupt with the next
step.\\[2pt]
Golden response: \{golden\}\\
Model response: \{model\}\\
Task context: \{task\_context\}\\[2pt]
Rate relevance, specificity, actionability, and conciseness from 1--5. A 5
means a near-perfect match to the golden response, 3 means acceptable but
flawed, and 1 means wrong or harmful. Return only a JSON object containing the
four integer scores and one-sentence reasoning.
\end{tcolorbox}
For a judged true positive $i$, let $q_i=(\bar{s}_i-1)/4$, where
$\bar{s}_i$ is the mean of its four 1--5 scores. Over all $N$ decision points,
\[
  \mathrm{PQS}=\frac{\mathrm{TN}+\sum_{i\in\mathrm{TP}}q_i}{N}.
\]
Thus correct silences receive full credit, true-positive interruptions receive
quality-weighted credit, and false positives, false negatives, and invalid
outputs receive zero credit.
\paragraph{Judge scope.}
PQS is a deterministic, reference-aligned comparison under one fixed rubric.
We report its four response dimensions separately so that timing and language
quality can be inspected independently. GPT-5.6 is excluded as a PQS candidate
because it is the judge.
\paragraph{Statistical scope.}
All inference uses deterministic decoding and checkpoints selected before the
frozen evaluation. Table~\ref{tab:eval-denominators} reports the exact procedure
and decision-point denominators, preserving the procedure-level clustering used
by Procedure Macro.
\paragraph{Reproducibility note.}
All matched conditions use the same frozen evaluation examples, visual inputs,
dialogue histories, prompts, model versions, and deterministic decoding. Frames are
sampled chronologically at 2\,fps from each 8\,s interval, yielding 16 frames
when the complete interval is available. Procedure Macro is derived from Turn
predictions with identical denominators and reference histories.

\section{Per-Dataset Statistics}
\label{app:stats}
This section reports detailed statistics for the five evaluation datasets and
the six-source benchmark collection. Ego4D contributes training data but is not
part of the frozen evaluation split.
\paragraph{Provenance.}
All values are computed deterministically from the released evaluation manifest
and its canonical semantic-unit identifiers. The same 26{,}874 decision points
are used for every matched model condition. The released statistics file and
the accompanying processing scripts reproduce
Tables~\ref{tab:app-eval-detail}--\ref{tab:app-train-stats}.
\subsection{Evaluation Set Distributions}
\begin{table*}[t]
\centering
\caption{Detailed statistics for the frozen evaluation split. Int. and Sil.
denote the percentages of interrupt and silent decision points. Response length
is measured on reference responses for interrupt points only. OOP denotes all
rows belonging to canonical EgoProactive out-of-plan procedures.}
\label{tab:app-eval-detail}
\setlength{\tabcolsep}{4pt}
\renewcommand{\arraystretch}{1.12}
\small
\begin{tabular*}{\textwidth}{@{\extracolsep{\fill}}l r r r r r r@{}}
\toprule
\textbf{Dataset} & \textbf{Eval $n$} & \textbf{Int.\%} & \textbf{Sil.\%} &
\textbf{Avg. resp. words} & \textbf{Procedures} & \textbf{OOP\%} \\
\midrule
EgoExo4D          & 2{,}734 & 56.7 & 43.3 & 9.5  & 286 & 0.0 \\
EPIC-KITCHENS     & 6{,}399 & 65.2 & 34.8 & 10.9 & 667 & 0.0 \\
HowTo100M         & 4{,}654 & 89.6 & 10.4 & 11.0 & 819 & 0.0 \\
HoloAssist        & 3{,}188 & 53.2 & 46.8 & 6.9  & 237 & 0.0 \\
EgoProactive      & 9{,}899 & 53.9 & 46.1 & 16.4 & 700 & 38.5 \\
\midrule
\textbf{Total}    & \textbf{26{,}874} & \textbf{63.0} & \textbf{37.0} &
\textbf{12.1} & \textbf{2{,}709} & \textbf{14.2} \\
\bottomrule
\end{tabular*}
\end{table*}
The split contains 16{,}924 interrupt and 9{,}950 silent decision points.
EgoProactive contains 3{,}810 rows from 247 out-of-plan procedures. Its
canonical pooled score includes both out-of-plan and non-out-of-plan
procedures.
\subsection{Procedure Length Distributions}
Table~\ref{tab:step-counts} summarizes the number of reference-plan steps per
canonical evaluation procedure.
\begin{table}[t]
\centering
\caption{Reference-plan step counts for evaluation procedures. Range gives the
minimum and maximum number of steps in a canonical semantic unit.}
\label{tab:step-counts}
\setlength{\tabcolsep}{6pt}
\renewcommand{\arraystretch}{1.12}
\begin{tabular}{@{}l c c r@{}}
\toprule
\textbf{Dataset} & \textbf{Range} & \textbf{Avg. steps} & \textbf{Procedures} \\
\midrule
EgoExo4D          & 2--15 & 5.4 & 286 \\
EPIC-KITCHENS     & 2--15 & 6.3 & 667 \\
HowTo100M         & 1--13 & 5.1 & 819 \\
HoloAssist        & 2--20 & 7.2 & 237 \\
EgoProactive      & 3--15 & 7.6 & 700 \\
\midrule
\textbf{All}      & 1--20 & 6.2 & 2{,}709 \\
\bottomrule
\end{tabular}
\end{table}
\subsection{Response Length Distributions}
Table~\ref{tab:resp-lengths} reports reference-response lengths for interrupt
decision points.
\begin{table}[t]
\centering
\caption{Reference response lengths for interrupt decision points.}
\label{tab:resp-lengths}
\setlength{\tabcolsep}{6pt}
\renewcommand{\arraystretch}{1.12}
\begin{tabular}{@{}l c c@{}}
\toprule
\textbf{Dataset} & \textbf{Range (words)} & \textbf{Avg. words} \\
\midrule
EgoExo4D          & 3--36 & 9.5 \\
EPIC-KITCHENS     & 2--38 & 10.9 \\
HowTo100M         & 2--46 & 11.0 \\
HoloAssist        & 2--17 & 6.9 \\
EgoProactive      & 1--62 & 16.4 \\
\midrule
\textbf{All}      & 1--62 & 12.1 \\
\bottomrule
\end{tabular}
\end{table}
EgoProactive has the longest reference responses because deviation and recovery
guidance often identifies both the error and the corrective action. HoloAssist
has the shortest responses, reflecting concise assembly instructions.
\subsection{Domain and Viewpoint Coverage}
Table~\ref{tab:domain-coverage} summarizes the activity and viewpoint coverage
of each benchmark source.
\begin{table}[t]
\centering
\caption{Activity and viewpoint coverage across benchmark sources. Ego4D is
used for training only; the remaining five sources form the frozen evaluation
split.}
\label{tab:domain-coverage}
\setlength{\tabcolsep}{4pt}
\renewcommand{\arraystretch}{1.12}
\small
\begin{tabular}{@{}l l l@{}}
\toprule
\textbf{Dataset} & \textbf{Viewpoint} & \textbf{Primary coverage} \\
\midrule
Ego4D             & Egocentric & Broad everyday activities \\
EgoExo4D          & Ego + exo & Skilled procedural activities \\
EPIC-KITCHENS     & Egocentric & Cooking \\
HowTo100M         & Third person & Broad instructional activities \\
HoloAssist        & Egocentric AR & Assembly and repair \\
EgoProactive      & Smart glasses & Cooking, tutorials, repair, arts and crafts \\
\bottomrule
\end{tabular}
\end{table}
\subsection{Training and Evaluation Coverage}
Table~\ref{tab:app-train-stats} identifies the sources used for training and
evaluation in the reported experiments.
\begin{table}[t]
\centering
\caption{Source participation in the reported experiments. HoloAssist and
EgoProactive are held out from planner and interaction-model training.}
\label{tab:app-train-stats}
\setlength{\tabcolsep}{5pt}
\renewcommand{\arraystretch}{1.12}
\small
\begin{tabular}{@{}l c c c@{}}
\toprule
\textbf{Source} & \textbf{Interaction train} & \textbf{Planner train} & \textbf{Evaluation} \\
\midrule
Ego4D             & \checkmark & \checkmark & -- \\
EgoExo4D          & \checkmark & \checkmark & \checkmark \\
EPIC-KITCHENS     & \checkmark & \checkmark & \checkmark \\
HowTo100M         & \checkmark & \checkmark & \checkmark \\
HoloAssist        & -- & -- & \checkmark \\
EgoProactive      & -- & -- & \checkmark \\
\bottomrule
\end{tabular}
\end{table}
Training and evaluation are separated at the canonical semantic-unit level,
with dataset-specific video and identity leakage checks applied before release.
All reported model conditions use the same frozen evaluation identifiers and
denominators.
\section{Per-Dataset Score Details}
\label{app:subj-detail}
Tables~\ref{tab:detail-egoexo4d}--\ref{tab:detail-egoproactive} report the complete per-dataset Turn and subjective breakdown. Turn columns report interrupt F1 (IF1), silent F1 (SF1), and their macro-average. Subjective columns report relevance (R), specificity (S), actionability (A), conciseness (C), their mean, and PQS. \textbf{No plan} omits planner state. \textbf{Plan} is the stronger of two matched controls: a fixed reference plan or a row-local dynamic plan produced by the same model. The selection is made independently for each model and dataset and is shared across its Turn, Session, and subjective summaries. \textbf{PWR} pairs the learned planner with the corresponding fine-tuned interaction model. Table labels abbreviate Gemini 3.6 Flash as Gemini-3.6, Claude Opus 5 as Opus-5, Gemma 4 31B as Gemma-31B, Muse Glimmer 30B as Muse-30B, and Qwen3.8-VL-27B as Qwen-27B. \colorbox{blue!8}{Fine-tuned configurations (Ours)}. \textbf{Bold} is best and \underline{underline} is second within a column.
\begin{table*}[t]
\centering\scriptsize
\setlength{\tabcolsep}{2.2pt}
\renewcommand{\arraystretch}{1.04}
\begin{tabular*}{\textwidth}{@{\extracolsep{\fill}}l l ccc cccccc@{}}
\toprule
& & \multicolumn{3}{c}{\textit{Turn}} & \multicolumn{6}{c}{\textit{Subjective}} \\
\cmidrule(lr){3-5}\cmidrule(lr){6-11}
\textbf{Model} & \textbf{Config} & \textbf{IF1} & \textbf{SF1} & \textbf{Macro-F1} & \textbf{R} & \textbf{S} & \textbf{A} & \textbf{C} & \textbf{Avg} & \textbf{PQS} \\
\midrule
GPT-5.6 & No plan & .443 & .546 & .494 & 2.78 & 2.85 & 4.19 & 4.86 & 3.67 & 0.434 \\
GPT-5.6 & Plan & .482 & .558 & .520 & 3.23 & 3.43 & 4.43 & 4.87 & 3.99 & 0.467 \\
Gemini-3.6 & No plan & .389 & .532 & .461 & 2.64 & 2.64 & 4.09 & 4.91 & 3.57 & 0.410 \\
Gemini-3.6 & Plan & .494 & .529 & .511 & 3.50 & 3.68 & 4.47 & 4.95 & 4.15 & 0.461 \\
Opus-5 & No plan & .604 & .525 & \underline{.564} & 3.11 & 3.18 & 4.41 & 4.47 & 3.79 & 0.468 \\
Opus-5 & Plan & .571 & .529 & .550 & 4.34 & 4.37 & 4.79 & 4.73 & 4.56 & 0.518 \\
Kimi-K3 & No plan & .436 & \underline{.567} & .501 & 2.92 & 3.03 & 4.33 & 4.72 & 3.75 & 0.451 \\
Kimi-K3 & Plan & .418 & .561 & .490 & 3.61 & 3.80 & 4.55 & 4.87 & 4.21 & 0.458 \\
Inkling & No plan & .603 & .494 & .548 & 3.01 & 3.05 & 4.30 & 4.43 & 3.70 & 0.445 \\
Inkling & Plan & .627 & .498 & .563 & 4.29 & 4.34 & 4.78 & 4.78 & 4.55 & 0.531 \\
Gemma-31B & No plan & .661 & .439 & .550 & 2.94 & 2.80 & 4.15 & 4.98 & 3.72 & 0.445 \\
\rowcolor{blue!8}Gemma-31B & Plan & .712 & .334 & .523 & 4.02 & 4.13 & 4.70 & \underline{4.99} & 4.46 & 0.531 \\
\rowcolor{blue!8}Gemma-31B & PWR & \textbf{.740} & .490 & \textbf{.615} & \textbf{4.90} & \textbf{4.93} & \textbf{4.97} & \textbf{5.00} & \textbf{4.95} & \textbf{0.649} \\
Muse-30B & No plan & .438 & .443 & .441 & 2.48 & 2.68 & 3.20 & 1.65 & 2.50 & 0.304 \\
\rowcolor{blue!8}Muse-30B & Plan & .342 & .503 & .423 & 2.86 & 3.11 & 3.25 & 1.65 & 2.72 & 0.350 \\
\rowcolor{blue!8}Muse-30B & PWR & .467 & \textbf{.646} & .557 & \underline{4.84} & \underline{4.87} & 4.11 & 1.85 & 3.92 & 0.524 \\
Qwen-27B & No plan & .674 & .250 & .462 & 2.81 & 2.77 & 4.12 & 4.93 & 3.66 & 0.388 \\
\rowcolor{blue!8}Qwen-27B & Plan & .582 & .431 & .506 & 4.43 & 4.41 & 4.76 & 4.97 & 4.64 & 0.488 \\
\rowcolor{blue!8}Qwen-27B & PWR & \underline{.720} & .358 & .539 & 4.60 & 4.63 & \underline{4.84} & \underline{4.99} & \underline{4.76} & \underline{0.580} \\
\bottomrule
\end{tabular*}
\caption{Per-dataset Turn and subjective results on EgoExo4D.}
\label{tab:detail-egoexo4d}
\end{table*}
\begin{table*}[t]
\centering\scriptsize
\setlength{\tabcolsep}{2.2pt}
\renewcommand{\arraystretch}{1.04}
\begin{tabular*}{\textwidth}{@{\extracolsep{\fill}}l l ccc cccccc@{}}
\toprule
& & \multicolumn{3}{c}{\textit{Turn}} & \multicolumn{6}{c}{\textit{Subjective}} \\
\cmidrule(lr){3-5}\cmidrule(lr){6-11}
\textbf{Model} & \textbf{Config} & \textbf{IF1} & \textbf{SF1} & \textbf{Macro-F1} & \textbf{R} & \textbf{S} & \textbf{A} & \textbf{C} & \textbf{Avg} & \textbf{PQS} \\
\midrule
GPT-5.6 & No plan & .420 & .498 & .459 & 2.11 & 2.13 & 4.03 & 4.88 & 3.29 & 0.378 \\
GPT-5.6 & Plan & .524 & .513 & .519 & 2.71 & 2.81 & 4.19 & 4.75 & 3.62 & 0.427 \\
Gemini-3.6 & No plan & .324 & .486 & .405 & 2.15 & 2.12 & 3.95 & 4.93 & 3.29 & 0.356 \\
Gemini-3.6 & Plan & .413 & .454 & .434 & 3.21 & 3.25 & 4.32 & 4.88 & 3.92 & 0.377 \\
Opus-5 & No plan & .624 & .457 & .540 & 2.39 & 2.38 & 4.21 & 4.54 & 3.38 & 0.406 \\
Opus-5 & Plan & .624 & .453 & .538 & 3.78 & 3.75 & 4.57 & 4.64 & 4.18 & 0.474 \\
Kimi-K3 & No plan & .393 & .507 & .450 & 2.18 & 2.20 & 4.12 & 4.71 & 3.30 & 0.381 \\
Kimi-K3 & Plan & .394 & .472 & .433 & 3.13 & 3.24 & 4.40 & 4.72 & 3.87 & 0.379 \\
Inkling & No plan & .607 & .451 & .529 & 2.17 & 2.11 & 4.05 & 4.35 & 3.17 & 0.380 \\
Inkling & Plan & .643 & .500 & \underline{.571} & 3.95 & 3.94 & 4.65 & 4.77 & 4.33 & 0.519 \\
Gemma-31B & No plan & .655 & .448 & .552 & 2.37 & 2.28 & 4.02 & 4.95 & 3.40 & 0.415 \\
\rowcolor{blue!8}Gemma-31B & Plan & \textbf{.769} & .344 & .556 & 3.82 & 3.91 & 4.63 & \underline{4.96} & 4.33 & 0.563 \\
\rowcolor{blue!8}Gemma-31B & PWR & \underline{.757} & \underline{.538} & \textbf{.648} & \textbf{4.65} & \textbf{4.65} & \textbf{4.88} & \textbf{4.97} & \textbf{4.79} & \textbf{0.655} \\
Muse-30B & No plan & .495 & .403 & .449 & 2.06 & 2.11 & 2.82 & 1.46 & 2.11 & 0.259 \\
\rowcolor{blue!8}Muse-30B & Plan & .363 & .446 & .405 & 2.76 & 2.90 & 3.12 & 1.55 & 2.58 & 0.302 \\
\rowcolor{blue!8}Muse-30B & PWR & .490 & \textbf{.576} & .533 & \underline{4.49} & \underline{4.53} & 4.05 & 2.05 & 3.78 & 0.469 \\
Qwen-27B & No plan & .738 & .178 & .458 & 2.14 & 2.12 & 4.01 & 4.95 & 3.30 & 0.366 \\
\rowcolor{blue!8}Qwen-27B & Plan & .649 & .283 & .466 & 3.67 & 3.65 & 4.52 & \underline{4.96} & 4.20 & 0.442 \\
\rowcolor{blue!8}Qwen-27B & PWR & .748 & .333 & .540 & 4.43 & 4.42 & \underline{4.79} & \underline{4.96} & \underline{4.65} & \underline{0.587} \\
\bottomrule
\end{tabular*}
\caption{Per-dataset Turn and subjective results on EPIC-KITCHENS.}
\label{tab:detail-epickitchens}
\end{table*}
\begin{table*}[t]
\centering\scriptsize
\setlength{\tabcolsep}{2.2pt}
\renewcommand{\arraystretch}{1.04}
\begin{tabular*}{\textwidth}{@{\extracolsep{\fill}}l l ccc cccccc@{}}
\toprule
& & \multicolumn{3}{c}{\textit{Turn}} & \multicolumn{6}{c}{\textit{Subjective}} \\
\cmidrule(lr){3-5}\cmidrule(lr){6-11}
\textbf{Model} & \textbf{Config} & \textbf{IF1} & \textbf{SF1} & \textbf{Macro-F1} & \textbf{R} & \textbf{S} & \textbf{A} & \textbf{C} & \textbf{Avg} & \textbf{PQS} \\
\midrule
GPT-5.6 & No plan & .544 & .211 & .377 & 2.66 & 2.71 & 4.16 & 4.85 & 3.59 & 0.301 \\
GPT-5.6 & Plan & .573 & .216 & .395 & 3.67 & 3.77 & 4.54 & 4.85 & 4.21 & 0.374 \\
Gemini-3.6 & No plan & .510 & .222 & .366 & 2.69 & 2.66 & 3.96 & 4.81 & 3.53 & 0.284 \\
Gemini-3.6 & Plan & .587 & .226 & .407 & 4.13 & 4.20 & 4.64 & 4.93 & 4.48 & 0.411 \\
Opus-5 & No plan & .738 & .257 & .498 & 2.80 & 2.85 & 4.26 & 4.43 & 3.59 & 0.420 \\
Opus-5 & Plan & .701 & .229 & .465 & 4.30 & 4.27 & 4.76 & 4.64 & 4.49 & 0.501 \\
Kimi-K3 & No plan & .420 & .195 & .308 & 2.79 & 2.88 & 4.28 & 4.62 & 3.64 & 0.243 \\
Kimi-K3 & Plan & .487 & .204 & .345 & 4.14 & 4.24 & 4.72 & 4.86 & 4.49 & 0.337 \\
Inkling & No plan & .743 & .187 & .465 & 2.50 & 2.56 & 4.15 & 4.17 & 3.35 & 0.376 \\
Inkling & Plan & .745 & .198 & .471 & 4.20 & 4.24 & 4.72 & 4.71 & 4.47 & 0.536 \\
Gemma-31B & No plan & .663 & .211 & .437 & 2.73 & 2.63 & 4.03 & 4.96 & 3.59 & 0.363 \\
\rowcolor{blue!8}Gemma-31B & Plan & \underline{.875} & \textbf{.272} & \textbf{.573} & 4.38 & 4.47 & \underline{4.81} & \textbf{4.99} & \underline{4.66} & \textbf{0.723} \\
\rowcolor{blue!8}Gemma-31B & PWR & .753 & .243 & .498 & \textbf{4.76} & \textbf{4.80} & \textbf{4.90} & \underline{4.98} & \textbf{4.86} & 0.608 \\
Muse-30B & No plan & .555 & .156 & .355 & 2.39 & 2.54 & 3.10 & 1.68 & 2.43 & 0.183 \\
\rowcolor{blue!8}Muse-30B & Plan & .419 & .153 & .286 & 2.53 & 2.80 & 3.03 & 1.65 & 2.50 & 0.147 \\
\rowcolor{blue!8}Muse-30B & PWR & .434 & .219 & .327 & \underline{4.64} & \underline{4.64} & 4.19 & 2.08 & 3.89 & 0.274 \\
Qwen-27B & No plan & \textbf{.877} & .105 & .491 & 2.40 & 2.48 & 4.00 & 4.84 & 3.43 & 0.481 \\
\rowcolor{blue!8}Qwen-27B & Plan & .800 & .176 & .488 & 4.09 & 4.13 & 4.64 & 4.93 & 4.45 & 0.590 \\
\rowcolor{blue!8}Qwen-27B & PWR & .864 & \underline{.268} & \underline{.566} & 4.27 & 4.31 & 4.70 & 4.96 & 4.56 & \underline{0.690} \\
\bottomrule
\end{tabular*}
\caption{Per-dataset Turn and subjective results on HowTo100M.}
\label{tab:detail-howto}
\end{table*}
\begin{table*}[t]
\centering\scriptsize
\setlength{\tabcolsep}{2.2pt}
\renewcommand{\arraystretch}{1.04}
\begin{tabular*}{\textwidth}{@{\extracolsep{\fill}}l l ccc cccccc@{}}
\toprule
& & \multicolumn{3}{c}{\textit{Turn}} & \multicolumn{6}{c}{\textit{Subjective}} \\
\cmidrule(lr){3-5}\cmidrule(lr){6-11}
\textbf{Model} & \textbf{Config} & \textbf{IF1} & \textbf{SF1} & \textbf{Macro-F1} & \textbf{R} & \textbf{S} & \textbf{A} & \textbf{C} & \textbf{Avg} & \textbf{PQS} \\
\midrule
GPT-5.6 & No plan & .417 & .532 & .474 & 2.41 & 2.64 & 4.16 & 4.85 & 3.52 & 0.412 \\
GPT-5.6 & Plan & .450 & .540 & .495 & 3.09 & 3.34 & 4.36 & 4.79 & 3.90 & 0.442 \\
Gemini-3.6 & No plan & .352 & .523 & .437 & 2.37 & 2.47 & 4.02 & 4.94 & 3.45 & 0.393 \\
Gemini-3.6 & Plan & .427 & .493 & .460 & 3.52 & 3.66 & 4.46 & 4.95 & 4.15 & 0.419 \\
Opus-5 & No plan & .600 & .495 & .547 & 2.90 & 3.08 & 4.37 & 4.30 & 3.66 & 0.441 \\
Opus-5 & Plan & .582 & .429 & .506 & 4.21 & 4.26 & 4.76 & 4.48 & 4.43 & 0.460 \\
Kimi-K3 & No plan & .440 & .582 & .511 & 2.57 & 2.77 & 4.25 & 4.78 & 3.59 & 0.455 \\
Kimi-K3 & Plan & .386 & .524 & .455 & 3.46 & 3.70 & 4.52 & 4.76 & 4.11 & 0.425 \\
Inkling & No plan & .604 & .495 & .550 & 2.51 & 2.68 & 4.14 & 4.55 & 3.47 & 0.427 \\
Inkling & Plan & .615 & .534 & .575 & 3.99 & 4.07 & 4.72 & 4.75 & 4.38 & 0.525 \\
Gemma-31B & No plan & .636 & .505 & .571 & 2.67 & 2.67 & 4.03 & 4.98 & 3.59 & 0.451 \\
\rowcolor{blue!8}Gemma-31B & Plan & .671 & .394 & .533 & 3.66 & 3.82 & 4.56 & \underline{4.99} & 4.26 & 0.493 \\
\rowcolor{blue!8}Gemma-31B & PWR & \textbf{.713} & .487 & \textbf{.600} & \textbf{4.97} & \textbf{4.97} & \textbf{4.99} & \textbf{5.00} & \textbf{4.98} & \textbf{0.630} \\
Muse-30B & No plan & .386 & .542 & .464 & 2.19 & 2.46 & 3.08 & 1.63 & 2.34 & 0.366 \\
\rowcolor{blue!8}Muse-30B & Plan & .195 & \underline{.596} & .396 & 2.55 & 2.91 & 3.07 & 1.67 & 2.55 & 0.422 \\
\rowcolor{blue!8}Muse-30B & PWR & .490 & \textbf{.671} & \underline{.581} & 4.67 & 4.66 & 4.00 & 1.80 & 3.78 & 0.542 \\
Qwen-27B & No plan & .650 & .255 & .452 & 2.73 & 2.77 & 4.06 & 4.96 & 3.63 & 0.372 \\
\rowcolor{blue!8}Qwen-27B & Plan & .643 & .296 & .469 & 4.32 & 4.35 & 4.69 & 4.95 & 4.58 & 0.481 \\
\rowcolor{blue!8}Qwen-27B & PWR & \underline{.707} & .300 & .504 & \underline{4.82} & \underline{4.85} & \underline{4.93} & \textbf{5.00} & \underline{4.90} & \underline{0.574} \\
\bottomrule
\end{tabular*}
\caption{Per-dataset Turn and subjective results on HoloAssist.}
\label{tab:detail-holoassist}
\end{table*}
\begin{table*}[t]
\centering\scriptsize
\setlength{\tabcolsep}{2.2pt}
\renewcommand{\arraystretch}{1.04}
\begin{tabular*}{\textwidth}{@{\extracolsep{\fill}}l l ccc cccccc@{}}
\toprule
& & \multicolumn{3}{c}{\textit{Turn}} & \multicolumn{6}{c}{\textit{Subjective}} \\
\cmidrule(lr){3-5}\cmidrule(lr){6-11}
\textbf{Model} & \textbf{Config} & \textbf{IF1} & \textbf{SF1} & \textbf{Macro-F1} & \textbf{R} & \textbf{S} & \textbf{A} & \textbf{C} & \textbf{Avg} & \textbf{PQS} \\
\midrule
GPT-5.6 & No plan & .602 & \textbf{.686} & .644 & 2.72 & 2.70 & 4.15 & 4.89 & 3.61 & 0.557 \\
GPT-5.6 & Plan & .586 & .665 & .626 & 3.73 & 3.67 & 4.52 & 4.88 & 4.20 & 0.577 \\
Gemini-3.6 & No plan & .590 & \underline{.685} & .638 & 2.78 & 2.62 & 3.92 & 4.72 & 3.51 & 0.548 \\
Gemini-3.6 & Plan & .640 & \underline{.685} & \underline{.662} & 4.08 & 4.03 & 4.52 & 4.78 & 4.35 & \textbf{0.615} \\
Opus-5 & No plan & .695 & .680 & \textbf{.688} & 2.91 & 2.93 & 4.29 & 4.46 & 3.65 & 0.568 \\
Opus-5 & Plan & .659 & .621 & .640 & 4.13 & 4.07 & 4.69 & 4.71 & 4.40 & 0.566 \\
Kimi-K3 & No plan & .462 & .648 & .555 & 2.83 & 2.87 & 4.30 & 4.39 & 3.60 & 0.510 \\
Kimi-K3 & Plan & .529 & .662 & .596 & 4.13 & 4.14 & 4.70 & 4.72 & 4.42 & 0.569 \\
Inkling & No plan & .646 & .493 & .569 & 2.59 & 2.69 & 4.19 & 4.21 & 3.42 & 0.433 \\
Inkling & Plan & .643 & .522 & .583 & 3.97 & 3.97 & 4.66 & 4.71 & 4.33 & 0.520 \\
Gemma-31B & No plan & .592 & .577 & .584 & 2.83 & 2.65 & 4.01 & 4.83 & 3.58 & 0.478 \\
\rowcolor{blue!8}Gemma-31B & Plan & \underline{.699} & .534 & .617 & 4.48 & 4.43 & 4.76 & 4.93 & 4.65 & \underline{0.597} \\
\rowcolor{blue!8}Gemma-31B & PWR & .602 & .594 & .598 & \underline{4.68} & 4.68 & \underline{4.84} & \textbf{4.95} & \underline{4.79} & 0.582 \\
Muse-30B & No plan & .502 & .396 & .449 & 2.57 & 2.71 & 3.32 & 1.84 & 2.61 & 0.290 \\
\rowcolor{blue!8}Muse-30B & Plan & .352 & .539 & .445 & 2.79 & 3.07 & 3.40 & 1.90 & 2.79 & 0.380 \\
\rowcolor{blue!8}Muse-30B & PWR & .347 & .670 & .509 & \underline{4.68} & \underline{4.75} & 4.28 & 2.29 & 4.00 & 0.533 \\
Qwen-27B & No plan & .486 & .413 & .450 & 2.52 & 2.57 & 4.06 & 4.82 & 3.49 & 0.354 \\
\rowcolor{blue!8}Qwen-27B & Plan & .577 & .404 & .491 & 4.40 & 4.36 & 4.74 & 4.91 & 4.60 & 0.472 \\
\rowcolor{blue!8}Qwen-27B & PWR & \textbf{.706} & .420 & .563 & \textbf{4.81} & \textbf{4.80} & \textbf{4.90} & \underline{4.94} & \textbf{4.87} & 0.594 \\
\bottomrule
\end{tabular*}
\caption{Per-dataset Turn and subjective results on EgoProactive.}
\label{tab:detail-egoproactive}
\end{table*}

\section{Per-Dataset Score Details}
% ============================================================

% \newpage
\section{More qualitative results}
\label{app:more_examples}

\Cref{fig:qual_silent,fig:qual_stepcomplete,fig:qual_oop_distraction,fig:qual_oop_mistake} show single-decision examples for each behavior regime, and \cref{fig:mt_gemini} traces full multi-turn sessions against frontier baselines.

\paragraph{Per-turn behavior.} On \emph{silent} timesteps, where the user is mid-step and no intervention is warranted, PWR withholds while baselines frequently interrupt (\cref{fig:qual_silent}). At genuine \emph{step boundaries} it interrupts with the correct next instruction rather than mistiming or mis-specifying the cue (\cref{fig:qual_stepcomplete}). The out-of-procedure cases are the most discriminative: under an external distraction PWR recognizes the deviation and re-engages with the task (\cref{fig:qual_oop_distraction}), and when the user makes a procedural mistake it both detects the error and names the specific corrective action, where baselines either miss it or offer generic advice (\cref{fig:qual_oop_mistake}).

\begin{figure*}[t]
% \vspace{-1cm}
\centering
% ------------------------------------------------------------------
% (A) SILENT TIMESTEPS — model correctly stays silent mid-step
% ------------------------------------------------------------------
\catbanner{colCat1}{(A)}{{\bf Silent Monitoring}: Step in Progress}
\setlength{\tabcolsep}{2pt}
\renewcommand{\arraystretch}{1.0}
\scriptsize
\catstripe{colCat1}{%
\begin{tabular}{@{}|>{\columncolor{colPlan}}p{0.25\linewidth}|>{\columncolor{colCue}}p{0.45\linewidth}|>{\columncolor{colGolden}}p{0.050\linewidth}|>{\columncolor{colOurs}}p{0.050\linewidth}|>{\columncolor{colThreeP}}p{0.20\linewidth}|@{}}
% --- Frames row (spans all 5 columns) ---
\multicolumn{5}{@{}c@{}}{\includegraphics[width=0.20\linewidth]{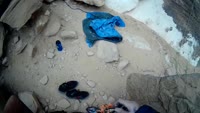} \includegraphics[width=0.20\linewidth]{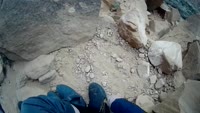} \includegraphics[width=0.20\linewidth]{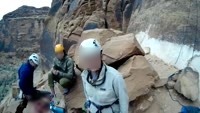} \includegraphics[width=0.20\linewidth]{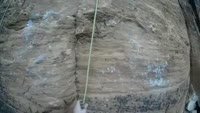} \includegraphics[width=0.20\linewidth]{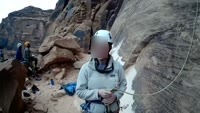}} \\
\hline
% --- Task (full-width) ---
\multicolumn{5}{|>{\columncolor{colTask}}p{\dimexpr \linewidth+6\tabcolsep+4\arrayrulewidth\relax}|}{\textbf{Task}: Ensure climbers are properly equipped and secured, and then ascend a rock face using proper climbing techniques and safety measures. \textit{[Source: \textbf{Ego4D}]}} \\
\hline
% --- Dialog history (full-width) ---
\multicolumn{5}{|>{\columncolor{colDialog}}p{\dimexpr \linewidth+6\tabcolsep+4\arrayrulewidth\relax}|}{\textbf{Dialog history}: \textbf{1.}\;Sure! Let's get started. First, Inspect and put on climbing shoes} \\
\hline
% --- Header row ---
\textbf{Plan (prev $\rightarrow$ curr)} & \textbf{Cue (\$ silent\$)} & \textbf{Golden} & \textbf{Ours}\,{\textcolor{darkgreen}{\cmark}} & \begin{adjustbox}{max width=\linewidth}\textbf{\texttt{gemini}}\,{\textcolor{red}{\xmark}}\end{adjustbox} \\
\hline
% --- Data row ---
\textit{Adjust and secure climbing harnesses and equipment} $\rightarrow$ \textbf{Tie a secure knot on the climbing rope} & Hands threading/looping rope, tails loose and moving, knot open or asymmetrical, fingers adjusting segments, attention on manipulation rather than inspection. & \textcolor{gray}{\textbf{[silent]}} & \textcolor{gray}{\textbf{[silent]}} & Step 2 is complete. Now, Ascend the rock face while ensuring safety
 \\
\hline
\end{tabular}%
}
% \end{figure*}

% \begin{figure*}[t]
\centering
\setlength{\tabcolsep}{2pt}
\renewcommand{\arraystretch}{1.0}
\scriptsize
\catstripe{colCat1}{%
\begin{tabular}{@{}|>{\columncolor{colPlan}}p{0.120\linewidth}|>{\columncolor{colCue}}p{0.260\linewidth}|>{\columncolor{colGolden}}p{0.070\linewidth}|>{\columncolor{colOurs}}p{0.070\linewidth}|>{\columncolor{colThreeP}}p{0.480\linewidth}|@{}}
% --- Frames row (spans all 5 columns) ---
\multicolumn{5}{@{}c@{}}{\includegraphics[width=0.20\linewidth]{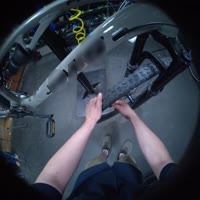} \includegraphics[width=0.20\linewidth]{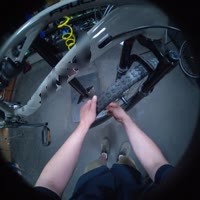} \includegraphics[width=0.20\linewidth]{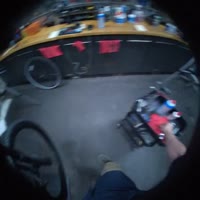} \includegraphics[width=0.20\linewidth]{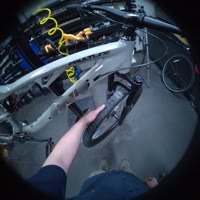} \includegraphics[width=0.20\linewidth]{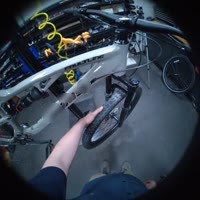}} \\
\hline
% --- Task (full-width) ---
\multicolumn{5}{|>{\columncolor{colTask}}p{\dimexpr 1.0000\linewidth+6\tabcolsep+4\arrayrulewidth\relax}|}{\textbf{Task}: Remove the front wheel from the bicycle for potential repair or replacement by deflating the tire, removing the through axle, and pulling the wheel off the fork. \textit{[Source: \textbf{EgoExo4D}]}} \\
\hline
% --- Dialog history (full-width) ---
\multicolumn{5}{|>{\columncolor{colDialog}}p{\dimexpr 1.0000\linewidth+6\tabcolsep+4\arrayrulewidth\relax}|}{\textbf{Dialog history}: } \\
\hline
% --- Header row ---
\textbf{Plan (prev $\rightarrow$ curr)} & \textbf{Cue (\$ silent\$)} & \textbf{Golden} & \textbf{Ours}\,{\textcolor{darkgreen}{\cmark}} & \textbf{\texttt{gpt}}\,{\textcolor{red}{\xmark}} \\
\hline
% --- Data row ---
\textbf{ $\rightarrow$ Unscrew the Presta valve to release air from the tire} & Hands rotating lock nut or depressing valve core, air visibly escaping, tire still firm, valve cap absent or partially removed. & \textcolor{gray}{\textbf{[silent]}} & \textcolor{gray}{\textbf{[silent]}} & You're skipping ahead—First, fully deflate the tire: remove the valve cap, unscrew the Presta valve tip a few turns, then press it in to release air until the tire is completely soft. Let me know when done and we'll move to the through axle. \\
\hline
\end{tabular}%
}
% \end{figure*}

% \begin{figure*}[t]
\centering
\setlength{\tabcolsep}{2pt}
\renewcommand{\arraystretch}{1.0}
\scriptsize
\catstripe{colCat1}{%
\begin{tabular}{@{}|>{\columncolor{colPlan}}p{0.185\linewidth}|>{\columncolor{colCue}}p{0.340\linewidth}|>{\columncolor{colGolden}}p{0.065\linewidth}|>{\columncolor{colOurs}}p{0.06\linewidth}|>{\columncolor{colThreeP}}p{0.35\linewidth}|@{}}
% --- Frames row (spans all 5 columns) ---
\multicolumn{5}{@{}c@{}}{\includegraphics[width=0.20\linewidth]{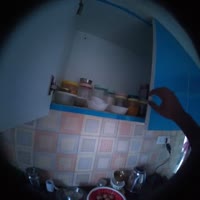} \includegraphics[width=0.20\linewidth]{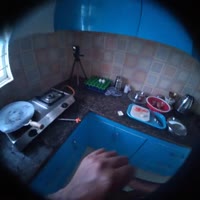} \includegraphics[width=0.20\linewidth]{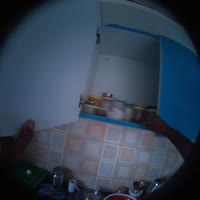} \includegraphics[width=0.20\linewidth]{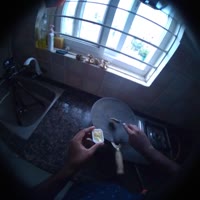} \includegraphics[width=0.20\linewidth]{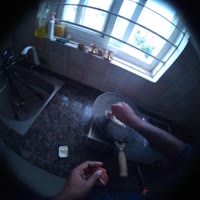}} \\
\hline
% --- Task (full-width) ---
\multicolumn{5}{|>{\columncolor{colTask}}p{\dimexpr \linewidth+6\tabcolsep+4\arrayrulewidth\relax}|}{\textbf{Task}: Prepare a meal by cracking eggs, mixing them with seasonings, cooking in a pan with butter, and cleaning up the workspace. \textit{[Source: \textbf{EgoExo4D}]}} \\
\hline
% --- Dialog history (full-width) ---
\multicolumn{5}{|>{\columncolor{colDialog}}p{\dimexpr \linewidth+6\tabcolsep+4\arrayrulewidth\relax}|}{\textbf{Dialog history}: \textbf{1.}\;Sure! Let's get started. First, Open the kitchen cabinet and retrieve a spice or seasoning container. \textbf{2.}\;Light the gas stove and heat a pan. \textbf{3.}\;Crack eggs into a bowl and add seasonings. \textbf{4.}\;Add butter to the heated pan and stir it.} \\
\hline
% --- Header row ---
\textbf{Plan (prev $\rightarrow$ curr)} & \textbf{Cue (\$ silent\$)} & \textbf{Golden} & \textbf{Ours}\,{\textcolor{darkgreen}{\cmark}} & \textbf{\texttt{claude}}\,{\textcolor{red}{\xmark}} \\
\hline
% --- Data row ---
\textit{Add butter to the heated pan and stir it.} $\rightarrow$ \textbf{Pour the egg mixture into the pan and cook.} & Active stirring or manipulation of the eggs \newline egg mixture is still being poured \newline spatula is actively engaged with the eggs \newline eggs are still visibly runny or not fully cooked & \textcolor{gray}{\textbf{[silent]}} & \textcolor{gray}{\textbf{[silent]}} & Good, it looks like you're pouring the egg mixture into the pan. Cook the eggs until they are fully set and no longer runny, stirring as needed. Once done, the next step will be to clean up the workspace. \\
\hline
\end{tabular}%
}
\centering
\setlength{\tabcolsep}{2pt}
\renewcommand{\arraystretch}{1.0}
\scriptsize
\catstripe{colCat1}{%
\begin{tabular}{@{}|>{\columncolor{colPlan}}p{0.250\linewidth}|>{\columncolor{colCue}}p{0.270\linewidth}|>{\columncolor{colGolden}}p{0.070\linewidth}|>{\columncolor{colOurs}}p{0.070\linewidth}|>{\columncolor{colThreeP}}p{0.340\linewidth}|@{}}
% --- Frames row (spans all 5 columns) ---
\multicolumn{5}{@{}c@{}}{\includegraphics[width=0.20\linewidth]{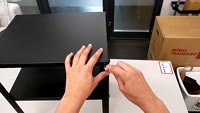} \includegraphics[width=0.20\linewidth]{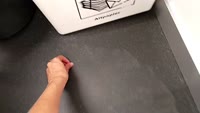} \includegraphics[width=0.20\linewidth]{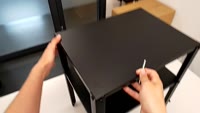} \includegraphics[width=0.20\linewidth]{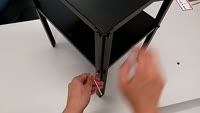} \includegraphics[width=0.20\linewidth]{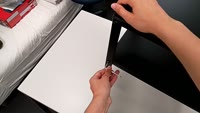}} \\
\hline
% --- Task (full-width) ---
\multicolumn{5}{|>{\columncolor{colTask}}p{\dimexpr 1.0000\linewidth+6\tabcolsep+4\arrayrulewidth\relax}|}{\textbf{Task}: Disassemble a nightstand. \textit{[Source: \textbf{HoloAssist}]}} \\
\hline
% --- Dialog history (full-width) ---
\multicolumn{5}{|>{\columncolor{colDialog}}p{\dimexpr 1.0000\linewidth+6\tabcolsep+4\arrayrulewidth\relax}|}{\textbf{Dialog history}: \textbf{1.}\;Sure! Let's get started. First, Unscrew the hex socket heads of the first leg. \textbf{2.}\;Unscrew the hex socket heads of the second leg.} \\
\hline
% --- Header row ---
\textbf{Plan (prev $\rightarrow$ curr)} & \textbf{Cue (\$ silent\$)} & \textbf{Golden} & \textbf{Ours}\,{\textcolor{darkgreen}{\cmark}} & \textbf{\texttt{gpt}}\,{\textcolor{red}{\xmark}} \\
\hline
% --- Data row ---
\textit{Unscrew the hex socket heads of the third leg.} $\rightarrow$ \textbf{Unscrew the hex socket heads of the fourth leg.} & Allen wrench actively turning a partially or fully seated screw, hand applying torque, leg still rigidly attached to frame. & \textcolor{gray}{\textbf{[silent]}} & \textcolor{gray}{\textbf{[silent]}} & You're on step 3 now: Finish unscrewing the hex screw(s) on the fourth leg counterclockwise until fully loose, then confirm the leg wiggles free. Next we'll remove the top shelf screws. \\
\hline
\end{tabular}%
}
\captionsetup{skip=2pt,font=footnotesize,labelsep=period,labelfont=bf}\caption{Qualitative examples: silent monitoring.}
\label{fig:qual_silent}
\end{figure*}

\begin{figure*}[t]
\ContinuedFloat
\centering
\setlength{\tabcolsep}{2pt}
\renewcommand{\arraystretch}{1.0}
\scriptsize
\catstripe{colCat1}{%
\begin{tabular}{@{}|>{\columncolor{colPlan}}p{0.290\linewidth}|>{\columncolor{colCue}}p{0.19\linewidth}|>{\columncolor{colGolden}}p{0.065\linewidth}|>{\columncolor{colOurs}}p{0.06\linewidth}|>{\columncolor{colThreeP}}p{0.395\linewidth}|@{}}
% --- Frames row (spans all 5 columns; series label sticks out to the left) ---
\multicolumn{5}{@{}c@{}}{
% \makebox[0pt][r]{\raisebox{5\height}{\large\bfseries 1h}\hspace{0.1cm}
%}
\includegraphics[width=0.20\linewidth]{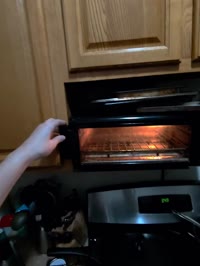} \includegraphics[width=0.20\linewidth]{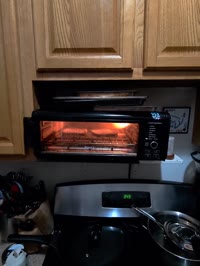} \includegraphics[width=0.20\linewidth]{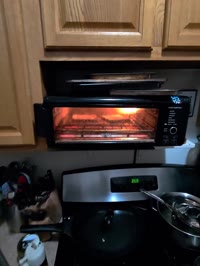} \includegraphics[width=0.20\linewidth]{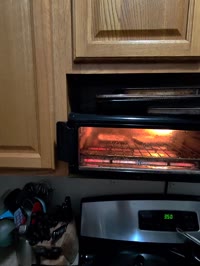} \includegraphics[width=0.20\linewidth]{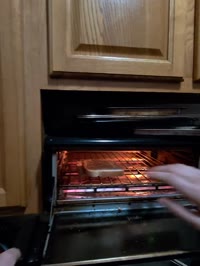}} \\
\hline
% --- Task (full-width) ---
\multicolumn{5}{|>{\columncolor{colTask}}p{\dimexpr 1.0000\linewidth+6\tabcolsep+4\arrayrulewidth\relax}|}{\textbf{Task}: Making Toast with Almond Butter and Honey. \textit{[Source: \textbf{EgoProactive}]}} \\
\hline
% --- Dialog history (full-width) ---
\multicolumn{5}{|>{\columncolor{colDialog}}p{\dimexpr 1.0000\linewidth+6\tabcolsep+4\arrayrulewidth\relax}|}{\textbf{Dialog history}: \textbf{1.}\;Sure! Let's get started. First, Open the bread bag and take out two slices of bread.  \textbf{2.}\;Place the two bread slices in the toaster oven, set the timer, and toast until lightly browned.} \\
\hline
% --- Header row ---
\textbf{Plan (prev $\rightarrow$ curr)} & \textbf{Cue (\$ silent\$)} & \textbf{Golden} & \textbf{Ours}\,{\textcolor{darkgreen}{\cmark}} & \textbf{\texttt{gpt}}\,{\textcolor{red}{\xmark}} \\
\hline
% --- Data row ---
\textit{Open the toaster oven, remove the toasted bread slices, and set them aside.} $\rightarrow$ \textbf{Place the toasted slices on a cutting board.} & % hands are actively holding or adjusting toast slices \newline one or more slices remain elevated above the board \newline visible movement indicates placement is still happening \newline the full set of slices has not yet been settled on the board
Hands actively placing toast slices, still in motion, not yet settled on the board. & \textcolor{gray}{\textbf{[silent]}} & \textcolor{gray}{\textbf{[silent]}} & Nice—both toasted slices are now flat on the cutting board. Next: open the almond butter jar, set the lid aside, and use a knife to scoop a small amount of almond butter. \\
\hline
\end{tabular}%
}
% \end{figure*}

% \begin{figure*}[t]
% \vspace{-1cm}
\centering
\setlength{\tabcolsep}{2pt}
\renewcommand{\arraystretch}{1.0}
\scriptsize
\catstripe{colCat1}{%
\begin{tabular}{@{}|>{\columncolor{colPlan}}p{0.300\linewidth}|>{\columncolor{colCue}}p{0.350\linewidth}|>{\columncolor{colGolden}}p{0.070\linewidth}|>{\columncolor{colOurs}}p{0.070\linewidth}|>{\columncolor{colThreeP}}p{0.210\linewidth}|@{}}
% --- Frames row (spans all 5 columns; series label sticks out to the left) ---
\multicolumn{5}{@{}c@{}}{
%\makebox[0pt][r]{\large\bfseries 1i\hspace{0.4cm}}
\includegraphics[width=0.20\linewidth]{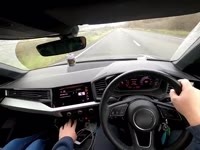} \includegraphics[width=0.20\linewidth]{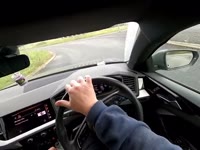} \includegraphics[width=0.20\linewidth]{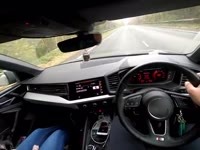} \includegraphics[width=0.20\linewidth]{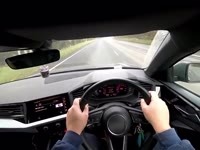} \includegraphics[width=0.20\linewidth]{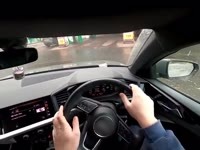}} \\
\hline
% --- Task (full-width) ---
\multicolumn{5}{|>{\columncolor{colTask}}p{\dimexpr 1.0000\linewidth+6\tabcolsep+4\arrayrulewidth\relax}|}{\textbf{Task}: Navigate the car from the starting point on the highway to a gas station. \textit{[Source: \textbf{Ego4D}]}} \\
\hline
% --- Dialog history (full-width) ---
\multicolumn{5}{|>{\columncolor{colDialog}}p{\dimexpr 1.0000\linewidth+6\tabcolsep+4\arrayrulewidth\relax}|}{\textbf{Dialog history}: \textbf{1.}\;Sure! Let's get started. First, Drive forward on a two-lane highway with both hands on the steering wheel.  \textbf{2.}\;Turn the steering wheel to the left to navigate a curve.  \textbf{3.}\;Struggle with the steering wheel, making adjustments to its position.  \textbf{4.}\;Drive on the highway, changing gears and steering smoothly.} \\
\hline
% --- Header row ---
\textbf{Plan (prev $\rightarrow$ curr)} & \textbf{Cue (\$ silent\$)} & \textbf{Golden} & \textbf{Ours}\,{\textcolor{darkgreen}{\cmark}} & \textbf{\texttt{qwen3 235B}} \\
\hline
% --- Data row ---
\textit{Slow down and make adjustments to the steering wheel.} $\rightarrow$ \textbf{Drive into a parking lot and turn the steering wheel clockwise.} & Hands gripping and rotating steering wheel clockwise, arms tense and in motion, motion blur or transitional scenery visible through windshield. & \textcolor{gray}{\textbf{[silent]}} & \textcolor{gray}{\textbf{[silent]}} & You have successfully navigated to the gas station. Park the car and turn off the engine. \\
\hline
\end{tabular}%
}
% \end{figure*}

% \begin{figure*}[t]
\centering
\setlength{\tabcolsep}{2pt}
\renewcommand{\arraystretch}{1.0}
\scriptsize
\catstripe{colCat1}{%
\begin{tabular}{@{}|>{\columncolor{colPlan}}p{0.155\linewidth}|>{\columncolor{colCue}}p{0.270\linewidth}|>{\columncolor{colGolden}}p{0.065\linewidth}|>{\columncolor{colOurs}}p{0.060\linewidth}|>{\columncolor{colThreeP}}p{0.45\linewidth}|@{}}
% --- Frames row (spans all 5 columns; series label sticks out to the left, vertically centered with the images) ---
\multicolumn{5}{@{}c@{}}{
%\makebox[0pt][r]{\raisebox{0.35in}{\large\bfseries 1j}\hspace{0.4cm}}
\includegraphics[width=0.20\linewidth]{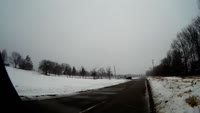} \includegraphics[width=0.20\linewidth]{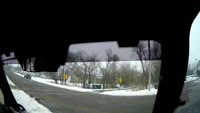} \includegraphics[width=0.20\linewidth]{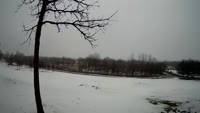} \includegraphics[width=0.20\linewidth]{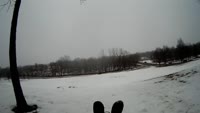} \includegraphics[width=0.20\linewidth]{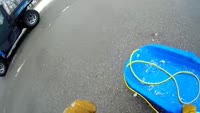}} \\
\hline
% --- Task (full-width) ---
\multicolumn{5}{|>{\columncolor{colTask}}p{\dimexpr 1.0000\linewidth+6\tabcolsep+4\arrayrulewidth\relax}|}{\textbf{Task}: Drive a UTV through snowy roads, pull a blue sled through a snowy field, slide down a snowy hill on the sled, load the sled onto the UTV, and transport it to the final destination. \textit{[Source: \textbf{Ego4D}]}} \\
\hline
% --- Dialog history (full-width) ---
\multicolumn{5}{|>{\columncolor{colDialog}}p{\dimexpr 1.0000\linewidth+6\tabcolsep+4\arrayrulewidth\relax}|}{\textbf{Dialog history}: \textbf{1.}\;Sure! Let's get started. First, Drive the UTV through snowy residential areas and stop at a specific location.  \textbf{2.}\;Exit the UTV and pull the blue sled through a snowy field.} \\
\hline
% --- Header row ---
\textbf{Plan (prev $\rightarrow$ curr)} & \textbf{Cue (\$ silent\$)} & \textbf{Golden} & \textbf{Ours}\,{\textcolor{darkgreen}{\cmark}} & \textbf{\texttt{gpt}}\,{\textcolor{red}{\xmark}} \\
\hline
% --- Data row ---
\textit{Slide down a snowy hill on the blue sled.} $\rightarrow$ \textbf{Load the blue sled onto the UTV.} & %Sled remains in contact with ground surface \newline Hands actively gripping or lifting sled edges \newline Person bent at waist with arms extended toward sled \newline Sled partially lifted or tilted at angle \newline Vehicle stationary while sled not yet moved from ground 
Hands gripping or lifting sled edges, person bent at waist, sled partially tilted but still in contact with ground, vehicle stationary. & \textcolor{gray}{\textbf{[silent]}} & \textcolor{gray}{\textbf{[silent]}} & You're at the UTV with the blue sled visible on your right—before sliding, you need to pull the sled out into the snowy field and head toward the hill. Grab the sled's rope/handle and start walking it across the snow toward the slope. \\
\hline
\end{tabular}%
}
% \end{figure*}

% \begin{figure*}[t]
\centering
\setlength{\tabcolsep}{2pt}
\renewcommand{\arraystretch}{1.0}
\scriptsize
\catstripe{colCat1}{%
\begin{tabular}{@{}|>{\columncolor{colPlan}}p{0.150\linewidth}|>{\columncolor{colCue}}p{0.180\linewidth}|>{\columncolor{colGolden}}p{0.070\linewidth}|>{\columncolor{colOurs}}p{0.070\linewidth}|>{\columncolor{colThreeP}}p{0.530\linewidth}|@{}}
% --- Frames row (spans all 5 columns; series label sticks out to the left) ---
\multicolumn{5}{@{}c@{}}{
%\makebox[0pt][r]{\large\bfseries 1k\hspace{0.4cm}}
\includegraphics[width=0.20\linewidth]{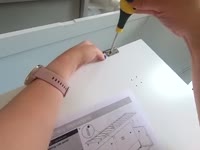} \includegraphics[width=0.20\linewidth]{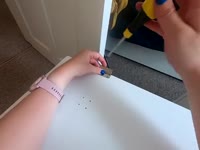} \includegraphics[width=0.20\linewidth]{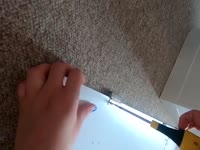} \includegraphics[width=0.20\linewidth]{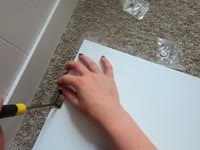} \includegraphics[width=0.20\linewidth]{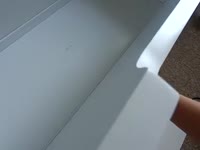}} \\
\hline
% --- Task (full-width) ---
\multicolumn{5}{|>{\columncolor{colTask}}p{\dimexpr 1.0000\linewidth+6\tabcolsep+4\arrayrulewidth\relax}|}{\textbf{Task}: Assemble a white piece of furniture by attaching hinges, metal brackets, and screws, and organize instructions and components to ensure all parts are properly secured and functional. \textit{[Source: \textbf{Ego4D}]}} \\
\hline
% --- Dialog history (full-width) ---
\multicolumn{5}{|>{\columncolor{colDialog}}p{\dimexpr 1.0000\linewidth+6\tabcolsep+4\arrayrulewidth\relax}|}{\textbf{Dialog history}: \textbf{1.}\;Sure! Let's get started. First, Screw a hinge onto a white piece of furniture using a screwdriver.  \textbf{2.}\;Place a metal clip on the cabinet and tighten the screws.  \textbf{3.}\;Attach metal attachments to the drawer.  \textbf{4.}\;Assemble the sides of the white piece of furniture together.  \textbf{5.}\;Screw screws into the corner of a white cabinet door.  \textbf{6.}\;Attach a metal bracket to a white wardrobe door.  \textbf{7.}\;Fix a door's hinge.  \textbf{8.}\;Screw black brackets into the corners of a white object.} \\
\hline
% --- Header row ---
\textbf{Plan (prev $\rightarrow$ curr)} & \textbf{Cue (\$ silent\$)} & \textbf{Golden} & \textbf{Ours}\,{\textcolor{darkgreen}{\cmark}} & \textbf{\texttt{gpt}}\,{\textcolor{red}{\xmark}} \\
\hline
% --- Data row ---
\textit{Remove screws from a white surface.} $\rightarrow$ \textbf{Attach metal brackets to the underside of a white piece of wood.} & % Hand actively gripping bracket or screwdriver \newline bracket held at an angle with visible space between bracket and wood \newline screws protruding or partially inserted \newline tool in motion near bracket \newline bracket not yet aligned with mounting points
Hand gripping bracket or screwdriver, bracket angled with gap from wood, screws partially inserted, bracket not yet aligned with mounting points. & \textcolor{gray}{\textbf{[silent]}} & \textcolor{gray}{\textbf{[silent]}} & It looks like you've removed the screws and set the panel down. Next, do Step 9: flip the white board so the underside is facing up, align the metal brackets with the pre-drilled holes, and screw them in flush (no gaps, screw heads level). If you don't see pilot holes, stop and check the instruction sheet to confirm you're on the correct side before drilling or forcing screws. \\
\hline
\end{tabular}%
}
\captionsetup{skip=2pt,font=footnotesize,labelsep=period,labelfont=bf}\caption{Silent monitoring (continued).}
\end{figure*}

\begin{figure*}[t]
\centering
\setlength{\tabcolsep}{2pt}
\renewcommand{\arraystretch}{1.0}
\scriptsize
\catstripe{colCat1}{%
\begin{tabular}{@{}|>{\columncolor{colPlan}}p{0.195\linewidth}|>{\columncolor{colCue}}p{0.185\linewidth}|>{\columncolor{colGolden}}p{0.060\linewidth}|>{\columncolor{colOurs}}p{0.060\linewidth}|>{\columncolor{colThreeP}}p{0.500\linewidth}|@{}}
% --- Frames row (spans all 5 columns; series label sticks out to the left) ---
\multicolumn{5}{@{}c@{}}{
%\makebox[0pt][r]{\large\bfseries 1l\hspace{0.4cm}}
\includegraphics[width=0.20\linewidth]{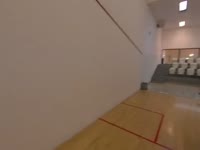} \includegraphics[width=0.20\linewidth]{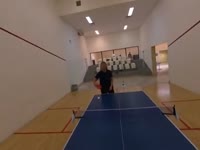} \includegraphics[width=0.20\linewidth]{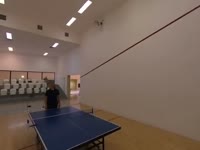} \includegraphics[width=0.20\linewidth]{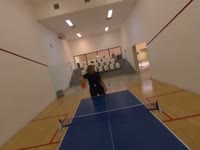} \includegraphics[width=0.20\linewidth]{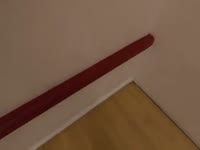}} \\
\hline
% --- Task (full-width) ---
\multicolumn{5}{|>{\columncolor{colTask}}p{\dimexpr 0.9900\linewidth+6\tabcolsep+4\arrayrulewidth\relax}|}{\textbf{Task}: Engage in a game of table tennis and then transition to a different activity by concluding the game and exploring the surroundings. \textit{[Source: \textbf{Ego4D}]}} \\
\hline
% --- Dialog history (full-width) ---
\multicolumn{5}{|>{\columncolor{colDialog}}p{\dimexpr 0.9900\linewidth+6\tabcolsep+4\arrayrulewidth\relax}|}{\textbf{Dialog history}: \textbf{1.}\;Sure! Let's get started. First, Serve the ball by bouncing it on the table before hitting it to the opponent's side.} \\
\hline
% --- Header row ---
\textbf{Plan (prev $\rightarrow$ curr)} & \textbf{Cue (\$ silent\$)} & \textbf{Golden} & \textbf{Ours}\,{\textcolor{darkgreen}{\cmark}} & \textbf{\texttt{claude}}\,{\textcolor{red}{\xmark}} \\
\hline
% --- Data row ---
\textit{Hit the ball back and forth across the table to engage in a rally.} $\rightarrow$ \textbf{Continue playing with consistent rallies.} & %Paddle actively swinging to strike ball \newline Ball in mid-air trajectory between players \newline Players positioned at table edges with ready stances \newline Eyes focused on ball movement \newline Repetitive hitting motions occurring 
Paddle mid-swing in hitting motion, ball in flight between players, both in ready stance with eyes tracking the ball. & \textcolor{gray}{\textbf{[silent]}} & \textcolor{gray}{\textbf{[silent]}} & It looks like you've stopped playing and moved away from the table (no ball/paddle action visible). Go ahead and conclude the game (say "good game," stop the rally, and set the paddle/ball down), then transition to exploring by walking out of the court area and looking around the nearby indoor spaces (hallway/lounge/adjacent rooms). \\
\hline
\end{tabular}%
}
% \end{figure*}

% \begin{figure*}[t]
\centering
\setlength{\tabcolsep}{2pt}
\renewcommand{\arraystretch}{1.0}
\scriptsize
\catstripe{colCat1}{%
\begin{tabular}{@{}|>{\columncolor{colPlan}}p{0.170\linewidth}|>{\columncolor{colCue}}p{0.270\linewidth}|>{\columncolor{colGolden}}p{0.050\linewidth}|>{\columncolor{colOurs}}p{0.050\linewidth}|>{\columncolor{colThreeP}}p{0.460\linewidth}|@{}}
% --- Frames row (spans all 5 columns; series label sticks out to the left) ---
\multicolumn{5}{@{}c@{}}{
%\makebox[0pt][r]{\large\bfseries 1n\hspace{0.4cm}}
\includegraphics[width=0.20\linewidth]{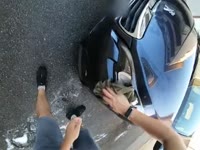} \includegraphics[width=0.20\linewidth]{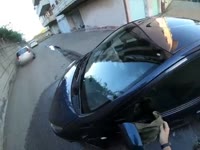} \includegraphics[width=0.20\linewidth]{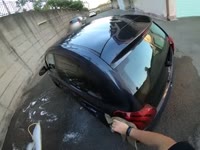} \includegraphics[width=0.20\linewidth]{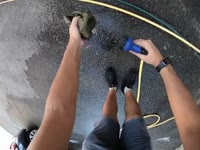} \includegraphics[width=0.20\linewidth]{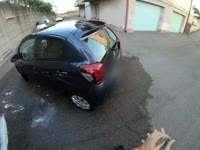}} \\
\hline
% --- Task (full-width) ---
\multicolumn{5}{|>{\columncolor{colTask}}p{\dimexpr 1.0000\linewidth+6\tabcolsep+4\arrayrulewidth\relax}|}{\textbf{Task}: Clean the car's exterior, windows, and mirrors thoroughly, and maintain the surrounding area's cleanliness. \textit{[Source: \textbf{Ego4D}]}} \\
\hline
% --- Dialog history (full-width) ---
\multicolumn{5}{|>{\columncolor{colDialog}}p{\dimexpr 1.0000\linewidth+6\tabcolsep+4\arrayrulewidth\relax}|}{\textbf{Dialog history}: \textbf{1.}\;Sure! Let's get started. First, Clean the car's hood and headlights using a cloth held in one hand, moving it across the surface to remove dirt and grime.  \textbf{2.}\;Clean the car's sides and windows using a larger towel, scrubbing from left to right.  \textbf{3.}\;Dry the car using a green cloth, starting from the rear left quarter window, moving to the tail light, and then the rear windshield.  \textbf{4.}\;Rinse a towel using a power washer, holding the green towel in one hand and the blue power washer in the other.} \\
\hline
% --- Header row ---
\textbf{Plan (prev $\rightarrow$ curr)} & \textbf{Cue (\$ silent\$)} & \textbf{Golden} & \textbf{Ours}\,{\textcolor{darkgreen}{\cmark}} & \textbf{\texttt{gpt}}\,{\textcolor{red}{\xmark}} \\
\hline
% --- Data row ---
\textit{Clean the ground using the power washer and towel.} $\rightarrow$ \textbf{Clean the car's windows and mirrors using a towel.} & 
Hands wipe towel across glass, continuous towel contact, motion blur from rubbing or circular strokes, water droplets or soap film still present, person facing glass with focused gaze
% hands are actively wiping the towel across windows or mirrors \newline towel maintains continuous contact with glass surfaces \newline visible motion blur indicating rubbing or circular cleaning motions \newline windows or mirrors still exhibit water droplets or soap film \newline person remains positioned directly in front of glass with focused gaze 
& \textcolor{gray}{\textbf{[silent]}} & \textcolor{gray}{\textbf{[silent]}} & You're holding the towel but it's not on the glass yet. Next, wipe the windows and mirrors: start at the driver-side window, then rear side window, then rear window, and finish with both side mirrors. Use firm overlapping strokes (top-to-bottom, then a quick left-to-right pass) until there are no droplets or streaks. \\
\hline
\end{tabular}%
}
% \end{figure*}

% \begin{figure*}
% \vspace{-1cm}
\centering
\setlength{\tabcolsep}{2pt}
\renewcommand{\arraystretch}{1.0}
\scriptsize
\catstripe{colCat1}{%
\begin{tabular}{@{}|>{\columncolor{colPlan}}p{0.230\linewidth}|>{\columncolor{colCue}}p{0.230\linewidth}|>{\columncolor{colGolden}}p{0.065\linewidth}|>{\columncolor{colOurs}}p{0.065\linewidth}|>{\columncolor{colThreeP}}p{0.410\linewidth}|@{}}
% --- Frames row (spans all 5 columns; series label sticks out to the left) ---
\multicolumn{5}{@{}c@{}}{
%\makebox[0pt][r]{\large\bfseries 1m\hspace{0.4cm}}
\includegraphics[width=0.20\linewidth]{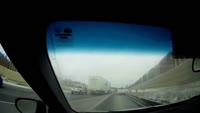} \includegraphics[width=0.20\linewidth]{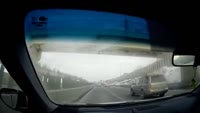} \includegraphics[width=0.20\linewidth]{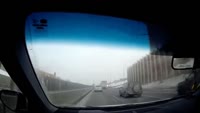} \includegraphics[width=0.20\linewidth]{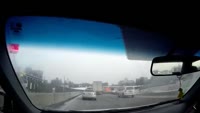} \includegraphics[width=0.20\linewidth]{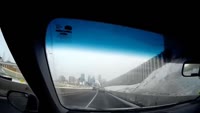}} \\
\hline
% --- Task (full-width) ---
\multicolumn{5}{|>{\columncolor{colTask}}p{\dimexpr \linewidth+6\tabcolsep+4\arrayrulewidth\relax}|}{\textbf{Task}: Safely navigate through various traffic and weather conditions on the highway while making progress on the journey. \textit{[Source: \textbf{Ego4D}]}} \\
\hline
% --- Dialog history (full-width) ---
\multicolumn{5}{|>{\columncolor{colDialog}}p{\dimexpr \linewidth+6\tabcolsep+4\arrayrulewidth\relax}|}{\textbf{Dialog history}: \textbf{1.}\;Sure! Let's get started. First, Accelerate and change lanes to overtake slow-moving vehicles on the highway  \textbf{2.}\;Navigate through traffic by changing lanes and adjusting speed to maintain safe distance} \\
\hline
% --- Header row ---
\textbf{Plan (prev $\rightarrow$ curr)} & \textbf{Cue (\$ silent\$)} & \textbf{Golden} & \textbf{Ours}\,{\textcolor{darkgreen}{\cmark}} & \textbf{\texttt{claude}}\,{\textcolor{red}{\xmark}} \\
\hline
% --- Data row ---
\textit{Drive through construction areas and cityscapes while adjusting speed and lane position} $\rightarrow$ \textbf{Adjust speed and maintain safe distance while driving under snowy conditions} & %Rapid closure rate between ego vehicle and lead vehicle \newline visible brake lights on lead vehicle with ego vehicle not decelerating \newline erratic steering corrections or lane position changes \newline hands actively manipulating steering wheel with urgency \newline significant speed differential between ego vehicle and surrounding traffic 
Rapid closure on lead vehicle, or lead vehicle braking while ego vehicle not decelerating, or urgent steering corrections, or significant speed differential from surrounding traffic. & \textcolor{gray}{\textbf{[silent]}} & \textcolor{gray}{\textbf{[silent]}} & Good progress. You're currently driving under snowy/wet conditions approaching the city skyline. Maintain a safe following distance from the vehicle ahead and reduce speed as needed for the road conditions. Next step: Navigate through traffic under bridges and overpasses while maintaining safe distance. \\
\hline
\end{tabular}%
}
% \end{figure*}

% ------------------------------------------------------------------
% (B) STEP COMPLETION INTERRUPTS
% ------------------------------------------------------------------
% \begin{figure*}[t]
\centering
% ------------------------------------------------------------------
% (B) STEP COMPLETION INTERRUPTS — model fires when a step finishes
% ------------------------------------------------------------------
\catbanner{colCat2}{(B)}{{\bf Step Completion}: Proactive Interrupt}
\setlength{\tabcolsep}{2pt}
\renewcommand{\arraystretch}{1.0}
\scriptsize
\catstripe{colCat2}{%
\begin{tabular}{@{}|>{\columncolor{colPlan}}p{0.363\linewidth}|>{\columncolor{colCue}}p{0.173\linewidth}|>{\columncolor{colGolden}}p{0.193\linewidth}|>{\columncolor{colOurs}}p{0.212\linewidth}|>{\columncolor{colThreeP}}p{0.060\linewidth}|@{}}
% --- Frames row (spans all 5 columns; series label sticks out to the left) ---
\multicolumn{5}{@{}c@{}}{
% \makebox[0pt][r]{\large\bfseries 2a\hspace{0.4cm}}
\includegraphics[width=0.20\linewidth]{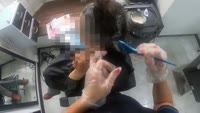} \includegraphics[width=0.20\linewidth]{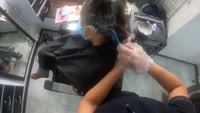} \includegraphics[width=0.20\linewidth]{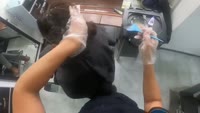} \includegraphics[width=0.20\linewidth]{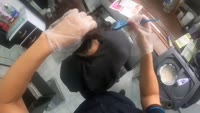} \includegraphics[width=0.20\linewidth]{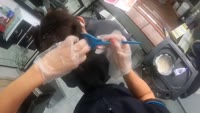}} \\
\hline
% --- Task (full-width) ---
\multicolumn{5}{|>{\columncolor{colTask}}p{\dimexpr 1.0010\linewidth+6\tabcolsep+4\arrayrulewidth\relax}|}{\textbf{Task}: Achieve the desired hair color and style for the clients by applying dye and product. \textit{[Source: \textbf{Ego4D}]}} \\
\hline
% --- Dialog history (full-width) ---
\multicolumn{5}{|>{\columncolor{colDialog}}p{\dimexpr 1.0010\linewidth+6\tabcolsep+4\arrayrulewidth\relax}|}{\textbf{Dialog history}: ---} \\
\hline
% --- Header row ---
\textbf{Plan (prev $\rightarrow$ curr)} & \textbf{Cue (\$ interrupt\$)} & \textbf{Golden} & \textbf{Ours}\,{\textcolor{darkgreen}{\cmark}} & \textbf{\texttt{gpt}}\,{\textcolor{red}{\xmark}} \\
\hline
% --- Data row ---
\textit{Using the blue brush, dip into the dye container and work the dye through the woman's hair section by section, using the free hand to pull and adjust for even coverage.} $\rightarrow$ \textbf{Keep working dye through the woman's hair using the blue brush, re-dipping into the container as needed and using the free hand to section the hair for full coverage.} & % Brush resting away from hair and dye container \newline hands withdrawn from manipulating hair sections \newline hair shows uniform dye saturation with no untreated strands visible \newline dye container not being accessed with tool \newline stylist's posture indicates transition to next task
Brush set aside and not in dye container, hands withdrawn, hair fully and evenly saturated, stylist posture shifting to next task. & 
Keep working dye through the woman's hair using the blue brush, re-dipping into the container as needed and using the free hand to section the hair for full coverage. & Continue applying dye to the woman's hair with the blue brush, ensuring thorough coverage by dipping the brush into the dye container multiple times and sectioning the hair with the other hand. & \textcolor{gray}{\textbf{[silent]}} \\
\hline
\end{tabular}%
}
\captionsetup{skip=2pt,font=footnotesize,labelsep=period,labelfont=bf}\caption{Qualitative examples: step-completion interrupts.}
\label{fig:qual_stepcomplete}
\end{figure*}

\begin{figure*}[t]
\ContinuedFloat
\centering
\setlength{\tabcolsep}{2pt}
\renewcommand{\arraystretch}{1.0}
\scriptsize
\catstripe{colCat2}{%
\begin{tabular}{@{}|>{\columncolor{colPlan}}p{0.40\linewidth}|>{\columncolor{colCue}}p{0.25\linewidth}|>{\columncolor{colGolden}}p{0.15\linewidth}|>{\columncolor{colOurs}}p{0.15\linewidth}|>{\columncolor{colThreeP}}p{0.050\linewidth}|@{}}
% --- Frames row (spans all 5 columns; series label sticks out to the left) ---
\multicolumn{5}{@{}c@{}}{
%\makebox[0pt][r]{\large\bfseries 2b\hspace{0.4cm}}
\includegraphics[width=0.20\linewidth]{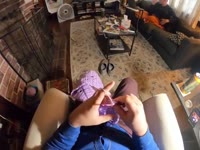} \includegraphics[width=0.20\linewidth]{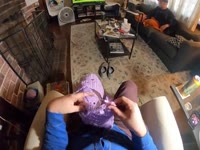} \includegraphics[width=0.20\linewidth]{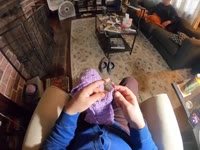} \includegraphics[width=0.20\linewidth]{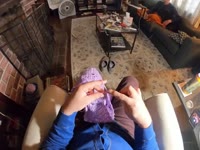} \includegraphics[width=0.20\linewidth]{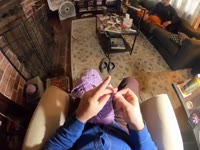}} \\
\hline
% --- Task (full-width) ---
\multicolumn{5}{|>{\columncolor{colTask}}p{\dimexpr 1.0000\linewidth+6\tabcolsep+4\arrayrulewidth\relax}|}{\textbf{Task}: Knit a purple item using two knitting needles and purple yarn while sitting comfortably in a chair, taking breaks as necessary, and completing the item to the desired specifications. \textit{[Source: \textbf{Ego4D}]}} \\
\hline
% --- Dialog history (full-width) ---
\multicolumn{5}{|>{\columncolor{colDialog}}p{\dimexpr 1.0000\linewidth+6\tabcolsep+4\arrayrulewidth\relax}|}{\textbf{Dialog history}: ---} \\
\hline
% --- Header row ---
\textbf{Plan (prev $\rightarrow$ curr)} & \textbf{Cue (\$ interrupt\$)} & \textbf{Golden} & \textbf{Ours}\,{\textcolor{darkgreen}{\cmark}} & \textbf{\texttt{qwen}}\,{\textcolor{red}{\xmark}} \\
\hline
% --- Data row ---
\textit{Settle into the white chair with legs crossed, pick up the knitting needles in both hands, and get ready to knit the purple cloth.} $\rightarrow$ \textbf{Tighten the yarn on the knitting needle until it sits snug.} & % Yarn appears taut with no visible slack \newline hands are stationary or moving away from the needle \newline knitting needle is held steady without active manipulation \newline yarn is neatly wrapped and positioned on the needle \newline fingers are relaxed and not engaged in tension adjustment
Yarn taut and neatly wrapped on needle, hands stationary or withdrawing, fingers relaxed with no active tension adjustment. & 
Tighten the yarn on the knitting needle until it sits snug. & Adjust the yarn on the knitting needle to ensure it is tight. & \textcolor{gray}{\textbf{[silent]}} \\
\hline
\end{tabular}%
}
% \end{figure*}

% \begin{figure*}[t]
\centering
\setlength{\tabcolsep}{2pt}
\renewcommand{\arraystretch}{1.0}
\scriptsize
\catstripe{colCat2}{%
\begin{tabular}{@{}|>{\columncolor{colPlan}}p{0.22\linewidth}|>{\columncolor{colCue}}p{0.19\linewidth}|>{\columncolor{colGolden}}p{0.12\linewidth}|>{\columncolor{colOurs}}p{0.12\linewidth}|>{\columncolor{colThreeP}}p{0.35\linewidth}|@{}}
% --- Frames row (spans all 5 columns; series label sticks out to the left) ---
\multicolumn{5}{@{}c@{}}{
%\makebox[0pt][r]{\large\bfseries 3a\hspace{0.4cm}}
\includegraphics[width=0.20\linewidth]{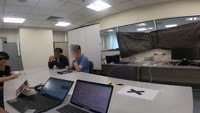} \includegraphics[width=0.20\linewidth]{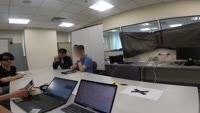} \includegraphics[width=0.20\linewidth]{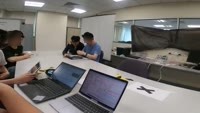} \includegraphics[width=0.20\linewidth]{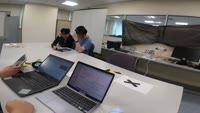} \includegraphics[width=0.20\linewidth]{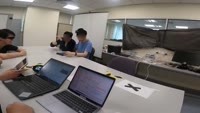}} \\
\hline
% --- Task (full-width) ---
\multicolumn{5}{|>{\columncolor{colTask}}p{\dimexpr 1.0000\linewidth+6\tabcolsep+4\arrayrulewidth\relax}|}{\textbf{Task}: Achieve a collaborative and immersive discussion among group members using laptops and virtual reality headsets, and successfully conclude the discussion. \textit{[Source: \textbf{Ego4D}]}} \\
\hline
% --- Dialog history (full-width) ---
\multicolumn{5}{|>{\columncolor{colDialog}}p{\dimexpr 1.0000\linewidth+6\tabcolsep+4\arrayrulewidth\relax}|}{\textbf{Dialog history}: ---} \\
\hline
% --- Header row ---
\textbf{Plan (prev $\rightarrow$ curr)} & \textbf{Cue (\$ interrupt\$)} & \textbf{Golden} & \textbf{Ours}\,{\textcolor{darkgreen}{\cmark}} & \textbf{\texttt{gpt}}\,{\textcolor{darkgreen}{\cmark}} \\
\hline
% --- Data row ---
\textit{Jump into the collaborative discussion, working with laptops and VR headsets.} $\rightarrow$ \textbf{Keep the collaborative project discussion going, working across laptops and VR headsets.} & % Participants removing headsets and placing them on the table \newline Laptops closed or screens displaying inactive states \newline Group members leaning back with relaxed posture and hands disengaged from devices \newline Participants gathering personal belongings or shifting attention away from shared workspace \newline Eye contact and body orientation turning away from collaborative focus toward individual departure 
Headsets coming off, laptops closing, participants leaning back and disengaging from devices, attention shifting away from the shared workspace. & 
Keep the collaborative project discussion going, working across laptops and VR headsets. & Continue the collaborative project discussion, utilizing laptops and virtual reality headsets. & Next step: assign roles and confirm everyone's setup.
1) Pick a facilitator and a note-taker/timekeeper.
2) Have each person open the shared doc/agenda on their laptop and confirm Wi‐Fi/audio works.
3) If you're using VR, distribute headsets now and do a quick fit + pairing check (one person at a time). \\
\hline
\end{tabular}%
}
% \end{figure*}

% \begin{figure*}[t]
\centering
\setlength{\tabcolsep}{2pt}
\renewcommand{\arraystretch}{1.0}
\scriptsize
\catstripe{colCat2}{%
\begin{tabular}{@{}|>{\columncolor{colPlan}}p{0.373\linewidth}|>{\columncolor{colCue}}p{0.297\linewidth}|>{\columncolor{colGolden}}p{0.148\linewidth}|>{\columncolor{colOurs}}p{0.132\linewidth}|>{\columncolor{colThreeP}}p{0.050\linewidth}|@{}}
% --- Frames row (spans all 5 columns; series label sticks out to the left) ---
\multicolumn{5}{@{}c@{}}{
% \makebox[0pt][r]{\large\bfseries 3b\hspace{0.4cm}}
\includegraphics[width=0.20\linewidth]{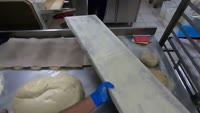} \includegraphics[width=0.20\linewidth]{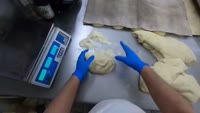} \includegraphics[width=0.20\linewidth]{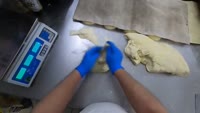} \includegraphics[width=0.20\linewidth]{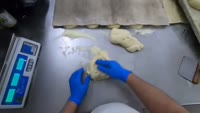} \includegraphics[width=0.20\linewidth]{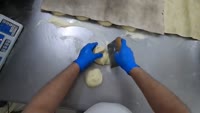}} \\
\hline
% --- Task (full-width) ---
\multicolumn{5}{|>{\columncolor{colTask}}p{\dimexpr \linewidth+6\tabcolsep+4\arrayrulewidth\relax}|}{\textbf{Task}: Prepare dough for baking by transforming it into uniform, shaped pieces, and positioning them for further processing or baking. \textit{[Source: \textbf{Ego4D}]}} \\
\hline
% --- Dialog history (full-width) ---
\multicolumn{5}{|>{\columncolor{colDialog}}p{\dimexpr \linewidth+6\tabcolsep+4\arrayrulewidth\relax}|}{\textbf{Dialog history}: \textbf{1.}\;Sure! Let's get started. First, Retrieve a large wooden rolling pin from a bread baking rack and position it on a metal table.  \textbf{2.}\;Move a large bread board from the rack to the metal table and rotate it to a suitable position.  \textbf{3.}\;Cut a large lump of dough into smaller pieces using a bench scraper.  \textbf{4.}\;Knead and shape the dough pieces into cylindrical forms and place them on a long wooden paddle.} \\
\hline
% --- Header row ---
\textbf{Plan (prev $\rightarrow$ curr)} & \textbf{Cue (\$ interrupt\$)} & \textbf{Golden} & \textbf{Ours}\,{\textcolor{darkgreen}{\cmark}} & \textbf{\texttt{qwen}}\,{\textcolor{red}{\xmark}} \\
\hline
% --- Data row ---
\textit{Transfer the shaped dough pieces from the wooden paddle to a rack for further preparation or proofing.} $\rightarrow$ \textbf{Mix additional dough with flour and water in a mixer.} & % Mixer bowl contains uniformly combined dough with no visible dry flour or liquid water \newline Mixing tool is stationary with no motion blur \newline Mixer power indicator is off or safety guard is closed \newline Hand is not interacting with mixer or ingredients \newline Ingredients containers are empty or set aside
Dough uniformly combined in bowl, mixer stationary and off, hands withdrawn, ingredient containers empty or set aside. & Mix additional dough with flour and water in a mixer. & 
Add more dough to the flour mixture in the mixer. & \textcolor{gray}{\textbf{[silent]}} \\
\hline
\end{tabular}%
}
\centering
\setlength{\tabcolsep}{2pt}
\renewcommand{\arraystretch}{1.0}
\scriptsize
\catstripe{colCat2}{%
\begin{tabular}{@{}|>{\columncolor{colPlan}}p{0.28\linewidth}|>{\columncolor{colCue}}p{0.15\linewidth}|>{\columncolor{colGolden}}p{0.14\linewidth}|>{\columncolor{colOurs}}p{0.14\linewidth}|>{\columncolor{colThreeP}}p{0.29\linewidth}|@{}}
% --- Frames row (spans all 5 columns; series label sticks out to the left) ---
\multicolumn{5}{@{}c@{}}{
%\makebox[0pt][r]{\large\bfseries 3d\hspace{0.4cm}}
\includegraphics[width=0.20\linewidth]{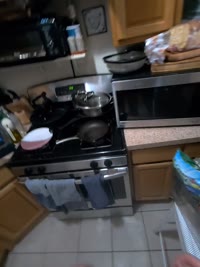} \includegraphics[width=0.20\linewidth]{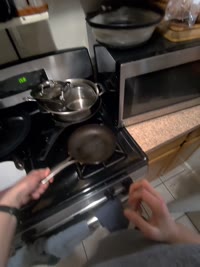} \includegraphics[width=0.20\linewidth]{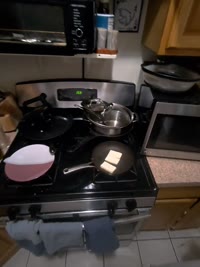} \includegraphics[width=0.20\linewidth]{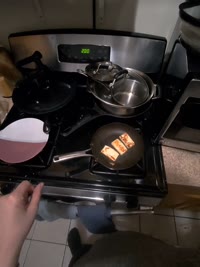} \includegraphics[width=0.20\linewidth]{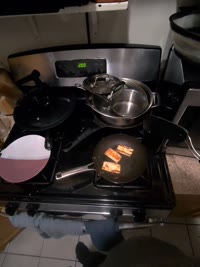}} \\
\hline
% --- Task (full-width) ---
\multicolumn{5}{|>{\columncolor{colTask}}p{\dimexpr \linewidth+6\tabcolsep+4\arrayrulewidth\relax}|}{\textbf{Task}: Making pan-seared halloumi with honey drizzle. \textit{[Source: \textbf{EgoProactive}]}} \\
\hline
% --- Dialog history (full-width) ---
\multicolumn{5}{|>{\columncolor{colDialog}}p{\dimexpr \linewidth+6\tabcolsep+4\arrayrulewidth\relax}|}{\textbf{Dialog history}: \textbf{1.}\;Sure! Let's get started. First, Prepare the halloumi: remove it from its packaging and slice the block into 1/2‐inch thick pieces.  \textbf{2.}\;Set up the cooking equipment: place a non‐stick skillet on the stove, turn the burner to medium heat, and let the pan warm up.  \textbf{3.}\;Eat a small snack (crackers) while the pan heats. (interruption, Body \& Health)  \textbf{4.}\;Add butter to the hot skillet, then place the halloumi slices in the pan without overlapping.} \\
\hline
% --- Header row ---
\textbf{Plan (prev $\rightarrow$ curr)} & \textbf{Cue (\$ interrupt\$)} & \textbf{Golden} & \textbf{Ours}\,{\textcolor{darkgreen}{\cmark}} & \textbf{\texttt{qwen3 235B}} \\
\hline
% --- Data row ---
\textit{Cook the halloumi: let it sizzle, flip the slices with a spatula, and continue cooking each side for 2‐3 minutes until golden brown.} $\rightarrow$ \textbf{Adjust the stove knob and pan position as needed to finish cooking the halloumi. \textbf{(fix)}} & 
Pan resting motionless on burner with knob off and utensil set aside, hands withdrawn as food browns evenly
 & Adjust the stove knob and pan position as needed to finish cooking the halloumi. & Adjust the stove knob and pan position as needed to finish cooking the halloumi. & The halloumi is now golden brown and fully cooked. Turn off the stove, then transfer the slices to a serving
  plate and arrange them neatly. Next, drizzle honey generously over the warm halloumi. \\
\hline
\end{tabular}%
}
\captionsetup{skip=2pt,font=footnotesize,labelsep=period,labelfont=bf}\caption{Step-completion interrupts (continued).}
\end{figure*}

% ------------------------------------------------------------------
% (C) OOP EXTERNAL INTERRUPTIONS
% ------------------------------------------------------------------
\begin{figure*}[t]
\centering
% ------------------------------------------------------------------
% (C) OOP EXTERNAL INTERRUPTIONS — user is distracted off-task; model fires and guides back
% ------------------------------------------------------------------
\catbanner{colCat3}{(C)}{{\bf OOP Distractions}: Detection and Re-engagement}
\setlength{\tabcolsep}{2pt}
\renewcommand{\arraystretch}{1.0}
\scriptsize
\catstripe{colCat3}{%
\begin{tabular}{@{}|>{\columncolor{colPlan}}p{0.28\linewidth}|>{\columncolor{colCue}}p{0.15\linewidth}|>{\columncolor{colGolden}}p{0.10\linewidth}|>{\columncolor{colOurs}}p{0.12\linewidth}|>{\columncolor{colThreeP}}p{0.35\linewidth}|@{}}
% --- Frames row (spans all 5 columns; series label sticks out to the left) ---
\multicolumn{5}{@{}c@{}}{
% \makebox[0pt][r]{\large\bfseries 4a\hspace{0.4cm}}
\includegraphics[width=0.20\linewidth]{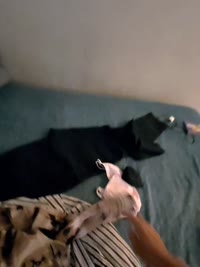} \includegraphics[width=0.20\linewidth]{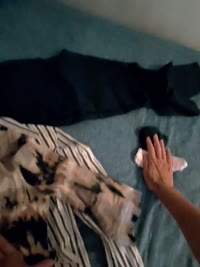} \includegraphics[width=0.20\linewidth]{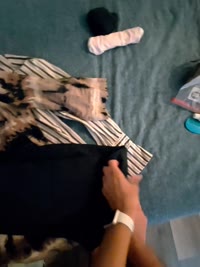} \includegraphics[width=0.20\linewidth]{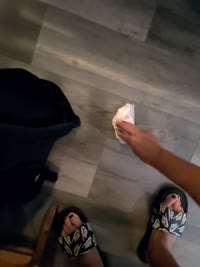} \includegraphics[width=0.20\linewidth]{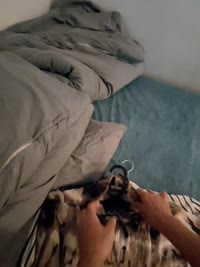}} \\
\hline
% --- Task (full-width) ---
\multicolumn{5}{|>{\columncolor{colTask}}p{\dimexpr 1.0000\linewidth+6\tabcolsep+4\arrayrulewidth\relax}|}{\textbf{Task}: How to pack a carry-on bag efficiently. \textit{[Source: \textbf{EgoProactive}]}} \\
\hline
% --- Dialog history (full-width) ---
\multicolumn{5}{|>{\columncolor{colDialog}}p{\dimexpr 1.0000\linewidth+6\tabcolsep+4\arrayrulewidth\relax}|}{\textbf{Dialog history}: \textbf{1.}\;Sure! Let's get started. First, Lay out and organize clothing items on the bed, including shirts, pants, socks, and undergarments.  \textbf{2.}\;Roll each clothing item tightly (shirts, pants, socks) to save space.} \\
\hline
% --- Header row ---
\textbf{Plan (prev $\rightarrow$ curr)} & \textbf{Cue (\$ interrupt\$)} & \textbf{Golden} & \textbf{Ours}\,{\textcolor{darkgreen}{\cmark}} & \textbf{\texttt{gpt}}\,{\textcolor{darkgreen}{\cmark}} \\
\hline
% --- Data row ---
\textit{(interruption, Object \& Task-Management Interruptions) Pause to wipe the floor after noticing a spill, using a wipe from the package.} $\rightarrow$ \textbf{ Resume rolling the remaining clothing items and place them aside. \textbf{(fix)} } & %All remaining garments are in compact rolled form and positioned in a separate pile away from the active workspace \newline Hands are empty and withdrawn from the clothing items, indicating no further manipulation \newline The work surface shows cleared space where rolling was performed, with no flat garments remaining \newline Rolled items maintain stable cylindrical shape without being held or adjusted
All garments rolled and piled away, hands withdrawn, workspace cleared, rolls stable without adjustment. & \textbf{(fix)} Resume rolling the remaining clothing items and place them aside. & 
Pick back up rolling the remaining clothing items and set them aside. & You've paused rolling—resume by rolling the remaining flat garments (the patterned/striped shirt) into tight cylinders and move each finished roll into a separate 'done' pile off to one side. Once everything is rolled, tell me and we'll move to packing toiletries in a ziplock. \\
\hline
\end{tabular}%
}
% \end{figure*}

% \begin{figure*}[t]
\centering
\setlength{\tabcolsep}{2pt}
\renewcommand{\arraystretch}{1.0}
\scriptsize
\catstripe{colCat3}{%
\begin{tabular}{@{}|>{\columncolor{colPlan}}p{0.36\linewidth}|>{\columncolor{colCue}}p{0.20\linewidth}|>{\columncolor{colGolden}}p{0.18\linewidth}|>{\columncolor{colOurs}}p{0.20\linewidth}|>{\columncolor{colThreeP}}p{0.060\linewidth}|@{}}
% --- Frames row (spans all 5 columns; series label sticks out to the left) ---
\multicolumn{5}{@{}c@{}}{
%\makebox[0pt][r]{\large\bfseries 4b\hspace{0.4cm}}
\includegraphics[width=0.20\linewidth]{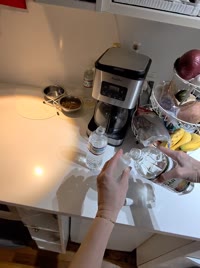} \includegraphics[width=0.20\linewidth]{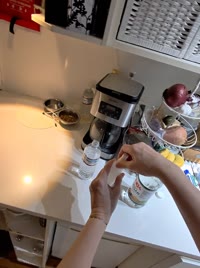} \includegraphics[width=0.20\linewidth]{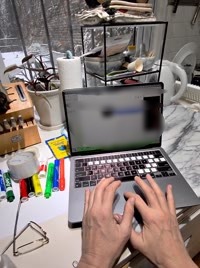} \includegraphics[width=0.20\linewidth]{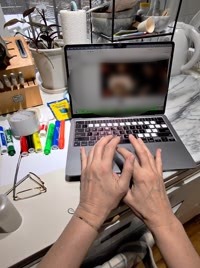} \includegraphics[width=0.20\linewidth]{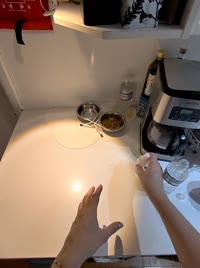}} \\
\hline
% --- Task (full-width) ---
\multicolumn{5}{|>{\columncolor{colTask}}p{\dimexpr \linewidth+6\tabcolsep+4\arrayrulewidth\relax}|}{\textbf{Task}: How to remove wrinkles with a spray bottle. \textit{[Source: \textbf{EgoProactive}]}} \\
\hline
% --- Dialog history (full-width) ---
\multicolumn{5}{|>{\columncolor{colDialog}}p{\dimexpr \linewidth+6\tabcolsep+4\arrayrulewidth\relax}|}{\textbf{Dialog history}: \textbf{1.}\;Sure! Let's get started. First, Fill a clean spray bottle with distilled water and add a few drops of white vinegar.  \textbf{2.}\;Shake the spray bottle to mix the solution.} \\
\hline
% --- Header row ---
\textbf{Plan (prev $\rightarrow$ curr)} & \textbf{Cue (\$ interrupt\$)} & \textbf{Golden} & \textbf{Ours}\,{\textcolor{darkgreen}{\cmark}} & \textbf{\texttt{qwen}}\,{\textcolor{red}{\xmark}} \\
\hline
% --- Data row ---
\textit{(interruption, Media \& Entertainment Interruptions) Switch from music to a podcast (or vice versa) while shaking the bottle.} $\rightarrow$ \textbf{Resume shaking the spray bottle to ensure the solution is well mixed. \textbf{(fix)} } & %spray bottle held completely still with no motion blur visible \newline hand grip relaxes from shaking position to a neutral hold or releases the bottle \newline bottle returned to upright orientation and remains motionless for several beats \newline user's attention shifts to next action such as aiming the nozzle \newline any visible liquid inside appears uniform and settled without agitation bubbles 
Spray bottle still, upright and stable, grip relaxed, liquid settled, attention shifting to next action. & \textbf{(fix)} 
Pick back up shaking the spray bottle to make sure the solution is well mixed. & Resume shaking the spray bottle to ensure the solution is well mixed. & \textcolor{gray}{\textbf{[silent]}} \\
\hline
\end{tabular}%
}
% \end{figure*}

% \begin{figure*}[t]
\centering
\setlength{\tabcolsep}{2pt}
\renewcommand{\arraystretch}{1.0}
\scriptsize
\catstripe{colCat3}{%
\begin{tabular}{@{}|>{\columncolor{colPlan}}p{0.34\linewidth}|>{\columncolor{colCue}}p{0.195\linewidth}|>{\columncolor{colGolden}}p{0.115\linewidth}|>{\columncolor{colOurs}}p{0.11\linewidth}|>{\columncolor{colThreeP}}p{0.24\linewidth}|@{}}
% --- Frames row (spans all 5 columns; series label sticks out to the left) ---
\multicolumn{5}{@{}c@{}}{
%\makebox[0pt][r]{\large\bfseries 4c\hspace{0.4cm}}
\includegraphics[width=0.20\linewidth]{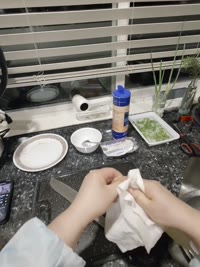} \includegraphics[width=0.20\linewidth]{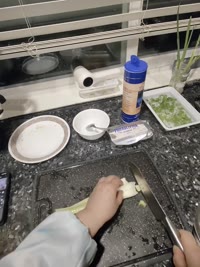} \includegraphics[width=0.20\linewidth]{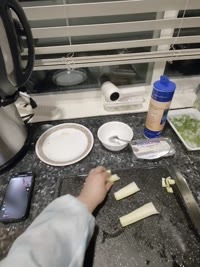} \includegraphics[width=0.20\linewidth]{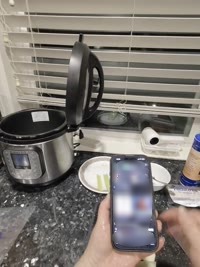} \includegraphics[width=0.20\linewidth]{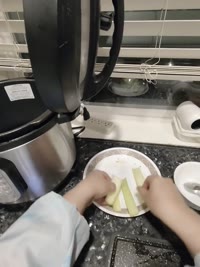}} \\
\hline
% --- Task (full-width) ---
\multicolumn{5}{|>{\columncolor{colTask}}p{\dimexpr \linewidth+6\tabcolsep+4\arrayrulewidth\relax}|}{\textbf{Task}: Making herbed cream cheese stuffed celery sticks. \textit{[Source: \textbf{EgoProactive}]}} \\
\hline
% --- Dialog history (full-width) ---
\multicolumn{5}{|>{\columncolor{colDialog}}p{\dimexpr \linewidth+6\tabcolsep+4\arrayrulewidth\relax}|}{\textbf{Dialog history}: \textbf{1.}\;Sure! Let's get started. First, Wash the celery stalks under running water and pat them dry with a paper towel.  \textbf{2.}\;Trim the ends of the celery and cut the stalks into 4‐inch sticks, then arrange the sticks on a plate.} \\
\hline
% --- Header row ---
\textbf{Plan (prev $\rightarrow$ curr)} & \textbf{Cue (\$ interrupt\$)} & \textbf{Golden} & \textbf{Ours}\,{\textcolor{darkgreen}{\cmark}} & \textbf{\texttt{qwen3 235B}} \\
\hline
% --- Data row ---
\textit{Switch the music or podcast playback on the smartphone. (interruption, Media \& Entertainment Interruptions)} $\rightarrow$ \textbf{Keep arranging the celery sticks on the plate until neatly set. \textbf{(fix)}} & % All intended pieces are resting on the serving plate with none left on the preparation surface \newline hands are withdrawn from the arrangement area and positioned away from the items \newline pieces are aligned in a uniform orientation with consistent spacing \newline no motion blur or active manipulation is visible around the arranged items \newline the layout remains stable and unchanged for a moment indicating finalization
All pieces arranged on serving plate, evenly spaced and aligned, hands withdrawn, layout settled and stable. & Keep arranging the celery sticks on the plate until neatly set. & Continue arranging the celery sticks neatly on the plate. & Open the cream cheese package, scoop the softened cream cheese into a bowl, chop fresh herbs, and add the herbs plus a pinch of salt to the bowl. \\
\hline
\end{tabular}%
}
\captionsetup{skip=2pt,font=footnotesize,labelsep=period,labelfont=bf}\caption{Qualitative examples: out-of-procedure distractions.}
\label{fig:qual_oop_distraction}
\end{figure*}

\begin{figure*}[t]
\ContinuedFloat
\centering
\setlength{\tabcolsep}{2pt}
\renewcommand{\arraystretch}{1.0}
\scriptsize
\catstripe{colCat3}{%
\begin{tabular}{@{}|>{\columncolor{colPlan}}p{0.34\linewidth}|>{\columncolor{colCue}}p{0.19\linewidth}|>{\columncolor{colGolden}}p{0.19\linewidth}|>{\columncolor{colOurs}}p{0.22\linewidth}|>{\columncolor{colThreeP}}p{0.060\linewidth}|@{}}
% --- Frames row (spans all 5 columns; series label sticks out to the left) ---
\multicolumn{5}{@{}c@{}}{
% \makebox[0pt][r]{\large\bfseries 4d\hspace{0.4cm}}
\includegraphics[width=0.20\linewidth]{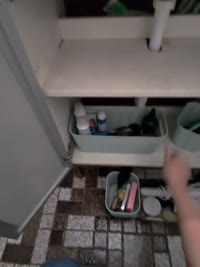} \includegraphics[width=0.20\linewidth]{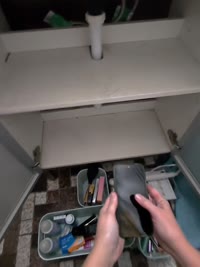} \includegraphics[width=0.20\linewidth]{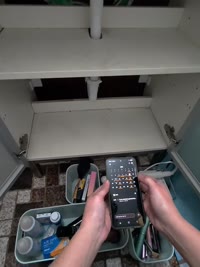} \includegraphics[width=0.20\linewidth]{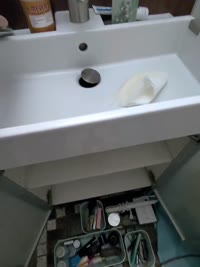} \includegraphics[width=0.20\linewidth]{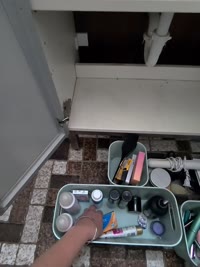}} \\
\hline
% --- Task (full-width) ---
\multicolumn{5}{|>{\columncolor{colTask}}p{\dimexpr 1.0000\linewidth+6\tabcolsep+4\arrayrulewidth\relax}|}{\textbf{Task}: How to organize under the bathroom sink. \textit{[Source: \textbf{EgoProactive}]}} \\
\hline
% --- Dialog history (full-width) ---
\multicolumn{5}{|>{\columncolor{colDialog}}p{\dimexpr 1.0000\linewidth+6\tabcolsep+4\arrayrulewidth\relax}|}{\textbf{Dialog history}: \textbf{1.}\;Sure! Let's get started. First, Open the bathroom sink cabinet doors and remove all items, placing them on the floor or countertop.} \\
\hline
% --- Header row ---
\textbf{Plan (prev $\rightarrow$ curr)} & \textbf{Cue (\$ interrupt\$)} & \textbf{Golden} & \textbf{Ours}\,{\textcolor{darkgreen}{\cmark}} & \textbf{\texttt{claude}}\,{\textcolor{red}{\xmark}} \\
\hline
% --- Data row ---
\textit{(interruption, Information \& Planning) Check the to‐do list on the smartphone.} $\rightarrow$ \textbf{Continue sorting the items under the sink, grouping similar items together (e.g., toiletries, cleaning supplies, extra toilet paper). \textbf{(fix)} } & % hands withdrawn from the sorting area and resting away, indicating no active manipulation \newline items visibly grouped by category within separate containers with no cross-mixing \newline cabinet floor cleared of loose items with everything placed inside containers \newline consistent alignment and orientation of similar items within each group \newline containers arranged neatly in rows or zones within the cabinet space 
Items sorted and grouped in separate containers, cabinet cleared, containers neatly arranged, hands withdrawn. & Continue sorting the items under the sink, grouping similar items together (e.g., toiletries, cleaning supplies, extra toilet paper). & Pick back up sorting the items under the sink, grouping similar ones together such as toiletries, cleaning supplies, and extra toilet paper. & \textcolor{gray}{\textbf{[silent]}} \\
\hline
\end{tabular}%
}
% \end{figure*}

% \begin{figure*}[t]
\centering
\setlength{\tabcolsep}{2pt}
\renewcommand{\arraystretch}{1.0}
\scriptsize
\catstripe{colCat3}{%
\begin{tabular}{@{}|>{\columncolor{colPlan}}p{0.38\linewidth}|>{\columncolor{colCue}}p{0.25\linewidth}|>{\columncolor{colGolden}}p{0.15\linewidth}|>{\columncolor{colOurs}}p{0.15\linewidth}|>{\columncolor{colThreeP}}p{0.070\linewidth}|@{}}
% --- Frames row (spans all 5 columns; series label sticks out to the left) ---
\multicolumn{5}{@{}c@{}}{
%\makebox[0pt][r]{\large\bfseries 4e\hspace{0.4cm}}
\includegraphics[width=0.20\linewidth]{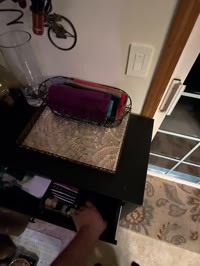} \includegraphics[width=0.20\linewidth]{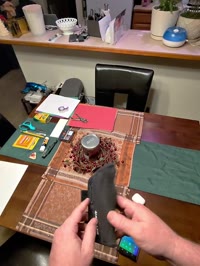} \includegraphics[width=0.20\linewidth]{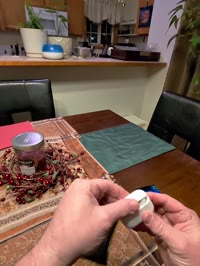} \includegraphics[width=0.20\linewidth]{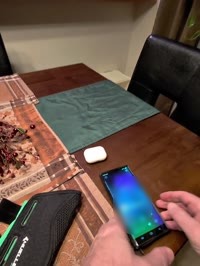} \includegraphics[width=0.20\linewidth]{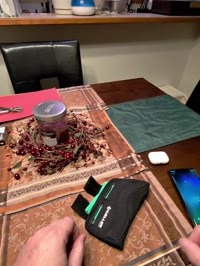}} \\
\hline
% --- Task (full-width) ---
\multicolumn{5}{|>{\columncolor{colTask}}p{\dimexpr 1.0000\linewidth+6\tabcolsep+4\arrayrulewidth\relax}|}{\textbf{Task}: How to apply a wrist support brace. \textit{[Source: \textbf{EgoProactive}]}} \\
\hline
% --- Dialog history (full-width) ---
\multicolumn{5}{|>{\columncolor{colDialog}}p{\dimexpr 1.0000\linewidth+6\tabcolsep+4\arrayrulewidth\relax}|}{\textbf{Dialog history}: \textbf{1.}\;Sure! Let's get started. First, Open the drawer and retrieve the wrist support brace.  \textbf{2.}\;Sit at a stable surface such as a table or chair, place the brace on the table, and handle the packaging while preparing.} \\
\hline
% --- Header row ---
\textbf{Plan (prev $\rightarrow$ curr)} & \textbf{Cue (\$ interrupt\$)} & \textbf{Golden} & \textbf{Ours}\,{\textcolor{darkgreen}{\cmark}} & \textbf{\texttt{claude}}\,{\textcolor{red}{\xmark}} \\
\hline
% --- Data row ---
\textit{(interruption, Media \& Entertainment Interruptions) Notice the audio playing from the phone speaker and switch the output to headphones.} $\rightarrow$ \textbf{Return your attention to the task and position the wrist support brace directly in front of you on the table. \textbf{(fix)}  } & % wrist support rests motionless on the flat surface with its interior opening facing upward and accessible \newline both hands are released and positioned beside or hovering near the support without contact \newline the support is centered in the user's immediate workspace directly in front of their torso \newline the support appears fully unfolded and ready for the next action \newline a brief pause occurs with no manipulation indicating placement is finished 
Wrist support placed flat with opening facing up, centered in workspace, hands released and hovering nearby, brief pause indicating placement complete. & Return your attention to the task and position the wrist support brace directly in front of you on the table. & Return your attention to the task and position the wrist support brace directly in front of you on the table. & \textcolor{gray}{\textbf{[silent]}} \\
\hline
\end{tabular}%
}
% \end{figure*}

% \begin{figure*}[t]
\centering
\setlength{\tabcolsep}{2pt}
\renewcommand{\arraystretch}{1.0}
\scriptsize
\catstripe{colCat3}{%
\begin{tabular}{@{}|>{\columncolor{colPlan}}p{0.25\linewidth}|>{\columncolor{colCue}}p{0.16\linewidth}|>{\columncolor{colGolden}}p{0.12\linewidth}|>{\columncolor{colOurs}}p{0.12\linewidth}|>{\columncolor{colThreeP}}p{0.35\linewidth}|@{}}
% --- Frames row (spans all 5 columns; series label sticks out to the left) ---
\multicolumn{5}{@{}c@{}}{
%\makebox[0pt][r]{\large\bfseries 4f\hspace{0.4cm}}
\includegraphics[width=0.20\linewidth]{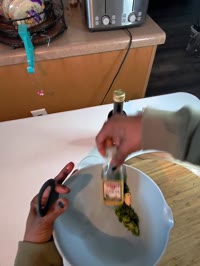} \includegraphics[width=0.20\linewidth]{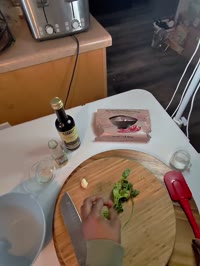} \includegraphics[width=0.20\linewidth]{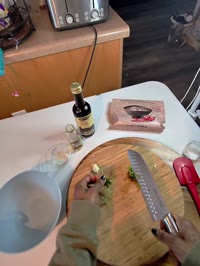} \includegraphics[width=0.20\linewidth]{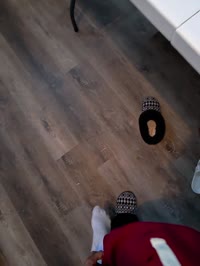} \includegraphics[width=0.20\linewidth]{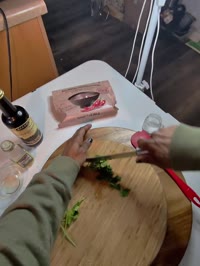}} \\
\hline
% --- Task (full-width) ---
\multicolumn{5}{|>{\columncolor{colTask}}p{\dimexpr 1.0000\linewidth+6\tabcolsep+4\arrayrulewidth\relax}|}{\textbf{Task}: Making fresh chimichurri sauce. \textit{[Source: \textbf{EgoProactive}]}} \\
\hline
% --- Dialog history (full-width) ---
\multicolumn{5}{|>{\columncolor{colDialog}}p{\dimexpr 1.0000\linewidth+6\tabcolsep+4\arrayrulewidth\relax}|}{\textbf{Dialog history}: \textbf{1.}\;Sure! Let's get started. First, Gather and prepare ingredients: retrieve fresh parsley, cilantro, garlic, red wine vinegar, olive oil, and red pepper flakes from the pantry/fridge and place them on the cutting board.  \textbf{2.}\;Transfer the parsley and cilantro to the cutting board, separate leaves from stems, and roughly chop the herbs.} \\
\hline
% --- Header row ---
\textbf{Plan (prev $\rightarrow$ curr)} & \textbf{Cue (\$ interrupt\$)} & \textbf{Golden} & \textbf{Ours}\,{\textcolor{darkgreen}{\cmark}} & \textbf{\texttt{gpt}}\,{\textcolor{darkgreen}{\cmark}} \\
\hline
% --- Data row ---
\textit{Put on slippers to warm feet before continuing. (interruption, Sensory \& Comfort Adjustments)} $\rightarrow$ \textbf{Resume chopping the herbs until they are finely minced, then transfer the chopped herbs to the mixing bowl. \textbf{(fix)}} & % herbs are reduced to uniformly small, consistent pieces with no large leaf fragments visible \newline the knife is set down or resting stationary away from the herb pile \newline hands are positioned to gather or sweep the herbs rather than in chopping grip \newline the mixing bowl contains the freshly chopped herbs \newline the cutting board is largely cleared of herbs after transfer 
Herbs finely and evenly chopped, knife set aside, hands shifting to gather or sweep herbs into mixing bowl, cutting board cleared. & Resume chopping the herbs until they are finely minced, then transfer the chopped herbs to the mixing bowl. & Pick back up chopping the herbs until finely minced, then sweep them into the mixing bowl. & Skip the slippers for now—focus on the sauce.
Next step: strip the parsley/cilantro leaves from the thicker stems and pile the leaves on the cutting board. Then mince the garlic clove(s). After that, finely chop the herbs together (keep your fingertips tucked and use a rocking motion). \\
\hline
\end{tabular}%
}
\captionsetup{skip=2pt,font=footnotesize,labelsep=period,labelfont=bf}\caption{Out-of-procedure distractions (continued).}
\end{figure*}

\begin{figure*}[t]
\centering
\setlength{\tabcolsep}{2pt}
\renewcommand{\arraystretch}{1.0}
\scriptsize
\catstripe{colCat3}{%
\begin{tabular}{@{}|>{\columncolor{colPlan}}p{0.35\linewidth}|>{\columncolor{colCue}}p{0.25\linewidth}|>{\columncolor{colGolden}}p{0.17\linewidth}|>{\columncolor{colOurs}}p{0.18\linewidth}|>{\columncolor{colThreeP}}p{0.050\linewidth}|@{}}
% --- Frames row (spans all 5 columns; series label sticks out to the left) ---
\multicolumn{5}{@{}c@{}}{
%\makebox[0pt][r]{\large\bfseries 4g\hspace{0.4cm}}
\includegraphics[width=0.20\linewidth]{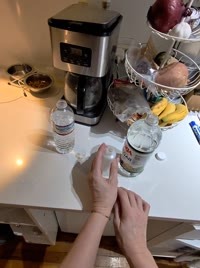} \includegraphics[width=0.20\linewidth]{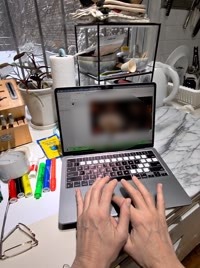} \includegraphics[width=0.20\linewidth]{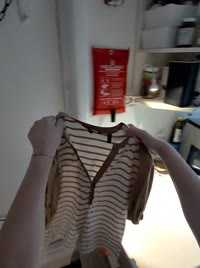} \includegraphics[width=0.20\linewidth]{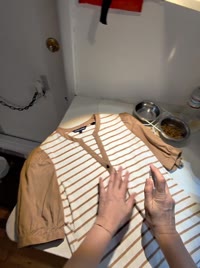} \includegraphics[width=0.20\linewidth]{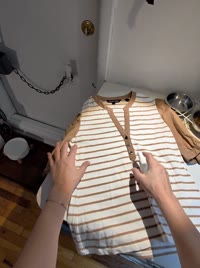}} \\
\hline
% --- Task (full-width) ---
\multicolumn{5}{|>{\columncolor{colTask}}p{\dimexpr \linewidth+6\tabcolsep+4\arrayrulewidth\relax}|}{\textbf{Task}: How to remove wrinkles with a spray bottle. \textit{[Source: \textbf{EgoProactive}]}} \\
\hline
% --- Dialog history (full-width) ---
\multicolumn{5}{|>{\columncolor{colDialog}}p{\dimexpr \linewidth+6\tabcolsep+4\arrayrulewidth\relax}|}{\textbf{Dialog history}: \textbf{1.}\;Sure! Let's get started. First, Fill a clean spray bottle with distilled water and add a few drops of white vinegar.  \textbf{2.}\;Shake the spray bottle to mix the solution.  \textbf{3.}\;(interruption, Media \& Entertainment Interruptions) Switch from music to a podcast (or vice versa) while shaking the bottle.  \textbf{4.}\;\textbf{(fix)} Resume shaking the spray bottle to ensure the solution is well mixed.  \textbf{5.}\;Lay the wrinkled shirt flat on a clean, dry surface such as a table or ironing board.  \textbf{6.}\;Lightly mist the wrinkled areas of the shirt with the spray, keeping the bottle about 6 inches away.} \\
\hline
% --- Header row ---
\textbf{Plan (prev $\rightarrow$ curr)} & \textbf{Cue (\$ interrupt\$)} & \textbf{Golden} & \textbf{Ours}\,{\textcolor{darkgreen}{\cmark}} & \begin{adjustbox}{max width=\linewidth}\textbf{\texttt{gemini}}\,{\textcolor{red}{\xmark}}\end{adjustbox} \\
\hline
% --- Data row ---
\textit{(interruption, Body \& Health) Pause to adjust a bracelet (or watch/ring) that feels uncomfortable.} $\rightarrow$ \textbf{Return to the shirt and continue misting any remaining wrinkled sections \textbf{(fix)}.} & % spray bottle placed aside or held idle with no mist being produced \newline hands perform final broad smoothing passes across the entire garment surface \newline the fabric appears evenly damp with no isolated dry wrinkled areas being targeted \newline hands rest in a relaxed position near the garment edges indicating work pause \newline visual attention shifts from specific wrinkles to overall garment inspection 
Spray bottle idle, no mist, hands making final broad smoothing passes across evenly damp fabric, hands relaxing near garment edges. & Return to the shirt and continue misting any remaining wrinkled sections. & Go back to the shirt and keep misting any sections still wrinkled. & \textcolor{gray}{\textbf{[silent]}} \\
\hline
\end{tabular}%
}
% ------------------------------------------------------------------
% (D) OOP USER MISTAKES — user makes procedural error; model detects and corrects
% ------------------------------------------------------------------
\catbanner{colCat4}{(D)}{{\bf OOP USER MISTAKES}: Error detection and Correction}
\setlength{\tabcolsep}{2pt}
\renewcommand{\arraystretch}{1.0}
\scriptsize
\catstripe{colCat4}{%
\begin{tabular}{@{}|>{\columncolor{colPlan}}p{0.25\linewidth}|>{\columncolor{colCue}}p{0.13\linewidth}|>{\columncolor{colGolden}}p{0.12\linewidth}|>{\columncolor{colOurs}}p{0.12\linewidth}|>{\columncolor{colThreeP}}p{0.38\linewidth}|@{}}
% --- Frames row (spans all 5 columns; series label sticks out to the left) ---
\multicolumn{5}{@{}c@{}}{
% \makebox[0pt][r]{\large\bfseries 3e\hspace{0.4cm}}
\includegraphics[width=0.20\linewidth]{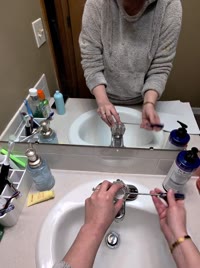} \includegraphics[width=0.20\linewidth]{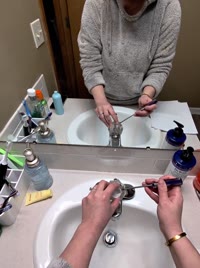} \includegraphics[width=0.20\linewidth]{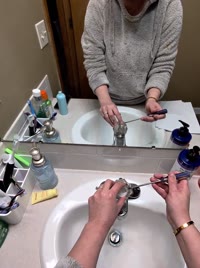} \includegraphics[width=0.20\linewidth]{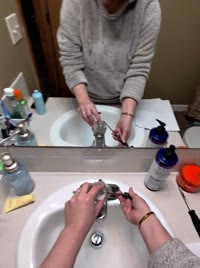} \includegraphics[width=0.20\linewidth]{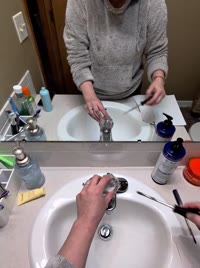}} \\
\hline
% --- Task (full-width) ---
\multicolumn{5}{|>{\columncolor{colTask}}p{\dimexpr 1.0000\linewidth+6\tabcolsep+4\arrayrulewidth\relax}|}{\textbf{Task}: Tightening loose faucet handle. \textit{[Source: \textbf{EgoProactive}]}} \\
\hline
% --- Dialog history (full-width) ---
\multicolumn{5}{|>{\columncolor{colDialog}}p{\dimexpr 1.0000\linewidth+6\tabcolsep+4\arrayrulewidth\relax}|}{\textbf{Dialog history}: ---} \\
\hline
% --- Header row ---
\textbf{Plan (prev $\rightarrow$ curr)} & \textbf{Cue (\$ interrupt\$)} & \textbf{Golden} & \textbf{Ours}\,{\textcolor{darkgreen}{\cmark}} & \textbf{\texttt{claude}}\,{\textcolor{darkgreen}{\cmark}} \\
\hline
% --- Data row ---
\textit{Grab a screwdriver without checking if it matches the faucet handle screw head and begin adjusting the knob. (mistake, Tool Selection Error)} $\rightarrow$ \textbf{Notice the screwdriver slipping, place it down, and switch to a properly fitting screwdriver before continuing. \textbf{(fix)}} & % the ill-fitting tool is resting on a stable surface away from the fastener \newline the user's hand is now gripping a different tool that appears properly sized for the fastener \newline the fastener is clear of any tool engagement, ready for the new tool to be applied \newline body posture shifts from reactive withdrawal to preparatory alignment with the new tool \newline visual focus moves to positioning the replacement tool rather than the previous slipping one 
Ill-fitting tool set aside, hand now gripping a properly sized replacement, posture shifting to align new tool with fastener. & Notice the screwdriver slipping, place it down, and switch to a properly fitting screwdriver before continuing. & Set down the slipping screwdriver and swap to one that fits properly before continuing. & Good — I can see you've set the screwdriver down on the counter and are now working the faucet handle by hand. That's fine for loosening the handle knob. Now, Step 2: Remove the faucet handle — carefully pull or lift the handle off to expose the internal stem and packing nut underneath. Set the handle aside on the counter. Once exposed, we'll tighten the packing nut or stem to fix the looseness. \\
\hline
\end{tabular}%
}
% \end{figure*}

% \begin{figure*}[t]
\centering
\setlength{\tabcolsep}{2pt}
\renewcommand{\arraystretch}{1.0}
\scriptsize
\catstripe{colCat4}{%
\begin{tabular}{@{}|>{\columncolor{colPlan}}p{0.40\linewidth}|>{\columncolor{colCue}}p{0.18\linewidth}|>{\columncolor{colGolden}}p{0.18\linewidth}|>{\columncolor{colOurs}}p{0.14\linewidth}|>{\columncolor{colThreeP}}p{0.10\linewidth}|@{}}
% --- Frames row (spans all 5 columns; series label sticks out to the left) ---
\multicolumn{5}{@{}c@{}}{
%\makebox[0pt][r]{\large\bfseries 4h\hspace{0.4cm}}
\includegraphics[width=0.20\linewidth]{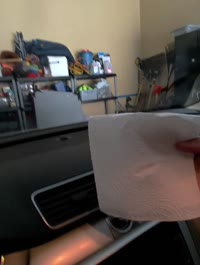} \includegraphics[width=0.20\linewidth]{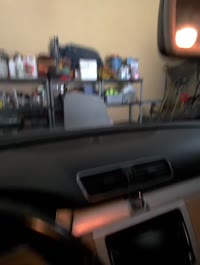} \includegraphics[width=0.20\linewidth]{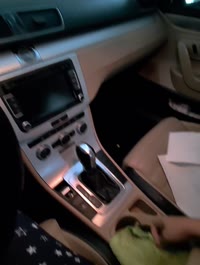} \includegraphics[width=0.20\linewidth]{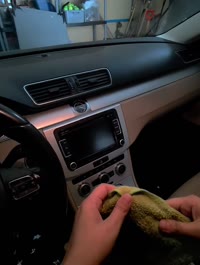} \includegraphics[width=0.20\linewidth]{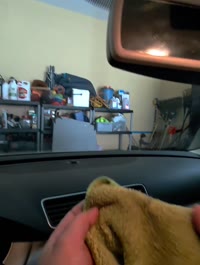}} \\
\hline
% --- Task (full-width) ---
\multicolumn{5}{|>{\columncolor{colTask}}p{\dimexpr 1.0000\linewidth+6\tabcolsep+4\arrayrulewidth\relax}|}{\textbf{Task}: How to install a phone mount in your car. \textit{[Source: \textbf{EgoProactive}]}} \\
\hline
% --- Dialog history (full-width) ---
\multicolumn{5}{|>{\columncolor{colDialog}}p{\dimexpr 1.0000\linewidth+6\tabcolsep+4\arrayrulewidth\relax}|}{\textbf{Dialog history}: ---} \\
\hline
% --- Header row ---
\textbf{Plan (prev $\rightarrow$ curr)} & \textbf{Cue (\$ interrupt\$)} & \textbf{Golden} & \textbf{Ours}\,{\textcolor{darkgreen}{\cmark}} & \begin{adjustbox}{max width=\linewidth}\textbf{\texttt{gemini}}\,{\textcolor{darkgreen}{\cmark}}\end{adjustbox} \\
\hline
% --- Data row ---
\textit{(mistake, Preparation/Setup Error) Wipe the chosen surface with a dry paper towel only, leaving oily residue that reduces adhesive grip.} $\rightarrow$ \textbf{Use a slightly damp cloth to wipe the surface again, then dry it thoroughly with a clean paper towel before proceeding. \textbf{(fix)}} & % cloth is lifted away and set aside with no further contact to the target surface \newline hands are free or holding a different tool, signaling the cleaning phase has ended \newline the surface appears uniformly dry and free of visible streaks or moisture \newline a separate drying material (e.g., paper towel) has been applied and removed \newline the person pauses and shifts attention toward the next installation component 
Cloth set aside, surface dry and streak-free, hands free or holding next tool, attention shifting to next component. & Use a slightly damp cloth to wipe the surface again, then dry it thoroughly with a clean paper towel before proceeding. & Wipe the surface again with a slightly damp cloth, then dry it  with a clean paper towel. & Next, peel the backing off the phone mount. \\
\hline
\end{tabular}%
}
\captionsetup{skip=2pt,font=footnotesize,labelsep=period,labelfont=bf}\caption{Qualitative examples: out-of-procedure user mistakes.}
\label{fig:qual_oop_mistake}
\end{figure*}

\FloatBarrier        % flush the (floating) per-category galleries within this section

\clearpage\begingroup\centering\setlength{\parskip}{0pt}\setlength{\topsep}{0pt}\vspace*{-1cm}
% \vspace{-1cm}
\centering
% \caption*{\textbf{Ours vs \texttt{gemini}: Out-of-Procedure capture.} Across the turns of \texttt{c3e58bd88741baa1}, our model catches the OOP mistake; \texttt{{gemini}} misses or mis-corrects.}
% \vspace{4pt}
\scriptsize \turnpad{\begin{tabular}{@{}|p{\dimexpr\linewidth+2\tabcolsep}|@{}}
\hline
\rowcolor{colDialog} \textbf{Task}: How to inflate a basketball properly. \textit{[Source: \textbf{EgoProactive}]}  \\
\hline
\rowcolor{colDialog}
\textbf{Dialog history}:
\textbf{1.}\;Sure! Let's get started. First, Locate a basketball and a hand pump with a needle suitable for inflating sports balls. 
% \textbf{2.}\;\textbf{(mistake,} Tool Selection Error) Grab a bicycle pump that has a different needle type and does not fit the basketball valve. 
% \textbf{3.}\;\textbf{(fix)} Set the bicycle pump aside and retrieve the correct hand pump with a universal or basketball‑specific needle. 
% \textbf{4.}\;Wipe the basketball's valve area with a dry cloth to remove any dirt or debris. 
\\
\hline
\end{tabular}}

% \vspace{6pt}
% --- Turn 1 (interrupt, [2.7-4.8s], ours=✓, gemini=✗) ---
\centering
\setlength{\tabcolsep}{2pt}
\renewcommand{\arraystretch}{1.05}
\scriptsize
\turnrib{2}{2.7--4.8s}{%
\begin{tabular}{@{}|>{\columncolor{colDialog}}p{0.4300\linewidth}|>{\columncolor{colDialog}}p{0.1800\linewidth}|>{\columncolor{colDialog}}p{0.1800\linewidth}|>{\columncolor{colDialog}}p{0.1600\linewidth}|>{\columncolor{colDialog}}p{0.0500\linewidth}|@{}}
\multicolumn{5}{@{}>{\columncolor{colDialog}}c@{}}{ %\makebox[0pt][r]{\raisebox{0.35in}{\large\bfseries T1}\hspace{0.4cm}}
\includegraphics[trim=0cm 0cm 0cm 8cm,clip,width=0.20\linewidth]{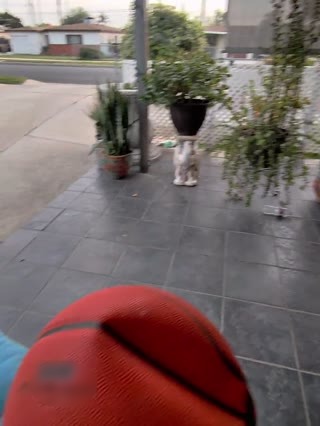}
\includegraphics[trim=0cm 0cm 0cm 8cm,clip,width=0.20\linewidth]{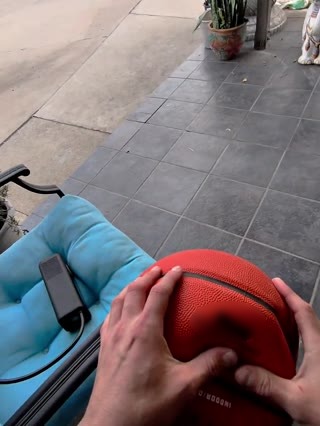}
\includegraphics[trim=0cm 0cm 0cm 8cm,clip,width=0.20\linewidth]{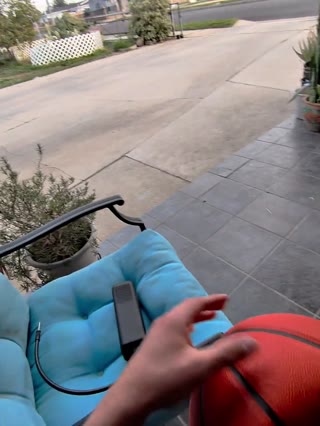}
\includegraphics[trim=0cm 0cm 0cm 8cm,clip,width=0.20\linewidth]{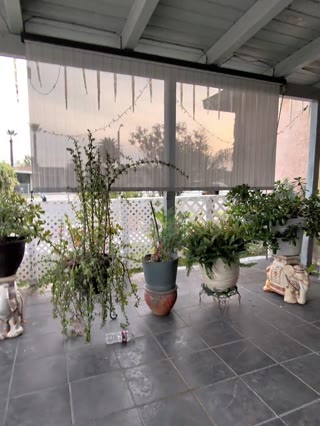}
\includegraphics[trim=0cm 0cm 0cm 8cm,clip,width=0.20\linewidth]{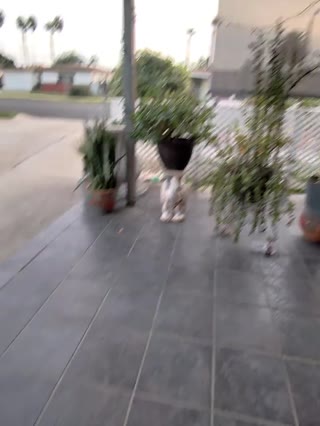}} \\
\hline
\textbf{Plan (prev $\rightarrow$ curr)} & \textbf{Cue (\$ interrupt\$)} & \textbf{Golden} & \textbf{Ours}\,{\textcolor{darkgreen}{\cmark}} & \begin{adjustbox}{max width=\linewidth}\textbf{\texttt{gemini}}\,{\textcolor{red}{\xmark}}\end{adjustbox} \\
\hline
\textit{Locate a basketball and a hand pump with a needle suitable for inflating sports balls.} $\rightarrow$ \textbf{\textbf{(mistake,} Tool Selection Error) Grab a bicycle pump that has a different needle type and does not fit the basketball valve.} &
% inflation needle remains visibly outside the valve opening despite multiple contact attempts \newline clear size mismatch between the needle tip and valve port is evident \newline hands pause with the tool pressed against the ball without achieving secure insertion \newline the ball shows no signs of inflation such as surface tension change \newline pumping motion has stopped indicating recognition of incompatibility
% Needle failing to seat in valve, size mismatch evident, ball showing no inflation, pumping stopped.
Needle failing to seat in valve, size mismatch evident, no inflation visible, pumping stopped. &
Tool Selection Error) Grab a bicycle pump that has a different needle type and does not fit the basketball valve. &
Grab a bicycle pump that has a different needle type and does not fit the basketball valve. &
\textcolor{gray}{\textbf{[silent]}} \\
\hline
\end{tabular}}

% \vspace{6pt}
% --- Turn 2 (interrupt, [4.8-9.3s], ours=✓, gemini=✗) ---
\centering
\setlength{\tabcolsep}{2pt}
\renewcommand{\arraystretch}{1.05}
\turnrib{3}{4.8--9.3s}{%
\scriptsize\begin{tabular}{@{}|>{\columncolor{colDialog}}p{0.4100\linewidth}|>{\columncolor{colDialog}}p{0.1900\linewidth}|>{\columncolor{colDialog}}p{0.1900\linewidth}|>{\columncolor{colDialog}}p{0.1600\linewidth}|>{\columncolor{colDialog}}p{0.0500\linewidth}|@{}}
\multicolumn{5}{@{}>{\columncolor{colDialog}}c@{}}{
%\makebox[0pt][r]{\raisebox{0.35in}{\large\bfseries T2}\hspace{0.4cm}}
\includegraphics[trim=0cm 0cm 0cm 6cm,clip,width=0.20\linewidth]{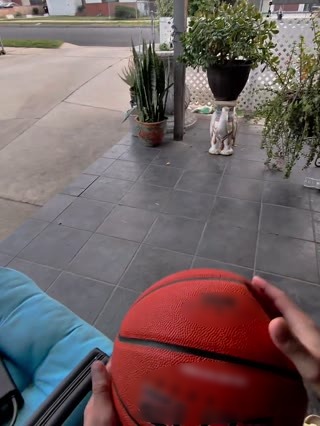}
\includegraphics[trim=0cm 0cm 0cm 6cm,clip,width=0.20\linewidth]{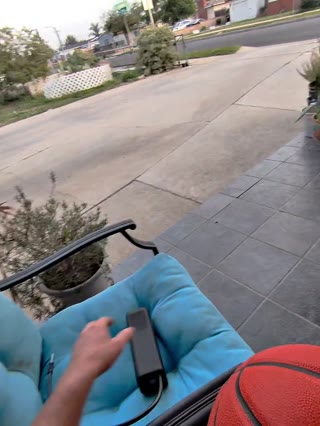}
\includegraphics[trim=0cm 0cm 0cm 6cm,clip,width=0.20\linewidth]{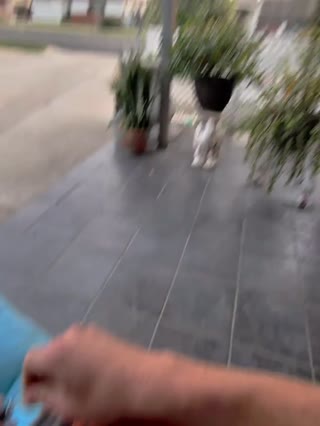}
\includegraphics[trim=0cm 0cm 0cm 6cm,clip,width=0.20\linewidth]{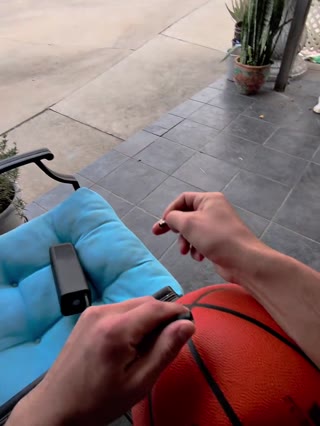}
\includegraphics[trim=0cm 0cm 0cm 6cm,clip,width=0.20\linewidth]{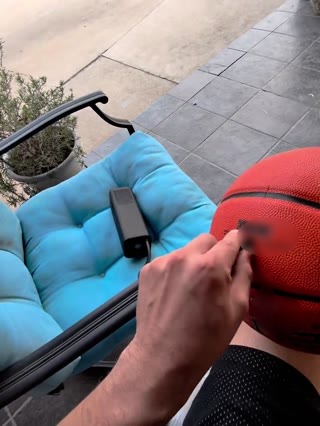}} \\
\hline
\textbf{Plan (prev $\rightarrow$ curr)} & \textbf{Cue (\$ interrupt\$)} & \textbf{Golden} & \textbf{Ours}\,{\textcolor{darkgreen}{\cmark}} & \begin{adjustbox}{max width=\linewidth}\textbf{\texttt{gemini}}\,{\textcolor{red}{\xmark}}\end{adjustbox} \\
\hline
\textit{\textbf{(mistake,} Tool Selection Error) Grab a bicycle pump that has a different needle type and does not fit the basketball valve.} $\rightarrow$ \textbf{\textbf{(fix)} Set the bicycle pump aside and retrieve the correct hand pump with a universal or basketball‑specific needle.} &
% The inflation needle is fully inserted into the ball's valve and connected to the pump currently held \newline The pump intended for sports balls is in active position while any alternative pump remains set aside and untouched \newline The user's hands have transitioned from tool swapping to a stable grip ready for pumping \newline The hose shows tension indicating a secure connection to the ball rather than being manipulated \newline Only one pump is engaged with the ball, confirming tool selection is finalized
% Needle fully inserted and pump actively engaged, hands stable and ready for pumping, hose taut confirming secure connection.
Needle fully inserted with pump engaged, hands stable and ready to pump, hose taut confirming secure connection. &
Set the bicycle pump aside and retrieve the correct hand pump with a universal or basketball‑specific needle. &
Put the bicycle pump aside and grab a hand pump with a universal or basketball-specific needle. &
\textcolor{gray}{\textbf{[silent]}} \\
\hline
\end{tabular}}

% \vspace{6pt}
% --- Turn 3 (interrupt, [9.3-13.3s], ours=✓, gemini=✗) ---
\centering
\setlength{\tabcolsep}{2pt}
\renewcommand{\arraystretch}{1.05}
\scriptsize
\turnrib{4}{9.3--13.3s}{%
\scriptsize\begin{tabular}{@{}|>{\columncolor{colDialog}}p{0.4100\linewidth}|>{\columncolor{colDialog}}p{0.1900\linewidth}|>{\columncolor{colDialog}}p{0.1900\linewidth}|>{\columncolor{colDialog}}p{0.1600\linewidth}|>{\columncolor{colDialog}}p{0.0500\linewidth}|@{}}
\multicolumn{5}{@{}>{\columncolor{colDialog}}c@{}}{
% \makebox[0pt][r]{\raisebox{0.35in}{\large\bfseries T3}\hspace{0.4cm}}
\includegraphics[trim=0cm 0cm 0cm 5cm,clip,width=0.20\linewidth]{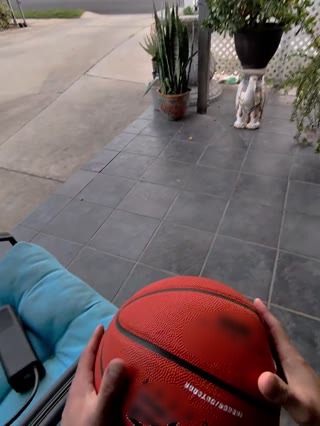}
\includegraphics[trim=0cm 0cm 0cm 5cm,clip,width=0.20\linewidth]{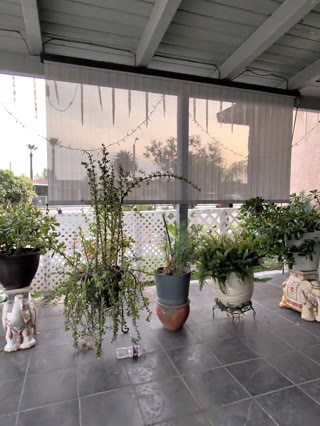}
\includegraphics[trim=0cm 0cm 0cm 5cm,clip,width=0.20\linewidth]{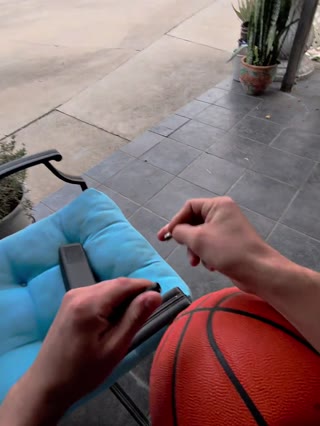}
\includegraphics[trim=0cm 0cm 0cm 5cm,clip,width=0.20\linewidth]{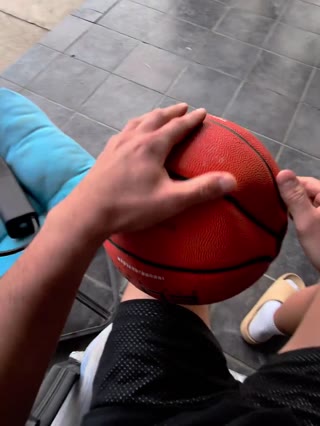}
\includegraphics[trim=0cm 0cm 0cm 5cm,clip,width=0.20\linewidth]{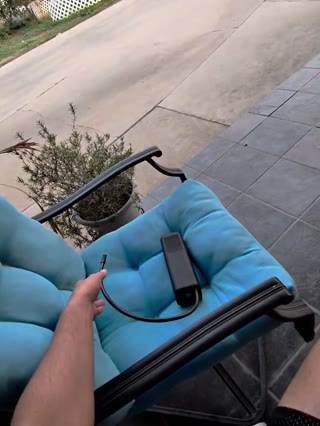}} \\
\hline
\textbf{Plan (prev $\rightarrow$ curr)} & \textbf{Cue (\$ interrupt\$)} & \textbf{Golden} & \textbf{Ours}\,{\textcolor{darkgreen}{\cmark}} & \begin{adjustbox}{max width=\linewidth}\textbf{\texttt{gemini}}\,{\textcolor{red}{\xmark}}\end{adjustbox} \\
\hline
\textit{\textbf{(fix)} Set the bicycle pump aside and retrieve the correct hand pump with a universal or basketball‑specific needle.} $\rightarrow$ \textbf{Wipe the basketball's valve area with a dry cloth to remove any dirt or debris.} &
% valve area fully exposed with no cloth or hand covering it after wiping motion has ceased \newline hands positioned away from the valve indicating the cleaning action has ended \newline cleaning cloth resting on a nearby surface rather than being applied to the ball \newline valve opening appears free of visible dirt or moisture \newline person transitions to preparing the inflation tool as the next action
Valve exposed and clean, hands and cloth withdrawn, person transitioning to inflation tool. &
Wipe the basketball's valve area with a dry cloth to remove any dirt or debris. &
Use a dry cloth to wipe any dirt or debris from the basketball's valve.
& \textcolor{gray}{\textbf{[silent]}} \\
\hline
\end{tabular}}

% \vspace{6pt}
% --- Turn 4 (interrupt, [13.3-17.9s], ours=✓, gemini=✗) ---
\centering
\setlength{\tabcolsep}{2pt}
\renewcommand{\arraystretch}{1.05}
\scriptsize
\turnrib{5}{13.3--17.9s}{%
\scriptsize\begin{tabular}{@{}|>{\columncolor{colDialog}}p{0.4000\linewidth}|>{\columncolor{colDialog}}p{0.2200\linewidth}|>{\columncolor{colDialog}}p{0.1800\linewidth}|>{\columncolor{colDialog}}p{0.1500\linewidth}|>{\columncolor{colDialog}}p{0.0500\linewidth}|@{}}
\multicolumn{5}{@{}>{\columncolor{colDialog}}c@{}}{
%\makebox[0pt][r]{\raisebox{0.35in}{\large\bfseries T4}\hspace{0.4cm}}
\includegraphics[trim=0cm 0cm 0cm 6cm,clip,width=0.20\linewidth]{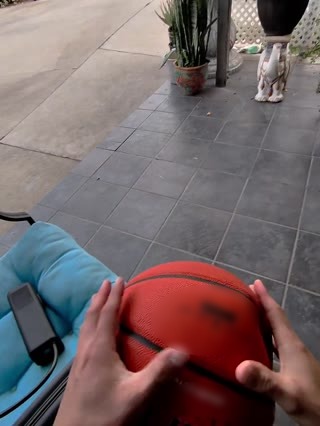}
\includegraphics[trim=0cm 0cm 0cm 6cm,clip,width=0.20\linewidth]{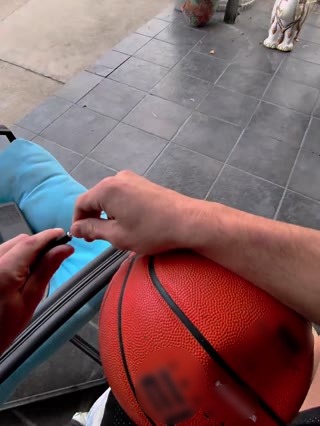}
\includegraphics[trim=0cm 0cm 0cm 6cm,clip,width=0.20\linewidth]{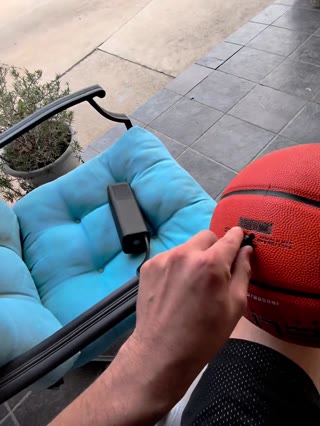}
\includegraphics[trim=0cm 0cm 0cm 6cm,clip,width=0.20\linewidth]{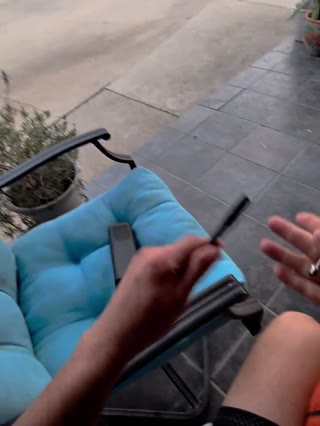}
\includegraphics[trim=0cm 0cm 0cm 6cm,clip,width=0.20\linewidth]{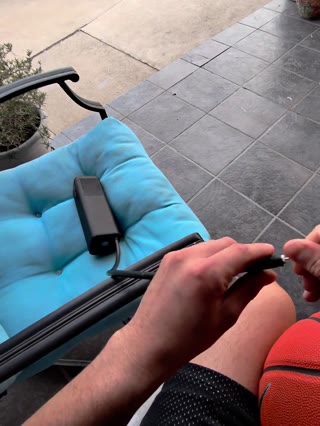}} \\
\hline
\textbf{Plan (prev $\rightarrow$ curr)} & \textbf{Cue (\$ interrupt\$)} & \textbf{Golden} & \textbf{Ours}\,{\textcolor{darkgreen}{\cmark}} & \begin{adjustbox}{max width=\linewidth}\textbf{\texttt{gemini}}\,{\textcolor{red}{\xmark}}\end{adjustbox} \\
\hline
\textit{Wipe the basketball's valve area with a dry cloth to remove any dirt or debris.} $\rightarrow$ \textbf{\textbf{(mistake,} Positioning/Alignment Error) Insert the pump needle at a slight angle, causing air to leak out.} &
% needle is fully seated in the valve and visibly tilted relative to the valve's central axis \newline hands have stopped active pushing and are now stabilizing rather than advancing the needle \newline no further forward motion of the needle is observed, indicating the insertion action has ended \newline the needle remains at a non-perpendicular angle to the valve opening, confirming the misalignment \newline the pump hose is stationary and connected, ready for inflation despite the angle error
% Needle fully seated but tilted off-axis, hands stabilizing rather than advancing, hose stationary and connected.
Needle fully seated but tilted off-axis, hands stabilizing, hose stationary and connected. &
Insert the pump needle at a slight angle, causing air to leak out. &
Insert the pump needle at a slight angle, letting air leak out. &
\textcolor{gray}{\textbf{[silent]}} \\
\hline
\end{tabular}}

% % \vspace{6pt}
% % --- Turn 5 (interrupt, [17.9-25.9s], ours=✓, gemini=✓) ---
\centering
\setlength{\tabcolsep}{2pt}
\renewcommand{\arraystretch}{1.05}
\scriptsize
\turnrib{6}{17.9--25.9s}{%
\begin{tabular}{@{}|>{\columncolor{colDialog}}p{0.3200\linewidth}|>{\columncolor{colDialog}}p{0.2100\linewidth}|>{\columncolor{colDialog}}p{0.1800\linewidth}|>{\columncolor{colDialog}}p{0.1500\linewidth}|>{\columncolor{colDialog}}p{0.1400\linewidth}|@{}}
\multicolumn{5}{@{}>{\columncolor{colDialog}}c@{}}{
% \makebox[0pt][r]{\raisebox{0.35in}{\large\bfseries T5}\hspace{0.4cm}}
\includegraphics[trim=0cm 0cm 0cm 6cm,clip,width=0.20\linewidth]{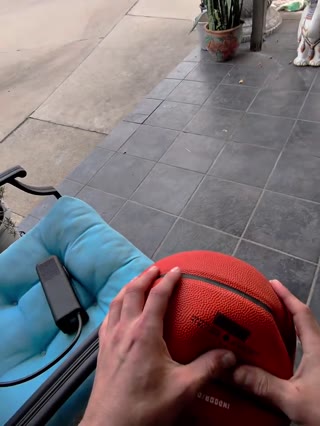}
\includegraphics[trim=0cm 0cm 0cm 6cm,clip,width=0.20\linewidth]{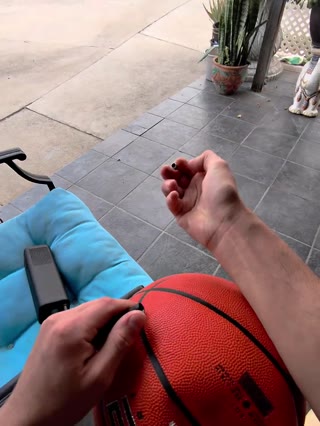}
\includegraphics[trim=0cm 0cm 0cm 6cm,clip,width=0.20\linewidth]{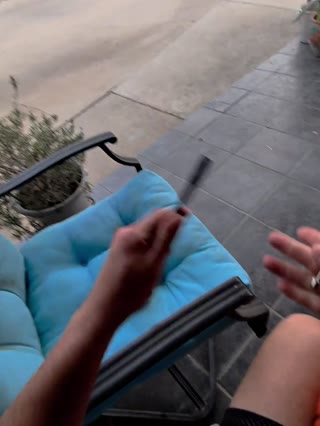}
\includegraphics[trim=0cm 0cm 0cm 6cm,clip,width=0.20\linewidth]{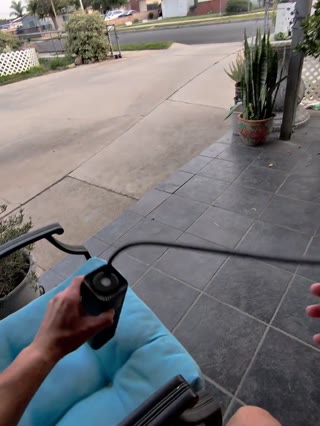}
\includegraphics[trim=0cm 0cm 0cm 6cm,clip,width=0.20\linewidth]{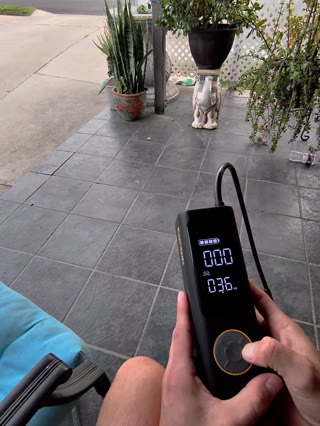}} \\
\hline
\textbf{Plan (prev $\rightarrow$ curr)} & \textbf{Cue (\$ interrupt\$)} & \textbf{Golden} & \textbf{Ours}\,{\textcolor{darkgreen}{\cmark}} & \begin{adjustbox}{max width=\linewidth}\textbf{\texttt{gemini}}\,{\textcolor{darkgreen}{\cmark}}\end{adjustbox} \\
\hline
\textit{\textbf{(mistake,} Positioning/Alignment Error) Insert the pump needle at a slight angle, causing air to leak out.} $\rightarrow$ \textbf{\textbf{(fix)} Remove the needle, realign it straight into the valve, and reinsert it firmly to create a tight seal.} &
% needle fully seated into the valve with no exposed shaft visible above the ball surface \newline hand releases the needle and moves away while the hose remains taut and stable \newline digital or analog pressure display begins to rise steadily indicating a sealed connection \newline needle positioned perpendicular to the ball surface forming a straight line with the valve \newline absence of visible air leakage signs such as movement or bubbles around the insertion point
Needle fully seated straight into valve, hose taut after hand release, pressure gauge rising steadily with no visible air leakage.” &
\textbf{(fix)} Remove the needle, realign it straight into the valve, and reinsert it firmly to create a tight seal. &
Remove the needle, realign it straight into the valve, and reinsert it firmly to create a tight seal. &
\textbf{(fix)} Remove the needle and re-insert it straight into the valve. \\
\hline
\end{tabular}}

\captionsetup{skip=2pt,font=footnotesize,labelsep=period,labelfont=bf}\captionof{figure}{\textbf{Ours vs \texttt{gemini}}: out-of-procedure capture across a session.}
\label{fig:mt_gemini}
\par\endgroup\medskip

\FloatBarrier

% ============================================================
\section{Training Details}
\label{app:training-recipe}
We train LoRA interaction models for Gemma~4 31B, Muse Glimmer 30B, and
Qwen3.8-VL-27B on EgoExo4D, EPIC-KITCHENS, HowTo100M, and Ego4D;
HoloAssist and EgoProactive are excluded from training. The four sources are
sampled with fixed square-root-proportional weights of 0.207, 0.225, 0.288,
and 0.280, respectively. Each example contains the current visual observation,
the preceding dialogue, and, when applicable, the plan state. The loss is
computed only on the final assistant turn so that supervision targets the
current \texttt{interrupt}/\texttt{silent} decision and any generated guidance.
All three models use LoRA rank 64, scaling factor 128, dropout 0.05, and no
bias adaptation. Gemma adapts all linear modules. Muse adapts the language
self-attention query, key, value, and output projections and keeps its vision
tower frozen. Qwen adapts those attention projections together with the
feed-forward gate, up, and down projections. We use AdamW with learning rate
$2\times10^{-5}$, $\beta=(0.9,0.95)$, weight decay 0.1, gradient clipping at
1.0, and cosine decay after a 3\% warmup. Inputs are truncated to 32,768 tokens
and at most 30 video frames. The exposure-matched schedules process up to
448,000 training examples, corresponding to 875 optimizer updates for Gemma
and 7,000 for Muse and Qwen under their respective distributed batch
configurations.
Checkpoint selection uses a deterministic, source-stratified 10\% validation
holdout whose session or procedure groups do not overlap training. We average
validation loss equally across the four sources, apply exponential smoothing
with coefficient 0.5, and select the checkpoint with the lowest smoothed macro
validation loss. Training stops early after 20 validation rounds without an
improvement of at least $10^{-3}$. PWR uses the selected interaction checkpoint
conditioned on the output of its separately trained planner. The
plan-conditioned control uses the matched plan context without changing the
reported evaluation examples.

% ============================================================
\section{Domain Taxonomy and Domain-Wise Results}
\label{app:domain-results}
Table~\ref{tab:domain-taxonomy} summarizes the source-reported domain labels in
the frozen evaluation rows. The generic label is retained verbatim where a
source does not provide a finer category; no semantic relabeling is inferred.
Table~\ref{tab:domain-results} reports the corresponding learned-planner Turn
results.
The 14-domain count in the main paper refers to the normalized taxonomy across
the full six-source training and evaluation collection. Table~\ref{tab:domain-taxonomy}
instead reports only labels materialized in the five-source frozen evaluation
manifest. Three sources expose only the generic label there, so the six rows
below are evaluation-manifest groups rather than a replacement for the
benchmark-wide 14-domain taxonomy.
\begin{table}[t]
\centering\small
\begin{tabular}{@{}l l r@{}}
\toprule
\textbf{Source domain} & \textbf{Evaluation datasets} & \textbf{Decision points} \\
\midrule
Arts and Crafts & EgoProactive & 2{,}391 \\
Chef & EgoProactive & 2{,}926 \\
Handyman & EgoProactive & 2{,}241 \\
Tutorial & EgoProactive & 2{,}341 \\
cooking & EPIC-KITCHENS & 6{,}399 \\
general & EgoExo4D{,} HoloAssist{,} HowTo100M & 10{,}576 \\
\bottomrule
\end{tabular}
\caption{Source-reported domain taxonomy in the frozen evaluation rows. Counts use the exact terminal evaluation UUID sets.}
\label{tab:domain-taxonomy}
\end{table}
\begin{table*}[t]
\centering\small
\setlength{\tabcolsep}{3pt}
\begin{tabular*}{\textwidth}{@{\extracolsep{\fill}}l l r ccc@{}}
\toprule
\textbf{Domain} & \textbf{Model} & \textbf{Decision points} & \textbf{Interrupt F1} & \textbf{Silent F1} & \textbf{Macro-F1} \\
\midrule
Arts and Crafts & Gemma-31B & 2{,}391 & \underline{.646} & \underline{.548} & \textbf{.597} \\
Arts and Crafts & Muse-30B & 2{,}391 & .363 & \textbf{.652} & .507 \\
Arts and Crafts & Qwen-27B & 2{,}391 & \textbf{.720} & .317 & \underline{.518} \\
\cmidrule(lr){1-6}
Chef & Gemma-31B & 2{,}926 & \underline{.532} & \underline{.632} & \textbf{.582} \\
Chef & Muse-30B & 2{,}926 & .299 & \textbf{.678} & .488 \\
Chef & Qwen-27B & 2{,}926 & \textbf{.662} & .501 & \underline{.581} \\
\cmidrule(lr){1-6}
Handyman & Gemma-31B & 2{,}241 & \underline{.601} & \underline{.548} & \textbf{.574} \\
Handyman & Muse-30B & 2{,}241 & .353 & \textbf{.657} & .505 \\
Handyman & Qwen-27B & 2{,}241 & \textbf{.729} & .405 & \underline{.567} \\
\cmidrule(lr){1-6}
Tutorial & Gemma-31B & 2{,}341 & \underline{.625} & \underline{.577} & \textbf{.601} \\
Tutorial & Muse-30B & 2{,}341 & .336 & \textbf{.653} & .494 \\
Tutorial & Qwen-27B & 2{,}341 & \textbf{.728} & .354 & \underline{.541} \\
\cmidrule(lr){1-6}
cooking & Gemma-31B & 6{,}399 & \textbf{.757} & \underline{.538} & \textbf{.648} \\
cooking & Muse-30B & 6{,}399 & .490 & \textbf{.576} & .533 \\
cooking & Qwen-27B & 6{,}399 & \underline{.748} & .333 & \underline{.540} \\
\cmidrule(lr){1-6}
general & Gemma-31B & 10{,}576 & \underline{.739} & \underline{.401} & \textbf{.570} \\
general & Muse-30B & 10{,}576 & .455 & \textbf{.504} & .480 \\
general & Qwen-27B & 10{,}576 & \textbf{.787} & .310 & \underline{.548} \\
\bottomrule
\end{tabular*}
\caption{Domain-wise Turn results for PWR. Results use the frozen predictions and source-reported evaluation-domain labels. This table analyzes variation within PWR; aggregate matched comparisons are reported in Table~\ref{tab:main}.}
\label{tab:domain-results}
\end{table*}
% ====================================================================================
\section{EgoProactive Collection Details}
\label{app:wp}
\label{app:EgoProactive}
This section provides additional details on EgoProactive, introduced in
\S\ref{sec:egoproactive}. EgoProactive is the only source in
Pro\textsuperscript{2}Bench collected specifically for proactive assistance.
It contains continuous smart-glasses video, out-of-plan annotations at the
procedure level, and recovery guidance.
\subsection{Collection Protocol}
\label{app:wp-protocol}
\paragraph{Device and recording format.}
Each released procedure is recorded as one continuous video from consumer
smart glasses. The first-person viewpoint matches the intended wearable
setting and naturally includes head motion, changing illumination, partial
occlusion, and variation in object placement. Participants perform the task at
a natural pace without narrating unless speech occurs naturally.
\paragraph{Task assignment.}
Each recording follows a procedural script containing a goal, an ordered set
of subgoals, and executable action steps. Participants may adapt incidental
details, such as the particular object or ingredient used, provided that the
task goal and procedural structure remain unchanged. The final release covers
procedures with 3 to 15 reference steps, with a mean of 7.6 steps.
\paragraph{Recording requirements.}
All required actions must remain visible from start to finish in a single
first-person recording. Written instructions may not appear in view, the
camera may not be placed on a surface, and the principal objects and hand
actions must remain sufficiently visible for temporal annotation. Videos with
an unrecoverable mismatch between the script and the visible execution are not
included in the released evaluation manifest.
\subsubsection{Released Dataset Accounting}
\label{app:wp-accounting-sec}
We distinguish \emph{procedures}, \emph{videos}, and \emph{decision points}.
A procedure is one recorded execution of a scripted task, represented by one
continuous video. A decision point is an observation window at which the
assistant must emit either \texttt{\$interrupt\$} with guidance or
\texttt{\$silent\$}. One procedure therefore contributes multiple decision
points.
\begin{table}[t]
\centering
\caption{EgoProactive accounting in the frozen evaluation manifest. OOP
decision points are all evaluated points belonging to OOP procedures, not a
count of deviation onsets.}
\label{tab:wp-accounting}
\setlength{\tabcolsep}{5pt}
\renewcommand{\arraystretch}{1.12}
\small
\begin{tabular}{@{}l r@{}}
\toprule
\textbf{Quantity} & \textbf{Count} \\
\midrule
Released procedures & 700 \\
Released videos & 700 \\
Video duration & 31.4 hours \\
Activity domains & 4 \\
\midrule
Evaluation decision points & 9{,}899 \\
\quad Interrupt & 5{,}338 \\
\quad Silent & 4{,}561 \\
\midrule
OOP procedures & 247 \\
Decision points in OOP procedures & 3{,}810 \\
\bottomrule
\end{tabular}
\end{table}
The median procedure contributes 14 decision points, with a range of 4 to 30.
All 700 EgoProactive procedures are held out from planner and interaction-model training
in the experiments reported in this paper. The released evaluation manifest
does not expose participant identifiers or demographic attributes, so we do
not report participant-level counts or demographic distributions.
\paragraph{Activity domains.}
The four domain labels in the final evaluation manifest are shown in
Table~\ref{tab:wp-domains}.
\begin{table}[t]
\centering
\caption{EgoProactive coverage by the domain labels in the frozen evaluation
manifest.}
\label{tab:wp-domains}
\setlength{\tabcolsep}{6pt}
\renewcommand{\arraystretch}{1.10}
\small
\begin{tabular}{@{}l r r@{}}
\toprule
\textbf{Domain} & \textbf{Procedures} & \textbf{Decision points} \\
\midrule
Chef & 174 & 2{,}926 \\
Arts and Crafts & 175 & 2{,}391 \\
Handyman & 175 & 2{,}241 \\
Tutorial & 176 & 2{,}341 \\
\midrule
\textbf{Total} & \textbf{700} & \textbf{9{,}899} \\
\bottomrule
\end{tabular}
\end{table}
\subsection{Out-of-Plan Scenario Design}
\label{app:wp-oop}
Of the 700 procedures, 247 contain a scripted departure from the expected
execution. Together, these procedures contribute 3{,}810 of the 9{,}899
decision points. This procedure-level membership is the canonical
EgoProactive-OOP slice used throughout the paper. It must not be interpreted as
3{,}810 distinct mistakes or deviation onsets: each OOP procedure also contains
ordinary progress, silence, deviation, redirection, and recovery points.
The scripts cover departures such as skipped or reordered actions, incorrect
tools or techniques, and attention-shifting interruptions. Each script is
designed around a continuous sequence
\[
\text{correct lead-in}\;\rightarrow\;\text{departure}\;\rightarrow\;
\text{recovery}\;\rightarrow\;\text{correct end state}.
\]
The reference plan and guidance identify the corrective action needed to
return to the task. For annotation, a departure begins at the first frame at
which the observed action can be distinguished from the expected execution.
Recovery points identify the action that restores progress toward the task
goal.
\paragraph{OOP target audit.}
\label{app:wp-oop-target-audit}
We re-audit every OOP interrupt target against the frames and dialogue
available at that decision point. A valid target must correct the observed
departure or re-engage the user with the procedure. Targets that instead
instruct the user to perform the scripted error, reveal a later recovery before
the departure is observable, or conflict with the visible state are corrected
or excluded before evaluation. Annotation tags and future mistake or recovery
text are retained only as scoring metadata and are removed from all model
inputs as described in Appendix~\ref{app:eval-input-sanitization}.
\paragraph{OOP annotation categories.}
The final manifest assigns one stable category to each procedure. Canonical
procedures are marked as ordinary executions; OOP procedures are categorized
by whether the record contains an explicit mistake and fix pair, an inferred
fix, a fix without a separately retained mistake point, or an interruption
with an explicit redirect. This procedure-level category defines the OOP
slice. Turn-level fields identify the particular mistake, deviation,
interruption, or recovery points inside that procedure.
\begin{table}[t]
\centering
\caption{Structured OOP support in EgoProactive relative to established
egocentric procedural sources. ``Partial'' denotes naturally occurring
instructor corrections without paired onset and recovery annotations.}
\label{tab:oop-comparison}
\setlength{\tabcolsep}{3pt}
\renewcommand{\arraystretch}{1.12}
\small
\begin{tabular}{@{}l cccc@{}}
\toprule
\textbf{Feature} & \textbf{EP} & \textbf{HA} & \textbf{EK} & \textbf{Ego4D} \\
\midrule
Scripted departures & \cmark & \xmark & \xmark & \xmark \\
In-video recovery & \cmark & Partial & \xmark & \xmark \\
Departure onset & \cmark & \xmark & \xmark & \xmark \\
Paired departure--recovery guidance & \cmark & \xmark & \xmark & \xmark \\
\bottomrule
\end{tabular}
\\[2pt]
{\footnotesize EP = EgoProactive, HA = HoloAssist, and EK =
EPIC-KITCHENS.}
\end{table}
\subsection{Annotation and Human Review}
\label{app:wp-annotation}
The annotation pipeline first proposes procedural steps, temporal boundaries,
decision labels, guidance, and visual cues. These are candidates rather than
ground truth. EgoProactive then undergoes a two-phase human review that checks
the visible execution against the proposed step sequence.
\paragraph{Phase 1: audit.}
For every candidate step, annotators inspect the corresponding video segment
and answer three diagnostic questions: (1) whether the description matches the
visible action, (2) whether the segment contains only the described action,
and (3) whether the action completes within the proposed boundary. A negative
answer routes the video to correction. Figure~\ref{fig:audit_workflow} shows a
boundary error detected by comparing adjacent steps.
\begin{figure}[t]
\centering
\begin{subfigure}[t]{0.48\linewidth}
\centering
\includegraphics[width=\linewidth]{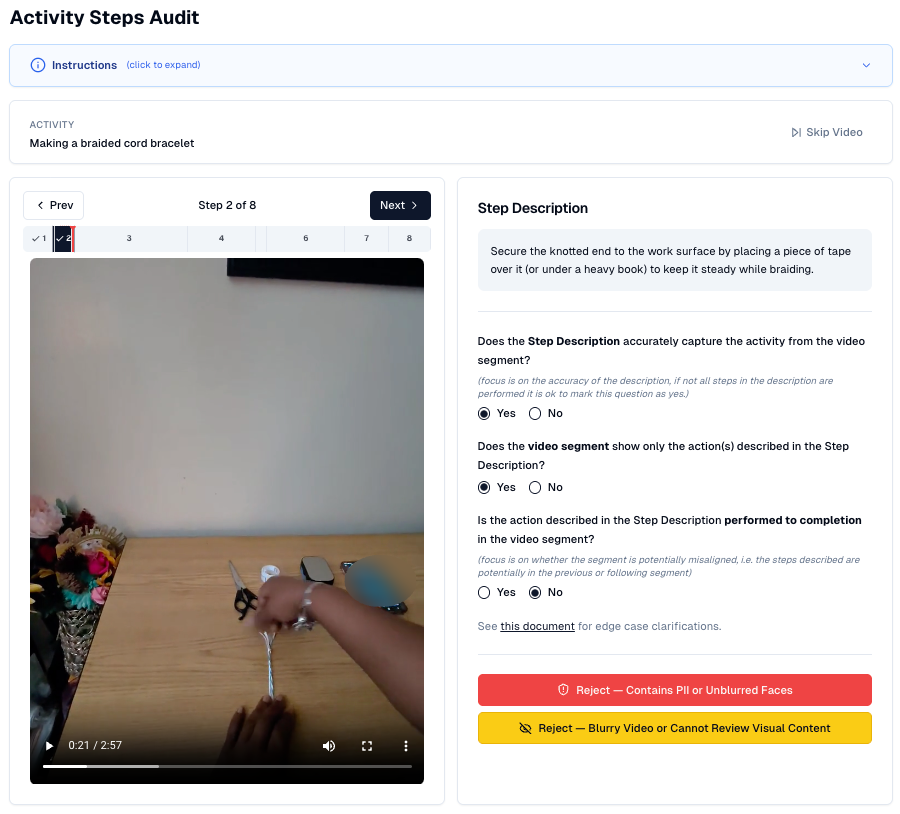}
\caption{The described securing action continues beyond the proposed segment,
so the step fails the completeness check.}
\label{fig:audit_step2}
\end{subfigure}
\hfill
\begin{subfigure}[t]{0.48\linewidth}
\centering
\includegraphics[width=\linewidth]{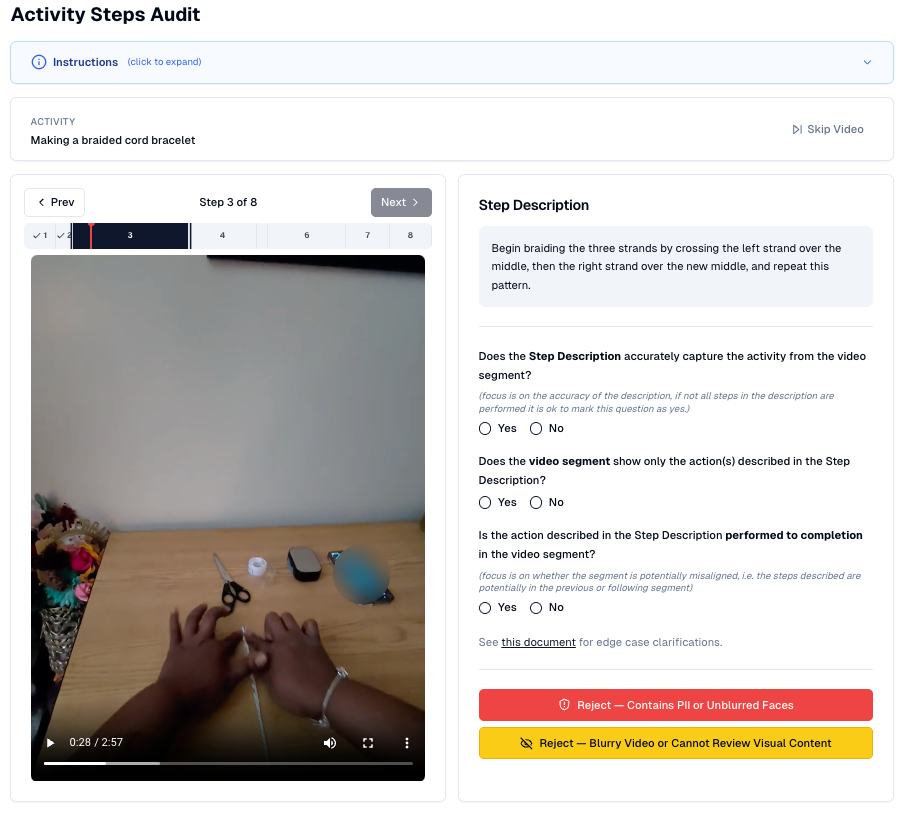}
\caption{The next segment begins with the preceding action, confirming the
boundary error.}
\label{fig:audit_step3}
\end{subfigure}
\caption{\textbf{Phase 1 audit.} Adjacent segments are inspected together to
identify textual and temporal inconsistencies before correction.}
\label{fig:audit_workflow}
\end{figure}
\paragraph{Phase 2: correction.}
Annotators adjust temporal boundaries at 0.1-second resolution and edit step
descriptions until the three checks pass. Boundary changes propagate to the
adjacent segment to preserve a contiguous timeline. The interface rejects
gaps and overlaps and presents a final preview of all edits. If a coherent
step decomposition cannot be recovered, the video is excluded rather than
forced into the release. Figures~\ref{fig:correction_flagged}--
\ref{fig:correction_preview} illustrate this workflow.
\paragraph{Review coverage and agreement.}
Every released EgoProactive procedure passes the audit above, and every
flagged procedure passes a subsequent correction review before release.
\begin{figure}[t]
\centering
\includegraphics[width=0.72\linewidth]{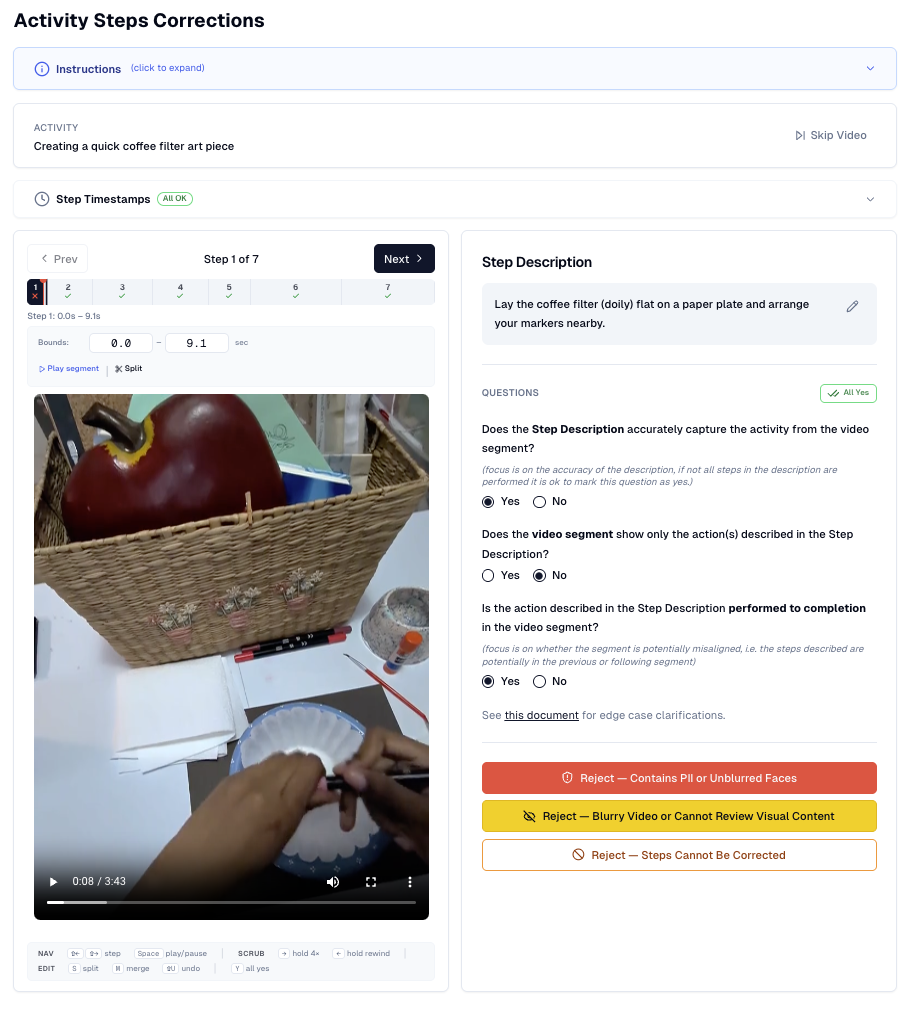}
\caption{\textbf{Correction, initial state.} A segment is flagged because it
contains action belonging to an adjacent step.}
\label{fig:correction_flagged}
\end{figure}
\begin{figure}[t]
\centering
\includegraphics[width=0.72\linewidth]{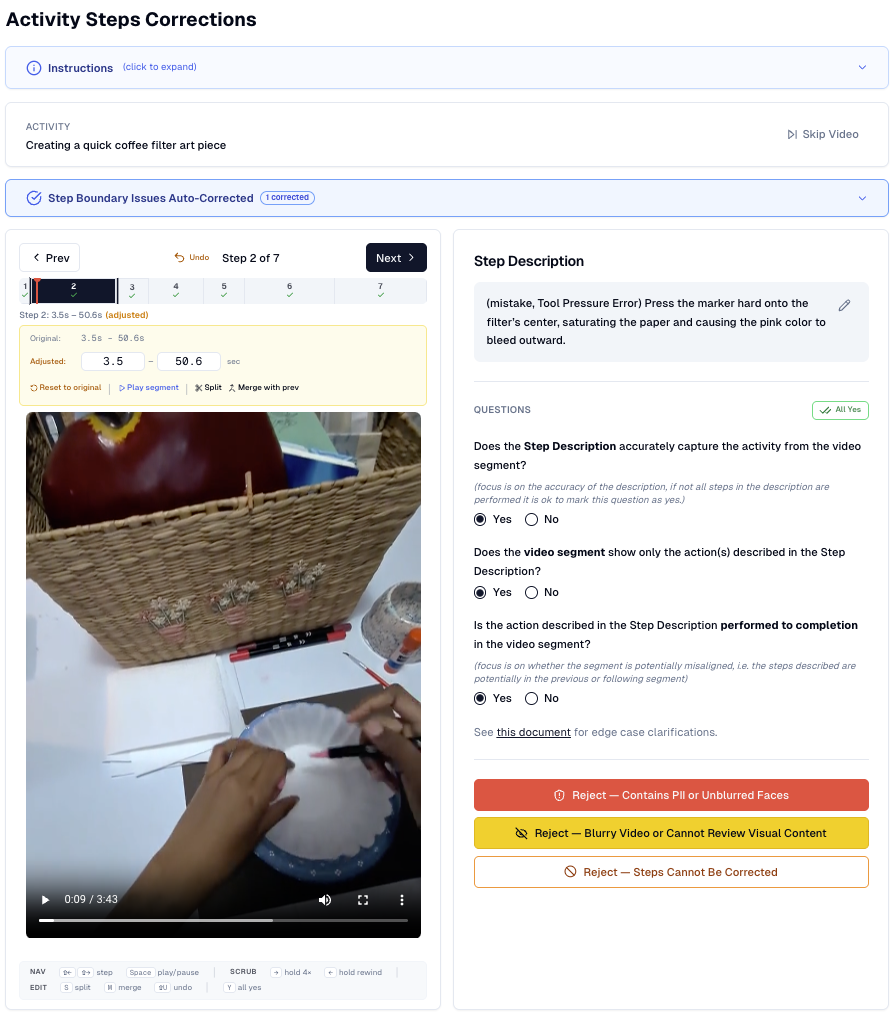}
\caption{\textbf{Correction, boundary edit.} The annotator moves the boundary,
replays the segment, and verifies the corrected description and timing.}
\label{fig:correction_adjusted}
\end{figure}
\begin{figure}[t]
\centering
\includegraphics[width=0.72\linewidth]{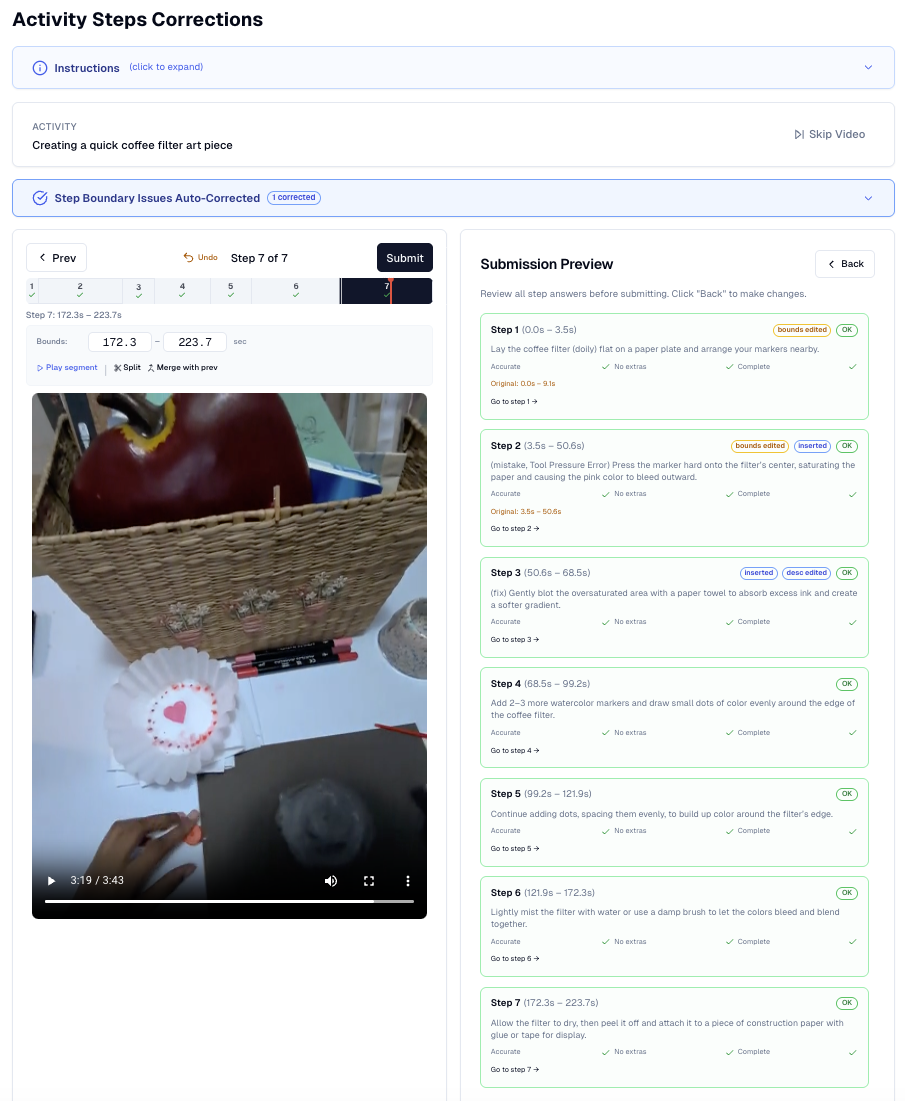}
\caption{\textbf{Correction, final review.} The preview exposes every boundary,
insertion, and text edit before submission.}
\label{fig:correction_preview}
\end{figure}
\paragraph{Label provenance.}
Temporal boundaries and step text begin as machine proposals and may be edited
during human review. OOP onset and recovery structure are attached to the
procedure annotation. The interrupt or silent target is then derived from the
verified event type, and every interrupt point carries a reference guidance
utterance. Visual completion and incompletion cues are retained when supported
by the reviewed observation. Although the generation prompt requests multiple
cues, the prompt itself is not treated as evidence that every released row has
a fixed minimum number.
\subsection{Quality Assurance and Release Scope}
\label{app:wp-qa}
Quality assurance checks script visibility, first-person viewpoint, visual
clarity, goal completion, step coverage, and temporal consistency. OOP
procedures receive additional checks that the departure is visible, the
recovery occurs in the same continuous recording, and both can be localized in
time. The counts in Table~\ref{tab:wp-accounting} describe the accepted frozen
evaluation manifest.
\paragraph{Consent and data handling.}
Participants consented to the research collection and dataset use before
recording. Collection instructions prohibited written scripts in view and
restricted recording to suitable private task settings. Participants were
asked to minimize incidental bystander capture, and the released metadata
omits participant identifiers and demographic attributes. Access is governed
by the dataset release terms linked from the main paper.
The released decision-point metadata contains no participant identity or
demographic attributes. This protects contributor privacy but prevents a
participant-level fairness analysis from the public metadata alone. More
broadly, proactive assistance can reduce procedural errors, but poorly timed or
incorrect guidance can distract users or encourage over-reliance. Systems
deployed in safety-sensitive settings therefore require additional calibration,
human-factors evaluation, and application-specific safeguards beyond the
benchmark evaluation reported here.

% ============================================================
\section{Step Alignment Pipeline Details}
\label{app:pipeline}
This section describes the automated step-alignment component used to generate
candidate procedural annotations. It builds on the Action100M methodology
\citep{Chen2026Action100M}. The component does not directly produce the final
benchmark. Its candidates enter the source-specific normalization and
iterative verification process in \S\ref{section:data_annotation_pipeline};
EgoProactive candidates additionally undergo the human audit and correction
described in \S\ref{app:wp-annotation}.
\subsection{Stage 1: Visual Feature Extraction}
\label{app:pipeline-s1}
We encode the video with V-JEPA~2 using a ViT-Giant encoder at 384-pixel
resolution. Frames are temporally subsampled and processed in overlapping
64-frame windows with stride 32. A frame covered by more than one window is
represented by the mean of its window-specific embeddings:
\begin{equation}
\mathbf{e}_t=
\frac{1}{|\mathcal{W}(t)|}
\sum_{w\in\mathcal{W}(t)}\mathbf{e}^{(w)}_t,
\end{equation}
where \(\mathcal{W}(t)\) is the set of windows containing frame \(t\). The
result is a temporally ordered embedding matrix
\(\mathbf{F}\in\mathbb{R}^{T\times d}\).
\subsection{Stage 2: Temporally Constrained Clustering}
\label{app:pipeline-s2}
We apply hierarchical agglomerative clustering with Ward linkage. A
tridiagonal connectivity matrix permits merges only between temporally
adjacent clusters, so every node of the resulting dendrogram represents a
contiguous time span. The merge distance records the visual change required to
join neighboring spans. We retain the full dendrogram because later stages may
need to refine a coarse segment without recomputing the embeddings.
\subsection{Stage 3: Dendrogram Partitioning}
\label{app:pipeline-s3}
The dendrogram is cut by binary search over its merge-distance threshold. When
a source provides an expected step count \(N\), the initial partition targets
\(\lfloor1.3N\rfloor\) segments. When no step count is available, the target
is estimated from video duration. This deliberate over-segmentation favors
recovering action boundaries: a later grouping stage can merge adjacent
segments, whereas it cannot reliably recover a transition hidden inside one
coarse segment.
\subsection{Stage 4: Visual Captioning and Recursive Refinement}
\label{app:pipeline-s4}
For each segment, Qwen3-VL-235B receives up to 16 uniformly spaced frames and
describes the visible hand actions, tools, and manipulated objects. If no task
action is visible, it returns ``No active task.'' If the response contains
multiple distinct actions, the segment is replaced by its two dendrogram
children and each child is captioned independently. Recursion stops when the
caption describes one coherent activity or the dendrogram reaches a leaf.
The captioning prompt is:
\begin{quote}
\footnotesize\ttfamily\raggedright
These frames are from an egocentric (first-person) video of someone performing
a hands-on task. Describe what you see, with an emphasis on any action the
person is performing. If no action is visible, say ``No active task.''
If there are multiple distinct actions involving a different tool, object, or
motion, list each on a separate line.
Output only the activity description or descriptions, one per line.
\end{quote}
\subsection{Stage 5: Procedural Step Grouping}
\label{app:pipeline-s5}
Visual captions are observations of local activity, while a procedural step
may span several neighboring segments. A text-only Llama~4 Maverick call
receives the chronological captions, their timestamps, the activity goal, and
any source-provided reference steps. It merges adjacent observations that
describe one action, absorbs inactive transition segments, maps generic object
descriptions to task-specific terminology, and rewrites the result as concise
instructions. Candidate mistake, interruption, and recovery segments remain
separate rather than being merged into canonical steps. These labels support
annotation and scoring only; they are stripped from model inputs by the
sanitization step in Appendix~\ref{app:eval-input-sanitization}.
The grouping call returns a JSON array of objects with \texttt{caption},
\texttt{start\_ts}, and \texttt{end\_ts}. The prompt constrains the output to
chronological, non-overlapping groups and asks for a number of groups close to
the source step count when one exists.
\subsection{Stage 6: Reference-Step Alignment}
\label{app:pipeline-s6}
A second text-only call assigns each grouped segment to the best-matching
reference step. It considers both semantic similarity and the segment's
relative position in the video. Several consecutive segments may map to the
same reference step, and unmatched transitions receive index 0. The output is
a JSON array containing exactly one integer for each grouped segment.
The alignment prompt is:
\begin{quote}
\footnotesize\ttfamily\raggedright
Given numbered procedural steps and chronological video-segment captions,
assign each segment to the best-matching step. Use both semantic match and
temporal position. Use 0 when no step matches. Multiple consecutive segments
may map to one step, and not every step must be matched. Return only a JSON
array of integers with exactly one entry per segment.
\end{quote}
\subsection{Stage 7: Automated Candidate Filtering}
\label{app:pipeline-s7}
We first verify that the generated segments form a valid temporal sequence and
cover the intended video span. A text-only Llama~4 Maverick call then scores
the complete caption sequence for coverage of distinct task phases, logical
order, and instructional relevance, each on a 1 to 10 scale. Sequences with an
overall score below 5 are removed before downstream annotation.
The scoring prompt asks whether a novice with the required materials could
complete the activity by following the captions. Coverage rewards distinct
hands-on phases rather than repeated descriptions of one phase; order measures
procedural coherence; and relevance measures specificity and actionability.
The call returns a JSON object containing the overall, coverage, order, and
relevance scores together with a short explanation and issue list.
\subsection{From Candidates to the Released Benchmark}
\label{app:pipeline-release}
Stage~7 is an automated prefilter, not the final annotation decision. The
remaining candidates pass through four additional layers:
\begin{enumerate}[nosep,leftmargin=1.5em]
    \item \textbf{Source-specific normalization} maps source metadata,
    timestamps, and task identifiers into the common benchmark schema.
    \item \textbf{Field-level verification and correction} checks steps,
    boundaries, decisions, guidance, and visual cues against the corresponding
    frames. A separate correction pass revises only the flagged fields, as
    described in \S\ref{section:data_annotation_pipeline}.
    \item \textbf{EgoProactive human review} audits textual fidelity, temporal
    specificity, and step completeness, and corrects or rejects inconsistent
    videos using the protocol in \S\ref{app:wp-annotation}.
    \item \textbf{Canonicalization and release checks} assign stable procedure
    and decision-point identities, remove duplicate or malformed rows, enforce
    source-specific split constraints, and freeze the manifests used by every
    matched evaluation condition. This stage also removes annotation-only tags,
    targets, reference responses, and future deviation or recovery information
    from model-facing fields.
\end{enumerate}
This separation is important: the seven-stage alignment component proposes a
structured transcript, while the later review and release stages determine
which annotations enter Pro\textsuperscript{2}Bench. The statistics in
Tables~\ref{tab:benchmark}, \ref{tab:wp-accounting}, and
\ref{tab:app-eval-detail} are computed from the final frozen manifests, not
from intermediate alignment output.
EgoProactive receives the additional human review in
Appendix~\ref{app:wp-annotation}. The five established sources use the
source-specific automated verification and correction process above.
The alignment pipeline does not define the evaluation sample. The exact
decision-point inclusion rule and per-source denominators are given in
Appendix~\ref{app:eval-population}.
\subsection{Reproducibility and Limitations}
\label{app:pipeline-limitations}
Each final decision point retains stable source, video, procedure, and row
identifiers together with its temporal interval and annotation provenance.
The prompt contracts used by the model-dependent stages are specified above,
and aggregate tables are regenerated deterministically from the frozen manifests.
The pipeline has four principal limitations. First, ambiguous or occluded
frames can cause visual-captioning errors. Second, a fixed 16-frame captioning
budget may undersample long segments. Third, independent processing does not
enforce identical language across different videos of the same task. Fourth,
text-only alignment can propagate a captioning error into the selected
reference step. The iterative verification, EgoProactive human review, and
release checks reduce these failure modes but do not eliminate them.

% \medskip

% \input{tex/checklist}

\end{document}